\documentclass[10pt,journal,compsoc]{IEEEtran}
\ifCLASSOPTIONcompsoc
  \usepackage[nocompress]{cite}
\else
  \usepackage{cite}
\fi
\ifCLASSINFOpdf
\else
\fi

\usepackage{times}
\usepackage{epsfig}
\usepackage{graphicx}
\usepackage{subcaption}
\usepackage{amsmath}
\usepackage{amssymb}
\usepackage{stfloats}
\usepackage{epsfig}
\usepackage{multirow}
\usepackage[lined,boxed,commentsnumbered]{algorithm2e}
\usepackage[percent]{overpic}
\usepackage{array}
\usepackage{booktabs}
\usepackage{enumitem}
\usepackage{sidecap}

\begin{document}
%
% paper title
% Titles are generally capitalized except for words such as a, an, and, as,
% at, but, by, for, in, nor, of, on, or, the, to and up, which are usually
% not capitalized unless they are the first or last word of the title.
% Linebreaks \\ can be used within to get better formatting as desired.
% Do not put math or special symbols in the title.
\title{Online Object Tracking, Learning and Parsing with And-Or Graphs}
%
%
% author names and IEEE memberships
% note positions of commas and nonbreaking spaces ( ~ ) LaTeX will not break
% a structure at a ~ so this keeps an author's name from being broken across
% two lines.
% use \thanks{} to gain access to the first footnote area
% a separate \thanks must be used for each paragraph as LaTeX2e's \thanks
% was not built to handle multiple paragraphs
%
%
%\IEEEcompsocitemizethanks is a special \thanks that produces the bulleted
% lists the Computer Society journals use for "first footnote" author
% affiliations. Use \IEEEcompsocthanksitem which works much like \item
% for each affiliation group. When not in compsoc mode,
% \IEEEcompsocitemizethanks becomes like \thanks and
% \IEEEcompsocthanksitem becomes a line break with idention. This
% facilitates dual compilation, although admittedly the differences in the
% desired content of \author between the different types of papers makes a
% one-size-fits-all approach a daunting prospect. For instance, compsoc 
% journal papers have the author affiliations above the "Manuscript
% received ..."  text while in non-compsoc journals this is reversed. Sigh.

\author{Tianfu Wu, Yang Lu and Song-Chun Zhu% <-this % stops a space
  \IEEEcompsocitemizethanks{\IEEEcompsocthanksitem T.F. Wu is with the Department of Electrical and Computer Engineering and the Visual Narrative Cluster, North Carolina State University. This work was mainly done when T.F. Wu was research assistant professor at UCLA. \protect\\
    E-mail: tianfu\_wu@ncsu.edu
    \IEEEcompsocthanksitem Y. Lu is with the Department of Statistics, University of California, Los Angeles.\protect\\
E-mail: yanglv@ucla.edu
\IEEEcompsocthanksitem S.-C. Zhu is with the Department of Statistics and Computer Science, University of California, Los Angeles.\protect\\
E-mail: sczhu@stat.ucla.edu
% note need leading \protect in front of \\ to get a newline within \thanks as
% \\ is fragile and will error, could use \hfil\break instead.
}% <-this % stops an unwanted space
\thanks{Manuscript received MM DD, YYYY; revised MM DD, YYYY.}}

% note the % following the last \IEEEmembership and also \thanks - 
% these prevent an unwanted space from occurring between the last author name
% and the end of the author line. i.e., if you had this:
% 
% \author{....lastname \thanks{...} \thanks{...} }
%                     ^------------^------------^----Do not want these spaces!
%
% a space would be appended to the last name and could cause every name on that
% line to be shifted left slightly. This is one of those "LaTeX things". For
% instance, "\textbf{A} \textbf{B}" will typeset as "A B" not "AB". To get
% "AB" then you have to do: "\textbf{A}\textbf{B}"
% \thanks is no different in this regard, so shield the last } of each \thanks
% that ends a line with a % and do not let a space in before the next \thanks.
% Spaces after \IEEEmembership other than the last one are OK (and needed) as
% you are supposed to have spaces between the names. For what it is worth,
% this is a minor point as most people would not even notice if the said evil
% space somehow managed to creep in.

% The paper headers
\markboth{ARXIV Version}%
{Shell \MakeLowercase{\textit{et al.}}: Online Object Tracking with And-Or Graphs}
% The only time the second header will appear is for the odd numbered pages
% after the title page when using the twoside option.
% 
% *** Note that you probably will NOT want to include the author's ***
% *** name in the headers of peer review papers.                   ***
% You can use \ifCLASSOPTIONpeerreview for conditional compilation here if
% you desire.

% The publisher's ID mark at the bottom of the page is less important with
% Computer Society journal papers as those publications place the marks
% outside of the main text columns and, therefore, unlike regular IEEE
% journals, the available text space is not reduced by their presence.
% If you want to put a publisher's ID mark on the page you can do it like
% this:
%\IEEEpubid{0000--0000/00\$00.00~\copyright~2015 IEEE}
% or like this to get the Computer Society new two part style.
%\IEEEpubid{\makebox[\columnwidth]{\hfill 0000--0000/00/\$00.00~\copyright~2015 IEEE}%
%\hspace{\columnsep}\makebox[\columnwidth]{Published by the IEEE Computer Society\hfill}}
% Remember, if you use this you must call \IEEEpubidadjcol in the second
% column for its text to clear the IEEEpubid mark (Computer Society jorunal
% papers don't need this extra clearance.)

% use for special paper notices
%\IEEEspecialpapernotice{(Invited Paper)}

% for Computer Society papers, we must declare the abstract and index terms
% PRIOR to the title within the \IEEEtitleabstractindextext IEEEtran
% command as these need to go into the title area created by \maketitle.
% As a general rule, do not put math, special symbols or citations
% in the abstract or keywords.
\IEEEtitleabstractindextext{%
\begin{abstract}
This paper presents a method, called \textit{AOGTracker}, for simultaneously tracking, learning and parsing (TLP) of unknown objects in video sequences with  a hierarchical and compositional And-Or graph (AOG) representation. %The AOG captures both structural and appearance variations  of a target object in a principled way. 
The TLP method  is formulated in the Bayesian framework with a spatial and a temporal dynamic programming (DP) algorithms inferring object bounding boxes on-the-fly.   
During online learning, the AOG is discriminatively learned using latent SVM~\cite{DPM} to account for appearance (e.g., lighting and partial occlusion) and structural (e.g., different poses and viewpoints) variations of a tracked object, as well as distractors (e.g., similar objects) in  background. Three key issues in online inference and learning are addressed: (i) maintaining  purity of positive and negative examples collected online, (ii) controling model complexity in latent structure learning, and (iii) identifying critical moments to re-learn the structure of AOG based on its intrackability. The intrackability measures uncertainty of an AOG based on its score maps in a frame. 
In experiments, our AOGTracker is tested on two popular tracking benchmarks with the same parameter setting: the TB-100/50/CVPR2013 benchmarks~\cite{trackingBenchmarkPAMI,trackingBenchmark}, and the VOT benchmarks~\cite{VOT}  --- VOT 2013, 2014, 2015 and TIR2015 (thermal imagery tracking). In the former, our AOGTracker outperforms state-of-the-art tracking algorithms including two trackers based on deep convolutional  network ~\cite{cnnTracker,rcnnTracker}. In the latter, our AOGTracker outperforms all other trackers in VOT2013 and is comparable to the state-of-the-art methods in VOT2014, 2015 and TIR2015.
\end{abstract}

% Note that keywords are not normally used for peerreview papers.
}

% make the title area
\maketitle

% To allow for easy dual compilation without having to reenter the
% abstract/keywords data, the \IEEEtitleabstractindextext text will
% not be used in maketitle, but will appear (i.e., to be "transported")
% here as \IEEEdisplaynontitleabstractindextext when the compsoc 
% or transmag modes are not selected <OR> if conference mode is selected 
% - because all conference papers position the abstract like regular
% papers do.
\IEEEdisplaynontitleabstractindextext
% \IEEEdisplaynontitleabstractindextext has no effect when using
% compsoc or transmag under a non-conference mode.

% For peer review papers, you can put extra information on the cover
% page as needed:
% \ifCLASSOPTIONpeerreview
% \begin{center} \bfseries EDICS Category: 3-BBND \end{center}
% \fi
%
% For peerreview papers, this IEEEtran command inserts a page break and
% creates the second title. It will be ignored for other modes.
\IEEEpeerreviewmaketitle

\IEEEraisesectionheading{\section{Introduction}\label{sec:introduction}}
% Computer Society journal (but not conference!) papers do something unusual
% with the very first section heading (almost always called "Introduction").
% They place it ABOVE the main text! IEEEtran.cls does not automatically do
% this for you, but you can achieve this effect with the provided
% \IEEEraisesectionheading{} command. Note the need to keep any \label that
% is to refer to the section immediately after \section in the above as
% \IEEEraisesectionheading puts \section within a raised box.

% The very first letter is a 2 line initial drop letter followed
% by the rest of the first word in caps (small caps for compsoc).
% 
% form to use if the first word consists of a single letter:
% \IEEEPARstart{A}{demo} file is ....
% 
% form to use if you need the single drop letter followed by
% normal text (unknown if ever used by the IEEE):
% \IEEEPARstart{A}{}demo file is ....
% 
% Some journals put the first two words in caps:
% \IEEEPARstart{T}{his demo} file is ....
% 
% Here we have the typical use of a "T" for an initial drop letter
% and "HIS" in caps to complete the first word.
%\vspace{-1mm}
\subsection{Motivation and Objective}
\begin{figure}
	\centering
	\includegraphics[width = 0.5\textwidth]{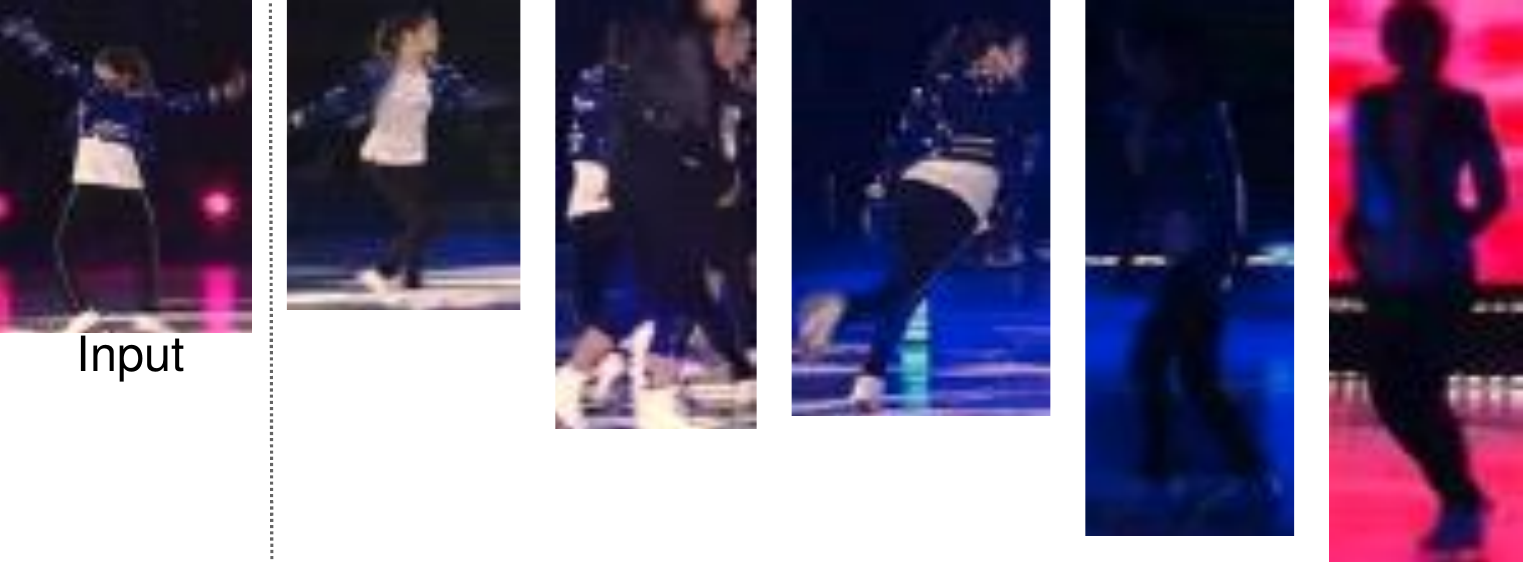}
	\caption{\label{fig:example} Illustration of some typical issues in online object tracking using the ``skating1" video in the benchmark~\cite{trackingBenchmarkPAMI}. Starting from the object specified in the first frame, a tracker needs to handle many variations in subsequent frames which include \textit{illuminative variation, scale variation, occlusion, deformation, fast motion, in-plane and out-of-plane rotation, background clutter}, etc. %To address these issues, the object representation must be sufficiently  expressive and the tracking algorithm must be robust and efficient.
	}
	%\vspace{-5mm} 
\end{figure}

\IEEEPARstart{O}{nline} object tracking is an innate capability in human and animal vision for learning visual concepts~\cite{Concepts}, and is an important task in computer vision. Given the state of an unknown object (e.g., its bounding box) in the first frame of a video, the task is to infer hidden states of the object in subsequent frames. 
Online object tracking, especially long-term tracking, is a difficult problem. It needs to handle variations of a tracked object, including  appearance and structural variations, scale changes, occlusions (partial or complete), etc. It also needs to tackle complexity of the scene, including camera motion, background clutter, distractors, illumination changes, frame cropping, etc. Fig.~\ref{fig:example} illustrates some typical issues in online object tracking.
In recent literature, object tracking has received much attention due to practical applications in video surveillance, activity and event prediction, human-computer interactions and traffic monitoring. 

\begin{SCfigure*}
  %\centering
  \caption{Overview of our AOGTracker. (a) Illustration of the  tracking, learning and parsing (TLP) framework. It consists of four components. (b) Examples of capturing structural and appearance variations of a tracked object by a series of object configurations inferred on-the-fly over key frames \#1, \#173, \#282, etc.  (c) Illustration of an object AOG, a parse tree and an object configuration in frame \#282. A parse tree is an instantiation of an AOG. A configuration is a layout of latent parts represented by terminal-nodes in a parse tree. An object AOG preserves ambiguities by capturing multiple parse trees. }\label{fig:overview} %\vspace{-4mm}
  \includegraphics[width=0.7\textwidth] {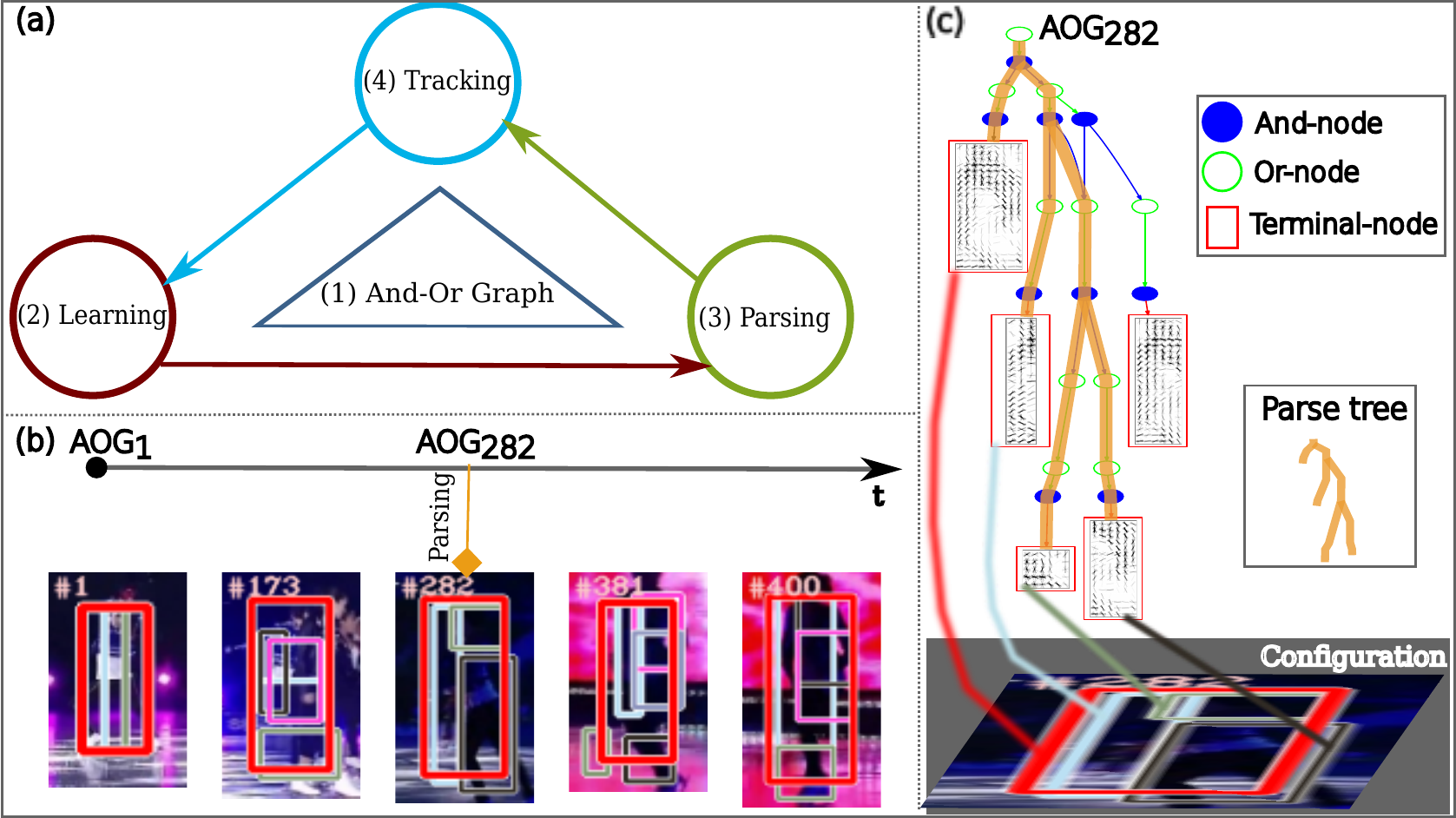}
\end{SCfigure*}

This paper presents an integrated framework for online tracking, learning and parsing (TLP) of unknown objects with a unified  representation. We focus on settings in which object state is represented by bounding box, without using pre-trained models. We address five issues associated with online object tracking in the following. 

\textit{Issue I: Expressive representation accounting for structural and appearance variations of unknown objects in tracking.} 
%part-based models provide an elegant framework~\cite{PictorialStructure}.
We are interested in hierarchical and compositional object models. 
Such models have shown promising performance in object detection~\cite{DPM,DisAOT,WLLSVM,pffGrammar,AOG} and object recognition~\cite{POP}. 
A popular modeling scheme represents object categories by mixtures of deformable part-based models (DPMs)~\cite{DPM}. The number of mixture components is usually predefined and the part configuration of each component is fixed after initialization or directly based on strong supervision. In online tracking, since a tracker can only access the ground-truth object state in the first frame, 
%it does not have enough information at the beginning to utilize popular modeling schema used in object detection. For example, in the mixture of deformable part-based models (DPMs)~\cite{DPM}, the number of mixture components is predefined and the part configuration of each component is fixed after initialization or directly based on strong supervision.  
it is not suitable for it to ``make decisions" on the number of mixture components and part configurations, and it does not have enough data to learn. It's desirable to have an object representation which has expressive power to represent a large number of part configurations, and can facilitate computationally effective inference and learning. 
We quantize the space of part configurations recursively in a principled way with a hierarchical and compositional And-Or graph (AOG) representation~\cite{DisAOT,AOG}. We learn and update the most discriminative part configurations online by pruning the quantized space based on part discriminability.

\textit{Issue II: Computing joint optimal solutions.} Online object tracking is usually posed as a maximum a posterior (MAP) problem using first order hidden Markov models (HMMs) \cite{Survey,trackingBenchmarkPAMI,HMM}. The likelihood or observation density is temporally inhomogeneous due to online updating of object models. Typically, the objective is to infer the most likely hidden state of a tracked object in a frame by maximizing a Bayesian marginal posterior probability given all the data observed so far. The maximization is based on either  particle filtering~\cite{particleFiltering} or dense sampling such as the tracking-by-detection methods ~\cite{TrackbyDet,TLD,SelfpacedTracking}. In most prior approaches (e.g., the 29 trackers evaluated in the TB-100 benchmark~\cite{trackingBenchmarkPAMI}), no feedback inspection is applied to the history of inferred trajectory. We utilize tracking-by-parsing with hierarchical models in inference. %We allow feedback inspection by maximizing a Bayesian joint posterior probability of trajectory given all the observation. %Thus, a tracker can trace back the trajectory to potentially improve accuracy at each step as more evidence has been observed. By doing so, we simultaneously address another key issue in online learning (Issue III).
By computing joint optimal solutions, we can not only improve prediction accuracy in a new frame by integrating past estimated trajectory, but also potentially correct errors in past estimated trajectory. Furthermore, we simultaneously address another key issue in online learning (Issue III).

 \textit{Issue III: Maintaining the purity of a training dataset.} The dataset consists of a set of positive examples computed based on the current trajectory, and a set of negative examples mined from outside the current trajectory. 
In the dataset, we can only guarantee that the positives and the negatives in the first frame are true positives and true negatives respectively. 
A tracker needs to carefully choose frames from which it can learn to avoid model drifting (i.e., self-paced learning). Most prior approaches do not address this issue since they focus on marginally optimal solutions with which object models are updated, except for the P-N learning in TLD~\cite{TLD} and the self-paced learning for tracking~\cite{SelfpacedTracking}. %Since we allow feedback inspection in tracking, we can correct previous errors in the training dataset. 
Since we compute joint optimal solutions in online tracking, we can maintain the purity of an online collected training dataset in a better way. 

\textit{Issue IV: Failure-aware online learning of object models.} 
In online learning, we mostly update model parameters incrementally after inference in a frame. Theoretically speaking, after an initial object model is learned in the first frame, model drifting is inevitable in general setting. Thus, in addition to maintaining the purity of a training dataset, it is also  important that we can identify critical moments (caused by different structural and appearance variations) automatically. At those moments, a tracker needs to re-learn both the structure and the parameters of object model using the current whole training dataset. We address this issue by computing uncertainty of an object model in a frame based on its response maps. 

\textit{Issue V: Computational efficiency by dynamic search strategy.} Most tracking-by-detection methods run detection in the whole frame since they usually use relatively simple models such as a single object template. With hierarchical models in tracking and sophisticated online inference and updating strategies, the computational complexity is high. To speed up tracking, we need to utilize a dynamic search strategy. This strategy must take into account the trade-off between generating a conservative proposal state space for efficiency and allowing an exhaustive search for accuracy (e.g., to handle the situation where the object is completely occluded for a while or moves out the camera view and then reappears). We address this issue by adopting a simple search cascade with which we run detection in the whole frame only when local search has  failed.

Our TLP method obtains state-of-the-art performance on one popular tracking benchmark~\cite{trackingBenchmarkPAMI}.  We give a brief overview of our method in the next subsection.

%\vspace{-2mm}
\subsection{Method Overview}\label{sec:overview}
As illustrated in Fig.\ref{fig:overview} (a), the TLP method consists of four components. We  introduce them briefly as follows.  

(1) \textit{An AOG quantizing the space of part configurations.}
Given the bounding box of an object in the first frame, we assume object parts are also of rectangular shapes. We first divide it evenly into a small cell-based grid (e.g., $3\times 3$) and a cell defines the smallest part. We then enumerate all possible parts with different aspect ratios and different sizes which can be placed inside the grid. All the enumerated parts are organized into a hierarchical and compositional AOG. Each part is represented by a terminal-node. Two types of nonterminal nodes as compositional rules: an And-node represents the decomposition of a large part into two smaller ones, and an Or-node represents alternative ways of decompositions through different horizontal or vertical binary splits. %We elaborate the details in Section~\ref{sec:AOG}.   
We call it \textbf{the full structure AOG}\footnote{By “full structure”, it means all the possible compositions on top of the  grid with binary composition being used for And-nodes}. It is capable of exploring a large number of latent part configurations (see some examples in Fig.~\ref{fig:overview} (b)), meanwhile it makes the problem of online model learning feasible.

(2) \textit{Learning object AOGs.}  
An object AOG is a subgraph learned from the full structure AOG (see Fig.~\ref{fig:overview} (c)~\footnote{We note that there are some Or-nodes in the object AOGs which have only one child node since they are subgraphs of the full structure AOG and we keep their original structures.}). Learning an object AOG consists of two steps: (i) The initial object AOG are learned by pruning branches of Or-nodes in the full structure AOG based on discriminative power, following breadth-first search (BFS) order. The discriminative power of a node is measured based on its training error rate. 
We keep multiple branches for each encountered Or-node to preserve ambiguities, whose training error rates are not bigger than the minimum one by a small positive value. (ii) We retrain the initial object AOG using latent SVM (LSVM) as it was done in learning the DPMs~\cite{DPM}. LSVM utilizes positive re-labeling (i.e., inferring the best configuration for each positive example) and hard negative mining. To further control the model complexity, we prune the initial object AOG through majority voting of latent assignments in positive re-labeling.   

(3) \textit{A spatial dynamic programming (DP) algorithm for computing all the proposals in a frame with the current object AOG}.
 Thanks to the DAG structure of the object AOG, a DP parsing algorithm is utilized to compute the matching scores and the optimal parse trees of all sliding windows inside the search region in a frame. A parse tree is an instantiation of the object AOG which selects the best child for each encountered Or-node according to matching score. A configuration is obtained by collapsing a parse tree onto the image domain, capturing layout of latent parts of a tracked object in a frame.
 %The search region can be either some local neighborhood around the location and scale predicted based on the previous bounding box or the whole image domain when the local search failed. 
 
(4) \textit{A temporal DP algorithm for inferring the most likely trajectory}. We maintain a DP table memorizing the candidate object states computed by the spatial DP in the past frames. Then, based on the first-order HMM assumption, a temporal DP algorithm is used to find the optimal solution for the past frames jointly with pair-wise motion constraints (i.e., the Viterbi path~\cite{HMM}). The joint solution can help correct potential tracking errors (i.e., false negatives and false positives collected online) by leveraging more spatial and temporal information. This is similar in spirit to methods of keeping N-best maximal decoder for part models~\cite{nbestPartModels} and maintaining diverse M-best solutions in MRF~\cite{mbestMRF}.

%%%%%%%%%%%%%%%%%%%%%%%%%%%%%%%%%%%%%%%%%%%%%%%%%%%%%%%%%%%%%%%%%%%%%%%%%
%\vspace{-2mm}
\section{Related Work}
In the literature of object tracking, either single object tracking or multiple-object tracking,  there are often two settings. 

\textit{Offline visual tracking}~\cite{optimalGreedyTracking1,optimalGreedyTracking,KShortestTracking,minCostFlow}. These methods assume the whole video sequence has been recorded, and consist of two steps. i) It first computes object proposals in all frames using some pre-trained detectors (e.g., the DPMs~\cite{DPM}) and then form ``tracklets" in consecutive frames. ii) It seeks the optimal object trajectory (or trajectories for multiple objects) by solving an optimization problem (e.g., the K-shortest path or min-cost flow formulation) for the data association. Most work assumed first-order HMMs in the formulation. Recently, Hong and Han~\cite{TreeStructuredTracker} proposed an offline single object tracking method by sampling tree-structured graphical models which exploit the underlying intrinsic structure of input video in an orderless tracking~\cite{OrderlessTracking}.

\textit{Online visual tracking} for streaming videos. It starts tracking after the state of an object is specified in certain frame. 
In the literature, particle filtering~\cite{particleFiltering} has been widely adopted, which approximately represents the posterior probability in a non-parametric form by maintaining a set of particles (i.e., weighted candidates). In practice, particle filtering does not perform well in high-dimensional state spaces.
More recently, tracking-by-detection methods \cite{TrackbyDet,TLD} have become popular which learn and update object models online and encode the posterior probability using dense sampling through sliding-window based detection on-the-fly. Thus, object tracking is treated as instance-based object detection. To leverage the recent advance in object detection, object tracking research has made progress by incorporating discriminatively trained part-based models~\cite{DPM,DisAOT,partBasedTracking} (or more generally grammar models \cite{AOG, pffGrammar, WLLSVM}). 
Most popular methods also assume first-order HMMs except for the recently proposed online graph-based tracker~\cite{OnlineGraphbasedTracking}. There are four streams in the literature of online visual tracking: 
\begin{itemize} %[leftmargin=*]
	\item [i)] Appearance modeling of the whole object, such as incremental learning~\cite{IVT}, kernel-based~\cite{KMS}, particle filtering~\cite{particleFiltering}, sparse coding~\cite{sparseCoding} and 3D-DCT representation~\cite{3DDCT}; More recently, Convolutional neural networks are utilized in improving tracking performance~\cite{rcnnTracker,cnnTracker,ConvNetTracking}, which are usually pre-trained on some large scale image datasets such as the ImageNet~\cite{imagenet} or on video sequences in a benchmark with the testing one excluded.
	\item [ii)]  Appearance modeling of objects with parts, such as patch-based~\cite{patchBased}, coupled 2-layer models~\cite{coupled2layer} and adaptive sparse appearance~\cite{adaptive}. The major limitation of appearance modeling of a tracked object is the lack of background models, especially in preventing model drift from distracotrs (e.g., players in sport games). Addressing this issue leads to discriminant tracking.
	\item [iii)] Tracking by discrimination using a single classifier, such as  support vector tracking~\cite{sv}, multiple instance learning~\cite{MIL}, STRUCK~\cite{STRUCK}, circulant structure-based kernel method~\cite{circulant}, and  discriminant saliency based tracking \cite{saliency};
	\item [iv)] Tracking by part-based discriminative models, such as  online extensions of DPMs~\cite{onlineDPM}, and  structure preserving tracking method~\cite{structurePreserving,partBasedTracking}.
\end{itemize}

Our method belongs to the fourth stream of online visual tracking. Unlike  predefined or fixed part configurations with star-model structure used in previous work, our method learns both structure and appearance of object AOGs online,  which is, to our knowledge, the first method to address the problem of online explicit structure learning in tracking. 
The advantage of introducing AOG representation are three-fold.
\begin{itemize} %[leftmargin=*]
\item [i)] \textit{More representational power}: Unlike TLD~\cite{TLD} and many other methods (e.g., \cite{SelfpacedTracking}) which model an object as a single template or a mixture of templates and thus do not perform well in tracking objects with large structural and appearance variations, an AOG represents an object in a hierarchical and compositional graph expressing a large number of latent part configurations. 
\item [ii)] \textit{More robust tracking and online learning strategies}: While the whole object has large variations or might be partially occluded from time to time during tracking, some other parts remain stable and are less likely to be occluded. Some of the parts can be learned to robustly track the object, which can also improve accuracy of appearance adaptation of terminal-nodes. 
This idea is similar in spirit to finding good features to track objects \cite{goodFeatTrack}, and we find good part configurations online for both tracking and learning. 
\item [iii)]  \textit{Fine-grained tracking results}: In addition to predicting bounding boxes of a tracked object, outputs of our AOGTracker (i.e., the parse trees) have more information which are  potentially useful for other modules beyond tracking such as activity or event prediction. 
\end{itemize}

Our preliminary work has been published in \cite{AOGTracker} and the method for constructing full structure AOG was published in \cite{DisAOT}. This paper extends them by: (i) adding more experimental results with state-of-the-art performance obtained and full source code released;
(ii) elaborating details substantially in deriving the formulation of inference and learning algorithms; and (iii) adding more analyses on different aspects of our method.
This paper makes three contributions to the online object tracking problem:
\begin{itemize} %[leftmargin=*]
\item[i)] It presents a tracking-learning-parsing (TLP) framework which can learn and track objects AOGs. 

\item[ii)] It presents a spatial and a temporal DP algorithms for tracking-by-parsing with AOGs and outputs fine-grained tracking results using parse trees. 

\item[iii)] It outperforms the state-of-the-art tracking methods in a recent public benchmark, TB-100 ~\cite{trackingBenchmarkPAMI}, and obtains comparable performance on a series of VOT benchmarks \cite{VOT}. 
\end{itemize}

\noindent \textbf{Paper Organization.} The remainder of this paper is organized as follows. Section \ref{sec:formulation} presents the formulation of our TLP framework under the Bayesian framework.  Section \ref{sec:trackAOG} gives the details of spatial-temporal DP algorithm. Section \ref{sec:online} presents the online learning algorithm using the latent SVM method. Section \ref{sec:exp} shows the experimental results and analyses. Section \ref{sec:conclusion} concludes this paper and discusses issues and future work. 

%\vspace{-2mm}
\section{Problem Formulation}\label{sec:formulation}
 %We also briefly introduce the construction of the full structure AOG proposed in~\cite{DisAOT} to be self-contained. 

%\vspace{-2mm}
\subsection{Formulation of Online Object Tracking}\label{sec:tracking}
In this section,  we first derive a generic formulation from generative perspective in the Bayesian framework, and then derive the discriminative counterpart.
%We assume a first-order HMM in tracking as commonly adopted in the tracking field.  

%\vspace{-2mm} 
\subsubsection{Tracking with HMM}
Let $\Lambda$ denote the image lattice on which video frames are defined. 
Denote a sequence of video frames within time range $[1, T]$ by, 
\begin{equation}
I_{1:T} = \{I_1, \cdots, I_T\}. 
\end{equation}
Denote by $B_t$ the bounding box of a target object in $I_t$. In online object tracking, $B_1$ is given and $B_t$'s are inferred by a tracker ($t\in [2, T]$). 
With first-order HMM, we have,
\begin{align}
\text{The prior model:}\qquad \qquad \,\,\, B_1 &\sim p(B_1)\, ,\\
\text{The motion model:}\qquad B_t|B_{t-1} &\sim p(B_t|B_{t-1})\, ,\\
\text{The likelihood:}\qquad \quad\,\, I_t|B_t &\sim p(I_t|B_t).
\end{align}

Then, the prediction model is defined by,
\begin{equation}
p(B_t|I_{1:t-1})=\int_{\Omega_{B_{t-1}}} p(B_t|B_{t-1})p(B_{t-1}|I_{1:t-1})d B_{t-1},
\end{equation}
where $\Omega_{B_{t-1}}$ is the candidate space of $B_{t-1}$, and the updating model is defined by,
 \begin{equation}
p(B_t|I_{1:t})=p(I_t|B_t)p(B_t|I_{1:t-1})/p(I_t|I_{1:t-1}),
 \end{equation}   
 which is a marginal posterior probability. The tracking result, the best bounding box $B_t^*$, is computed by,
 \begin{equation}
 B_t^* = \arg \max_{B_t \in \Omega_{B_t}} p(B_t|I_{1:t}),
 \end{equation}
 which is usually solved using particle filtering~\cite{particleFiltering}  in practice.
 
To allow feedback inspection of the history of a trajectory, we seek to maximize a joint posterior probability, 
\begin{align}
\nonumber p(B_{1:t}|I_{1:t}) &= p(B_{1:t-1}|I_{1:t-1}) {{p(B_t|B_{t-1})p(I_t|B_t)}\over {p(I_t|I_{1:t-1})}}\\
&=p(B_{1}|I_{1}) \prod_{i=2}^t {{{p(B_i|B_{i-1})p(I_i|B_i)}\over {p(I_i|I_{1:i-1})}}}. \label{eq:joint}
\end{align}

By taking the logarithm of both sides of Eqn.(\ref{eq:joint}), we have,
\begin{align}
\nonumber B_{1:t}^* = &\arg \max_{B_{1:t}} \log p(B_{1:t}|I_{1:t})\\
\nonumber  = &\arg \max_{B_{1:t}}\{ \log p(B_1) + \log p(I_1|B_1)+\\
&\qquad \sum_{i=2}^t [\log p(B_i|B_{i-1}) + \log p(I_i|B_i)]\}. \label{eqn:dp} 
\end{align}
where the image data term $p(I_1)$ and $\sum_{i=2}^t p(I_i|I_{1:i-1})$ are not included in the maximization as they are treated as constant terms. 

Since we have ground-truth for $B_1$, $p(I_1|B_1)$ can also be treated as known after the object model is learned based on $B_1$. Then, Eqn.(\ref{eqn:dp}) can be reproduced as,
\begin{align}
 B_{2:t}^* = &\arg \max_{B_{2:t}} \log p(B_{2:t}|I_{1:t},B_1)  \label{eq:dp1} \\
\nonumber= &\arg \max_{B_{2:t}}\{\sum_{i=2}^t [\log p(B_i|B_{i-1}) + \log p(I_i|B_i)]\}.
\end{align}

%\vspace{-2mm}
\subsubsection{Tracking as Energy Minimization over Trajectories}\label{sec:discriminativeDerivation}
To derive the discriminative formulation of Eqn.(\ref{eq:dp1}), we  show that only the log-likelihood ratio matters in computing  $\log p(I_i|B_i)$ in Eqn.(\ref{eq:dp1}) with very mild assumptions. 

Let $\Lambda_{B_i}$ be the image domain occupied by a tracked object, and $\overline{\Lambda_{B_i}}$ the remaining domain (i.e., $\overline{\Lambda_{B_i}}\cup \Lambda_{B_i}=\Lambda$ and $\overline{\Lambda_{B_i}}\cap \Lambda_{B_i}=\emptyset$) in a frame  $I_i$. With the independence assumption between $I_{\Lambda_{B_i}}$ and $I_{\overline{\Lambda_{B_i}}}$ given $B_i$, we have,
\begin{align}
\nonumber p(I_i|B_i) &= p(I_{\Lambda_{B_i}}, I_{\overline{\Lambda_{B_i}}}|B_i)=p(I_{\Lambda_{B_i}}|B_i) p(I_{\overline{\Lambda_{B_i}}}|B_i)\\
&=p(I_{\Lambda_{B_i}}|B_i) q(I_{\overline{\Lambda_{B_i}}})= q(I_{\Lambda}) {{p(I_{\Lambda_{B_i}}|B_i)}\over q(I_{\Lambda_{B_i}})},
\end{align}
where $q(I_{\Lambda})$ is the probability model of background scene and we have $q(I_{\overline{\Lambda_{B_i}}})=p(I_{\overline{\Lambda_{B_i}}}|B_i)$ w.r.t. context-free assumption. So, $q(I_{\Lambda})$  does not need to be specified explicitly and can be omitted in the maximization. This derivation gives an alternative explanation for  discriminant tracking \textit{v.s.} tracking by generative appearance modeling of an object~\cite{surveyAppInTracking}. 

Based on Eqn.(\ref{eq:dp1}), we define an energy function by,
\begin{equation}
\mathcal{E}(B_{2:t}|I_{1:t},B_0)\propto -\log p(B_{2:t}|I_{1:t}, B_1). 
\end{equation}  
And, we do not compute $\log p(I_i|B_i)$ in the probabilistic way, instead we compute matching score defined by,  
%Denote the matching score by,
\begin{align}
\text{Score}(I_i|B_i)&=\log {{p(I_{\Lambda_{B_i}}|B_i)}\over q(I_{\Lambda_{B_i}})}\\
\nonumber & = \log p(I_i|B_i) - \log q(I_{\Lambda}).
\end{align}
which we can apply discriminative learning methods. 

Also, denote the motion cost by,
\begin{align}
\text{Cost}(B_i|B_{i-1}) = -\log p(B_i|B_{i-1}).
\end{align}
We use a thresholded motion model in experiments: the cost is $0$ if the transition is accepted based on the median flow \cite{TLD} (which is a forward-backward extension of the Lucas-Kanade optimal flow~\cite{LK}) and $+\infty$ otherwise. A similar method was explored  in~\cite{SelfpacedTracking}.

So, we can re-write Eqn.(\ref{eq:dp1}) in the minimization form, 
\begin{align}
B_{2:t}^* = &\arg \min_{B_{2:t}}\, \mathcal{E}(B_{2:t}|I_{1:t},B_1) \label{eq:dp2} \\
\nonumber = &\arg \min_{B_{2:t}}\, \{\sum_{i=2}^t [\text{Cost}(B_i|B_{i-1}) - \text{Score}(I_i|B_i)]\}. \label{eqn:dp2}
\end{align}

In our TLP framework, we compute $\text{Score}(I_i|B_i)$ in Eqn.(~\ref{eq:dp2}) with an object AOG. So, we interpret a sliding window by the optimal parse tree inferred from object AOG. We treat parts as latent variables which are modeled to leverage more information for inferring object bounding box. We note that we do not track parts explicitly in this paper. 

\begin{figure}
	\centering
	\includegraphics[width = 0.5\textwidth]{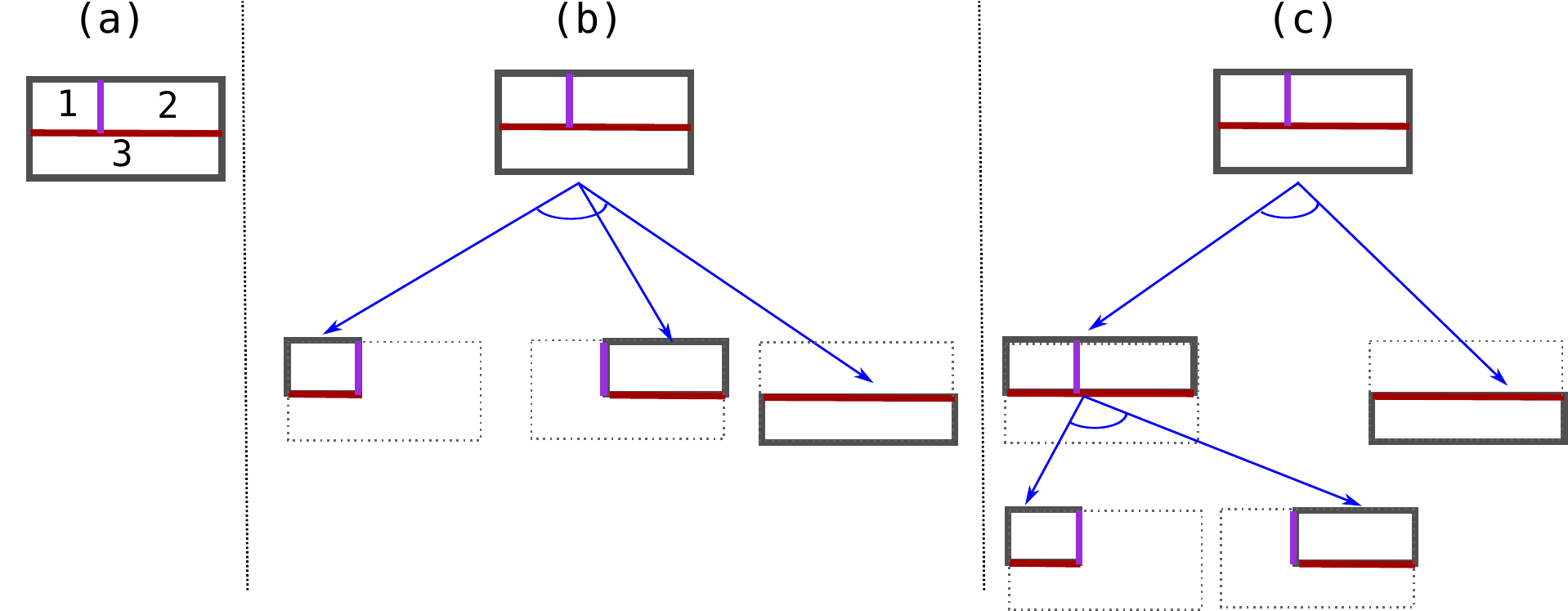}
	\caption{ We assume parts are of rectangular shapes. (a) shows a configuration with 3 parts. Two different, yet equivalent, decomposition rules in representing a configuration are shown in (b) for decomposition with  branching factor equal to the number of parts (i.e., a flat structure), and in (c) for a hierarchical decomposition with branching factor being set to 2 at all levels. 
	}
	\label{fig:binarySplitting}
	%\vspace{-3mm}
\end{figure}

\begin{figure}
	\centering
	\includegraphics[width = 0.5\textwidth]{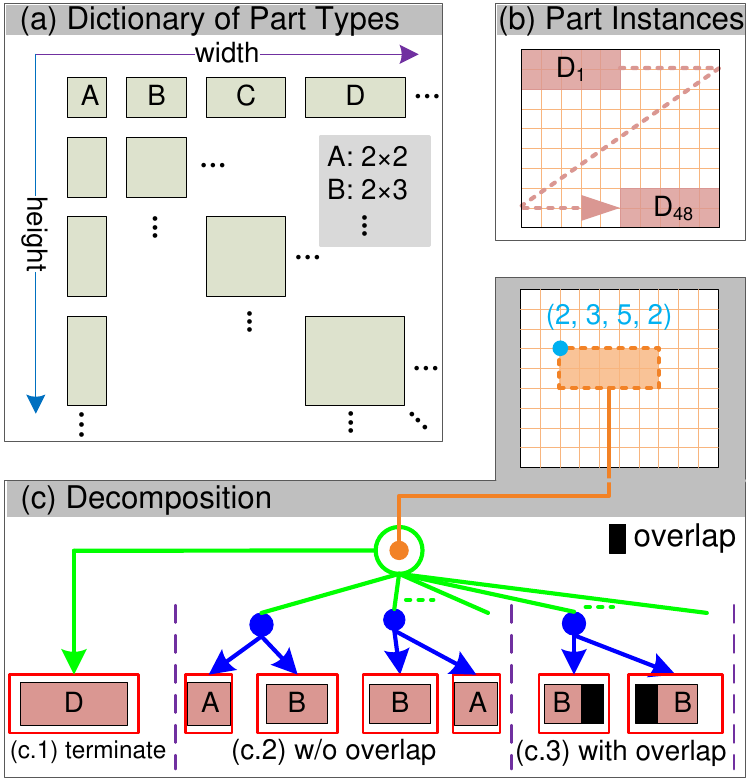}
	\caption{Illustration of (a) the dictionary of part types, and (b) part instances generated by placing a part type in a grid. Given part instances, (c) shows how a sub-grid is decomposed in different ways. We allow overlap between child nodes (see (c.3)).}
	\label{fig:parts}
%	\vspace{-4mm}
\end{figure}

\begin{figure*}
	\centering
	\includegraphics[width = 1.0\textwidth]{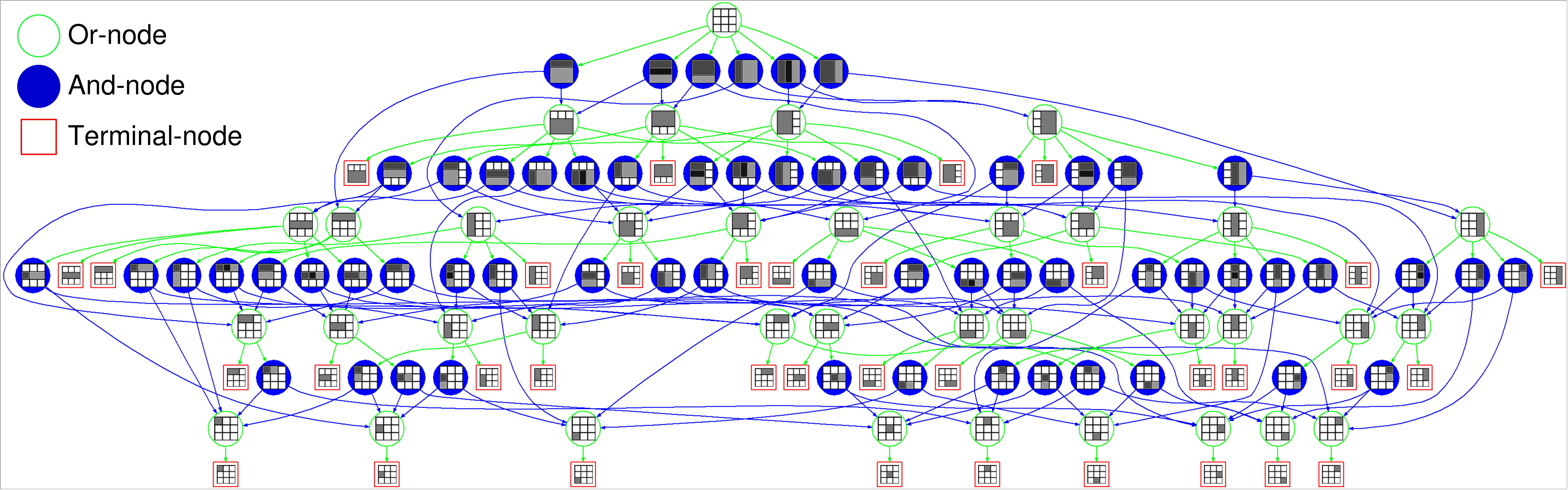}
	\caption{Illustration of full structure And-Or Graph (AOG) representing the space of part configurations. It is of directed acyclic graph (DAG) structure. For clarity, we show a toy example constructed for a $3\times 3$ grid. The AOG can generate all possible part configurations (the number is often huge for typical grid sizes, see Table.\ref{Table:numConfig}), while allowing efficient exploration with a DP algorithm due to the DAG structure. See text for details. (Best viewed in color and with magnification)}
	\label{fig:AOG}
	%\vspace{-3mm}
\end{figure*}

%\vspace{-2mm}
\subsection{Quantizing the Space of Part Configurations}\label{sec:AOG}
In this section, we first present the construction of a full structure AOG which quantizes the space of part configurations. We then introduce notations in defining an AOG. 

\textit{Part configurations.} For an input bounding box, a part configuration is defined by a partition with different number of parts of different shapes (see Fig.~\ref{fig:binarySplitting} (a)). Two natural questions arise: (i) How many part configurations (i.e., the space) can be defined in a bounding box? (ii) How to organize them into a compact representation? Without posing some structural constraints, it is a combinatorial problem. 

We assume rectangular shapes are used for parts. Then, a configuration can be treated as a tiling of input bounding box using either horizontal or vertical cuts. We utilize binary splitting rule only  in decomposition (see Fig.~\ref{fig:binarySplitting} (b) and (c)). 
With these two constraints, we represent all possible part configurations by a hierarchical and compositional AOG constructed in the following. 
 
Given a bounding box, we first divide it evenly into a cell-based grid (e.g., $9\times 10$ grid in the right of Fig.~\ref{fig:parts}). Then, in the grid, we define a dictionary of part types and enumerate all instances for all part types. 

\textit{A dictionary of part types.} A part type is defined by its width and height.  Starting from some minimal size (such as $2\times 2$ cells), we enumerate all possible part types with different aspect ratios and sizes which fit the grid (see $A, B, C, D$ in Fig.\ref{fig:parts} (a)). 

\textit{Part instances.} An instance of a part type is obtained by placing the part type at a position. Thus, a part instance is defined by a ``sliding window" in the grid. Fig.\ref{fig:parts} (b) shows an example of placing part type $D$ ($2\times 5$ cells) in a $9\times 10$ grid with $48$ instances in total.

To represent part configurations compactly, we exploit the compositional relationships between enumerated part instances.   

 \textbf{The full structure AOG.} For any sub-grid indexed by the left-top position, width and height (e.g., $(2,3,5,2)$ in the right-middle of Fig.\ref{fig:parts} (c)), we can either terminate it directly to the corresponding part instance (Fig.\ref{fig:parts} (c.1)), or decompose it into two smaller sub-grids using \textit{either horizontal or vertical binary splits}. Depending on the side length, we may have multiple valid splits along both directions (Fig.\ref{fig:parts} (c.2)). When splitting either side we allow overlaps between the two sub-grids up to some ratio (Fig.\ref{fig:parts} (c.3)). Then, we represent the sub-grid as an Or-node, which has a set of child nodes including a terminal-node (i.e. the part instance directly terminated from it), and a number of And-nodes (each of which represents a valid decomposition). This procedure is applied recursively for all child sub-grids. Starting from the whole grid and using BFS order, we construct a full structure AOG, all summarized in Algorithm~\ref{alg:AOG} (see Fig.~\ref{fig:AOG} for an example).   Table.~\ref{Table:numConfig} lists the number of part configurations for three cases from which we can see that full structure AOGs cover a large number of part configurations using a relatively small set of part instances.

\begin{algorithm}
 \SetAlgoLined
\KwIn{Image grid $\Lambda$ with $W\times H$ cells; Minimal size of a part type $(w_0, h_0)$; Maximal overlap ratio $r$ between two sub-grids.}
\KwOut{The And-Or graph $\mathcal{G}=<V, E>$ (see Fig.\ref{fig:AOG})}
 Initialization: Create an Or-node $O_{\Lambda}$ for the grid $\Lambda$, $V=\{O_{\Lambda}\}, E=\emptyset$, BFSqueue$=\{O_{\Lambda}\}$\;
 \While{BFSqueue is not empty}{
  Pop a node $v$ from the BFSqueue\; 
  \uIf{$v$ is an Or-node}{
  	i) Add a terminal-node $t$ (i.e. the part instance) $V=V\cup\{t\},\, E=E\cup\{<v, t>\}$\;
  	ii) Create And-nodes $A_i$ for all valid cuts\;
  	$E=E\cup\{<v, A_i>\}$\;
  	\If{$A_i\notin V$} {
  		$V=V\cup\{ A_i \}$\; 
  		Push $A_i$ to the back of BFSqueue\;
  	}
  }
  \ElseIf{$v$ is an And-node} {
  	Create two Or-nodes $O_i$ for the two sub-grids\;
  	$E=E\cup\{<v, O_i>\}$\;
  	\If{$O_i\notin V$} {
  		$V=V\cup\{ O_i \}$\;
  		Push $O_i$ to the back of BFSqueue\;
  	}
  }  
 }
  \caption{Constructing the grid AOG using BFS}\label{alg:AOG} 
\end{algorithm}

%\textit{Remark:} The construction of the full structure AOG can be treated as applying the Cocke--Younger--Kasami (CYK) parsing algorithm in NLP to a 2D grid with the binary splitting rule: we explore every possible partition (binary in our case) for every possible sub-grid in the input  grid. 

%\vspace{-2mm}
%\subsubsection{The Object AOG for Tracking}
%An object AOG is a subgraph of the full structure AOG with the same root Or-node, which is denoted by,
We denote an AOG by, 
\begin{equation}
\mathcal{G} = (V_{And}, V_{Or}, V_{T}, E, \Theta)
\end{equation} 
where $V_{And}, V_{Or}$ and $V_T$ represent a set of And-nodes, Or-nodes and terminal-nodes respectively, $E$ a set of edges and $\Theta$ a set of parameters (to be defined in Section~\ref{sec:scoreFunction}).  We have,
\begin{itemize}[leftmargin=*]
\item[i)] \textit{The object/root Or-node (plotted by green circles)}, which represents alternative object configurations;

\item[ii)] \textit{A set of And-nodes (solid blue circles)}, each of which represents the rule of decomposing a complex  structure (e.g., a walking person or  a running basketball player) into simpler ones;

\item[iii)] \textit{A set of part Or-nodes}, which handle local variations and configurations in a recursive way; 

\item[iv)] \textit{A set of terminal-nodes (red rectangles)}, which link an  object and its parts to image data (i.e., grounding symbols) to account for appearance variations and occlusions (e.g.,  head-shoulder of a walking person before and after opening a sun umbrella). 
\end{itemize}

\begin{table}
	 	\begin{tabular}{ | l | l | l | l | l |   }
	 		\hline 
	 		Grid & primitive part & $\#$Configuration & $\#$T-node  & $\#$And-node\\ \hline
	 		$3\times 3$ & $1\times 1$ & 319 & 35 &  48 \\ \hline
	 		$5\times 5$ & $1\times 1$ & 76,879,359 & 224 & 600 \\ \hline
	 		$10\times 12$ & $2\times 2$ & 3.8936e+009 & 1409 & 5209 \\ \hline
	 	\end{tabular}
	 	\caption{The number of part configurations generated from our AOG without considering overlapped compositions.} \label{Table:numConfig}
	 	%\vspace{-3mm}
	 \end{table}

\textit{An object AOG} is a subgraph of a full structure AOG with the same root Or-node. For notational simplicity, we also denote by $\mathcal{G}$ an object AOG. So, we will write $\text{Score}(I_i|B_i;\mathcal{G})$ in Eqn.(~\ref{eq:dp2}) with $\mathcal{G}$ added. 

\textit{A parse tree} is an instantiation of an object AOG with the best child node (w.r.t. matching scores) selected for each encountered Or-node. All the terminal-nodes in a parse tree represents a part configuration when collapsed to image domain. 

We note that an object AOG contains multiple parse trees to preserve ambiguities in interpreting a tracked object (see examples in Fig.~\ref{fig:overview} (c) and Fig.~\ref{fig:initAOG}).

%\vspace{-2mm}
\section{Tracking-by-Parsing with Object AOGs}\label{sec:trackAOG}
In this section, we present details of inference with object AOGs. We first define scoring functions of nodes in an AOG. Then, we present a spatial DP algorithm for computing $\text{Score}(I_i|B_i;\mathcal{G})$, and a temporal DP algorithm for inferring the trajectory $B_{2:t}^*$ in Eqn.(\ref{eq:dp2}). %Then, we introduce the online learning of object AOGs in Section~\ref{sec:online} since inference is a prerequisite for learning.

%\vspace{-2mm}
\subsection{Scoring Functions of Nodes in an AOG}\label{sec:scoreFunction}
%We first define scoring functions of an object AOG in computing the matching scores.

Let $\mathbb{F}$ be the feature pyramid computed for either the local ROI or the whole image $I_t$, and $\mathbf{\Lambda}$ the position space of  pyramid $\mathbb{F}$. Let $p=(l, x, y)\in \mathbf{\Lambda}$ specify a position $(x, y)$ in the $l$-th level of pyramid $\mathbb{F}$.

\begin{figure}
	 	\centering
	 	\includegraphics[width = 0.5\textwidth]{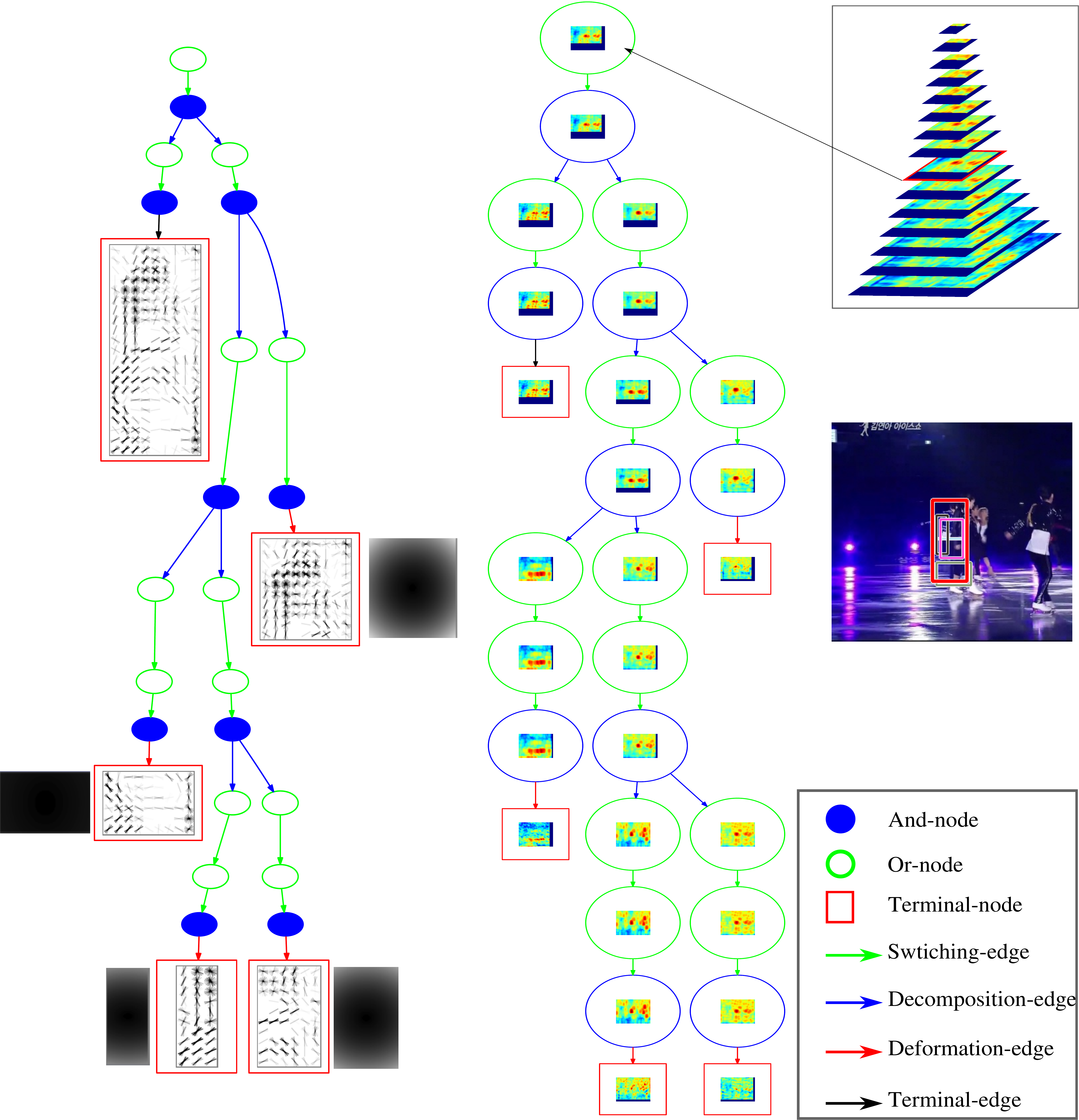}
	 	\caption{Illustration of the spatial DP algorithm for parsing with AOGs (e.g., $AOG_{172}$ in the \textit{left}).  \textit{Right-middle}: The input image (ROI in the 173-th frame in the ``Skating1" sequence) and the inferred object configuration. \textit{Right-top}: The score map pyramid for root Or-node. \textit{Middle}: For each node in AOG, we show one level of score map pyramid at which the optimal parse tree is retrieved.} \label{fig:dp}
	 	%\vspace{-3mm} 
	 \end{figure}

Given an AOG $\mathcal{G}=(V_{T}, V_{And}, V_{Or}, E, \Theta)$ (e.g., the left in Fig.\ref{fig:dp}), we define four types of edges, i.e., $E=E_{T}\cup E_{Def}\cup E_{Dec} \cup E_{Switch}$ as shown in Fig.\ref{fig:dp}. We elaborate the definitions of parameters $\Theta = (\Theta^{app}, \Theta^{def}, \Theta^{bias})$: 
\begin{itemize}[leftmargin=*]
\item [i)] Each terminal-node $\mathbf{t}\in V_T$  has appearance parameters $\theta^{app}_\mathbf{t}\subset \Theta^{app}$, which is used to ground a terminal-node to image data. 
\item [i)] The parent And-node $A$ of a part terminal-node with deformation edge has deformation parameters $\theta^{def}_A\subset \Theta^{def}$. They are used for penalizing local displacements when placing a terminal-node around its anchor position.  We note that the object template is not allowed to perturb locally in inference since we infer the optimal part configuration for each given object location in the pyramid with sliding window technique used, as done in the DPM~\cite{DPM}, so the parent And-node of the object terminal-node does not have deformation parameters.
\item [iii)]  A child And-node of the root Or-node has a bias term $\Theta^{bias}=\{b\}$. We do not define bias terms for child nodes of other Or-nodes.
\end{itemize}
    %In our object AOG, parts can be placed at either the same spatial resolution or twice the spatial resolution w.r.t. an object, which is determined automatically in learning the initial AOG in the first frame. 

\textit{Appearance Features.}  We use three types of features: histogram of oriented gradient (HOG) \cite{HOG}, local binary pattern features (LBP) \cite{LBP}, and RGB color histograms (for color videos). 

\textit{Deformation Features.} Denote by $\delta=[dx, dy]$ the displacement of placing a terminal-node around its anchor location. The deformation feature is defined by $\Phi^{def}(\delta) = [dx^2, dx, dy^2, dy]'$ as done in DPMs~\cite{DPM}.
 
  We use linear functions to evaluate both appearance scores and deformation scores. The score functions of nodes in an AOG are defined as follows:
\begin{itemize} %[leftmargin=*]
\item [i)] For a terminal-node $\mathbf{t}$, its score at a position $p$ is computed by, 
	\begin{equation}
	\text{Score}(\mathbf{t}, p|\mathbb{F}) = <\theta_\mathbf{t}^{app}, \mathbb{F}(\mathbf{t}, p)>
	\end{equation}
	where $<\cdot, \cdot>$ represents inner product and $\mathbb{F}(\mathbf{t},p)$ extracts features in feature pyramid.  
	
\item [ii)]For an Or-node $O$, its score at position $p$ takes the maximum score over its child nodes,
	\begin{equation}
	\text{Score}(O, p|\mathbb{F}) = \max_{c\in ch(O)} \text{Score}(c, p|\mathbb{F})
	\end{equation}
	where $ch(v)$ denotes the set of child nodes of a node $v$. 
	
\item [iii)] For an And-node $A$, we have three different functions w.r.t. the type of its out-edge (i.e., Terminal-,  Deformation-, or Decomposition-edge), 
	\begin{align}
	&\text{Score}(A, p|\mathbb{F}) = \label{eqn:andNodeScore}\\
	&\nonumber  \begin{cases}
	\text{Score}(\mathbf{t}, p|\mathbb{F}),  \quad e_{A,\mathbf{t}} \in E_{T}\\	
	\max_{\delta}[\text{Score}(\mathbf{t},p\oplus\delta|\mathbb{F}) - <\theta^{def}_A, \Phi^{def}(\delta)>],  e_{A,\mathbf{t}} \in E_{Def}\\
	\sum_{c\in ch(A)}\text{Score}(c, p|\mathbb{F}),  \quad e_{A, c}\in E_{Dec}
	\end{cases}	
	\end{align}	
	where the first case is for sharing score maps between the object terminal-node and its parent And-node since we do not allow local deformation for the whole object, the second case for computing transformed score maps of parent And-node of a part terminal-node which is allowed to find the best placement through distance transformation~\cite{DPM},  $\oplus$ represents the displacement operator in the position space in $\mathbf{\Lambda}$, and the third case for computing the score maps of an And-node which has two child nodes through composition. %The deformed score map can be computed by the distance transformation method efficiently \cite{DPM}. 
\end{itemize}

	 \begin{algorithm} [!ht]
	 	  \SetAlgoLined
	 	 \KwIn{An image $I_i$, a bounding box $B_i$, and an AOG $\mathcal{G}$}
	 	 \KwOut{$\text{Score}(I_i|B_i;\mathcal{G})$ in Eqn.(8) and the optimal configuration $\mathcal{C}_i^*$ from the parse tree for the object at frame $i$.}
	 	  \textit{Initialization}: Build the depth-first search (DFS) ordering queue ($Q_{DFS}$) of all nodes in the AOG\;
	 	 \underline{\textbf{Step 0:} Compute scores for all nodes in $Q_{DFS}$}\;
	 	  \While{$Q_{DFS}$ is not empty}{
	 	   Pop a node $v$ from the $Q_{DFS}$\; 
	 	   \uIf{$v$ is an Or-node}{
	 	   	Score(v) = $\max_{u\in ch(v)}$ Score(u); // $ch(v)$ is the set of child nodes of $v$
	 	   }
	 	   \uElseIf{$v$ is an And-node} {
	 	   	Score(v) = $\sum_{u\in ch(v)}$ LocalMax(Score(u))
	 	   }
	 	   \ElseIf{$v$ is a Terminal-node} {
	 	   	Compute the filter response map for $I_{\mathcal{N}(\Lambda_{v})}$. // $\mathcal{N}(\Lambda_{v})$ represents the image domain of the LocalMax operation of Terminal-node $v$.
	 	   }  
	 	  }
	 	  {$\text{Score}(I_i|B_i;\mathcal{G})$ = Score(RootOrNode)}.\;
	 	  \underline{\textbf{Step 1:} Compute $\mathcal{C}_i^*$ using the breadth-first search}\;
	 	  $Q_{BFS}=\{\text{RootOrNode}\}$, $\mathcal{C}_i^*=(B_i)$, $k=1$\;
	 	  \While{$Q_{BFS}$ is not empty}{
	 	   Pop a node $v$ from the $Q_{BFS}$\; 
	 	   \uIf{$v$ is an Or-node}{
	 	   	Push the child node $u$ with maximum score into $Q_{BFS}$(i.e., Score(u)=Score(v)).
	 	   }
	 	   \uElseIf{$v$ is an And-node} {
	 	   	Push all the child nodes $v$'s into $Q_{BFS}$.
	 	   } 
	 	   \ElseIf{$v$ is a Terminal-node} {
	 	   	Add $B_i^{(k)}=\text{Deformed}(\Lambda_v)$  to $\mathcal{C}_i^*=(\mathcal{C}_i^*, B_i^{(k)})$. Increase $k=k+1$.
	 	   }
	 	  }
	 	   \caption{The spatial DP algorithm for parsing with the   AOG, Parse($I_i|B_i;\mathcal{G}$)}\label{alg:parse}
	 	 \end{algorithm} 
	 
%\vspace{-2mm}
\subsection{Tracking-by-Parsing}
With scoring functions defined above, we present a spatial DP and a temporal DP algorithms in solving Eqn.(\ref{eq:dp2}). 

\textbf{Spatial DP:}
The DP algorithm (see Algorithm~\ref{alg:parse}) consists of two stages: 
\textit{(i) The bottom-up pass} computes score map pyramids (as illustrated in Fig.~\ref{fig:dp}) for all nodes following the depth-first-search (DFS) order of nodes. It computes matching scores of all possible parse trees at all possible positions in feature pyramid.
\textit{(ii) In the top-down pass}, we first find all candidate positions for the root Or-node $O$ based on its score maps and current threshold $\tau_{\mathcal{G}}$ of the object AOG, denoted by 
\begin{equation}
\Omega_{cand} = \{p; \text{Score}(O,p|\mathbb{F}) \geq \tau_{\mathcal{G}} \text{ and } p \in \mathbf{\Lambda}\}.\label{eqn:cand}
\end{equation}
Then, following BFS order of nodes, we retrieve the optimal parse tree at each $p\in \mathbb{P}$: starting from the root Or-node, we select the optimal branch (with the largest score) of each encountered Or-node, keep the two child nodes of each encountered And-node, and retrieve the optimal position of each encountered part terminal-node (by taking $\arg \max$ for the second case in Eqn.(\ref{eqn:andNodeScore})). 

After spatial parsing, we apply non-maximum suppression (NMS) in computing the optimal parse trees with a predefined intersection-over-union (IoU) overlap threshold, denoted by $\tau_{\text{\tiny{NMS}}}$. We keep top $N_{best}$ parse trees to infer the best $B_t^*$ together with a temporal DP algorithm, similar to the strategies used in~\cite{nbestPartModels,mbestMRF}. 

\textbf{Temporal DP:}
Assuming that all the N-best candidates for $B_2, \cdots, B_t$ are memoized after running spatial DP algorithm in $I_2$ to $I_t$, Eqn.(\ref{eq:dp2}) corresponds to the classic DP formulation of forward and backward inference for decoding HMMs with $- \text{Score}(I_i|B_i;\mathcal{G})$ being the singleton ``data"  term and $\text{Cost}(B_i|B_{i-1})$ the pairwise cost term.

Let $\mathbb{B}_i[B_i]$ be energy of the best object states in the first $i$ frames with the constraint that the $i$-th one is $B_i$. We have, 
\begin{align}
\nonumber \mathbb{B}_1[B_1] &= -\text{Score}(I_1|B_1;\mathcal{G}), \\
\nonumber \mathbb{B}_i[B_i] &= -\text{Score}(I_i|B_i;\mathcal{G}) \\
&\qquad +\min_{B_{i-1}}(\mathbb{B}_{i-1}[B_{i-1}]+\text{Cost}(B_{i}| B_{i-1})). \label{eq:DPassignment}
\end{align}
When $B_1$ is the input bounding box. Then, the temporal DP algorithm consists of two steps:
\begin{enumerate}
\item[i)] \textit{The forward step} for computing all $\mathbb{B}_i[B_i]$'s, and caching the optimal solution for $B_{i-1}$ as a function of $B_i$ for later back-tracing starting at $i=2$,
\begin{align}
\nonumber \mathbf{T}_i[B_i] = \arg\min_{B_{i-1}}\{&\mathbb{B}_{i-1}[B_{i-1}]+\text{Cost}(B_{i}| B_{i-1})\}.
\end{align}

\item[ii)] \textit{The backward step} for finding the optimal trajectory $(B_1, B_2^*, \cdots, B_t^*)$, where we first take, 
\begin{equation}
B_t^* = \arg \min_{B_t} \mathbb{B}_t[B_t],
\end{equation}
and then in the order of $i=t-1, \cdots, 2$ trace back,
\begin{equation}
B_i^* = \mathbf{T}_{i+1}[B_{i+1}^*].
\end{equation}
\end{enumerate}

In practice, we often do not need to run temporal DP in the whole time range $[1, t]$, especially for long-term tracking, since the target object might have changed significantly or we might have camera motion, instead we only focus on some short time range, $[t-\Delta t, t]$ (see settings in experiments).  

\textit{Remarks:} In our TLP method, we apply the spatial and the temporal DP algorithms in a stage-wise manner and without tracking parts explicitly. Thus, we do not introduce loops in inference. If we instead attempt to learn a joint spatial-temporal AOG, %for event or activity parsing, 
it will be a much more difficult problem due to loops in joint spatial-temporal inference, and approximate inference is used. 

\textbf{Search Strategy:} During tracking, at time $t$, $B_t$ is initialized by $B_{t-1}$, and then a rectangular region of interest (ROI) centered at the center of $B_t$ is used to compute feature pyramid and run parsing with AOG. The ROI is first computed as a square area with the side length being $s_{\text{\tiny{ROI}}}$ times longer than the maximum of width and height of $B_t$ and then is clipped with the image domain. If no candidates are found (i.e., $\Omega_{cand}$ is empty), we will run the parsing in whole image domain. So, our AOGTracker is capable of re-detecting a tracked object. If there are still no candidates (e.g., the target object was completely occluded or went out of camera view), the tracking result of this frame is set to be invalid and we do not need to run the temporal DP.  

%\begin{figure*}
%	\centering
%	\includegraphics[width = 0.95\textwidth]{Fig/LearnInitAOG-simple.pdf}
%	\caption{Illustration of online learning of object AOGs. (a) The three steps in learning the initial object AOG:  \textit{Left:} The structure of the initial object AOG, which is learned by pruning the full structure AOG based on the training error rates of nodes in the full structure AOG. \textit{Middle}: The object AOG with refined appearance templates for terminal-nodes after latent training and hard negative mining. \textit{Right}: The final object And-Or Tree (AOT) with the most discriminative configuration used (and therefore smaller model complexity) for more efficient tracking. (b) The three steps in updating the object AOG, which are similiar to those in (a). See text for details. (Best viewed in color and with magnification)} \label{fig:initAOG}
%	\vspace{-3mm} 
%\end{figure*}

%\vspace{-2mm}
\subsection{The Trackability of an Object AOG}\label{sec:intrackability}

To detect critical moments online, we need to measure the quality of an object AOG, $\mathcal{G}$ at time $t$.  We compute its trackability based on the score maps in which the optimal parse tree is placed. For each node $v$ in the parse tree, we have its position in score map pyramid (i.e., the level of pyramid and the location in that level), $(l_v, x_v, y_v)$. We define the trackability of node $v$ by,
\begin{align}
\text{Trackability}(v|I_t, \mathcal{G}) = 
S(l_v, x_v, y_v)-\mu_S \label{eq:intrackability}
\end{align}  
where $S(l_v, x_v, y_v)$ is the score of node $v$, $\mu_S$ the mean score computed from the whole score map. Intuitively, we expect the score map of a discriminative node $v$ has peak and steep landscape, as investigated in \cite{kwon_patch}. The trackabilities of part nodes are used to infer  partial occlusion and local structure variations, and trackability of the inferred parse tree indicate the ``goodness" of current object AOG. We note that we treat trackability and intrackability (i.e., the inverse of th trackability) exchangeably. More sophisticated definitions of intrackability in tracking are referred to \cite{intrackability}.

We model trackability by a Gaussian model whose mean and standard derivation are computed incrementally in $[2, t]$. At time $t$, a tracked  object is said to be ``intrackable" if its trackability is less than $mean_{trackability}(t) - 3\cdot std_{trackability}(t)$. We note that the tracking result could be still valid even if it is ``intrackable" (e.g., in the first few frames in which the target object is occluded partially, especially by similar distractors).  

%\vspace{-2mm}
\section{Online Learning of Object AOGs}\label{sec:online}
In this section, we present online learning of object AOGs, which consists of three components: (i) Maintaining a training dataset based on tracking results; (ii) Estimating parameters of a given object AOG; and (iii) Learning structure of the object AOG by pruning full structure AOG, which requires (ii) in the process. 

%\vspace{-2mm}
\subsection{Maintaining the Training Dataset Online}
Denote by $D_t=D^+_t \cup D^-_t$ the training dataset at time $t$, consisting of $D^+_t$, a positive dataset, and $D^-_t$, a negative dataset.

In the first frame, we have $D^+_1=\{(I_1, B_1)\}$ and let $B_1 = (x_1, y_1, w_1, h_1)$. We augment it with eight locally shifted positives, i.e., $\{I_1, B_{1,i}; i=1,\cdots,8\}$ where $x_{1, i} \in \{x_1 \pm d\}$ and $y_{1, i} \in \{ y_ \pm d\}$ with width and height not changed. $d$ is set to the cell size in computing HOG features. The initial $D^-_1$ uses the whole remaining image $I_{\overline{\Lambda_{B_1}}}$ for mining hard negatives in training. 

At time $t$, if $B_t$ is valid according to tracking-by-parsing, we have $D_t^+=D_{t-1}^+\cup \{(I_t, B_t)\}$, and add to $D_t^-$ all other candidates in $\Omega_{cand}$ (Eqn.~\ref{eqn:cand}) which are not suppressed by $B_t$ according to NMS (i.e., hard negatives). Otherwise, we have $D_t=D_{t-1}$.

%\vspace{-2mm}
\subsection{Estimating Parameters of a Given Object AOG}
We use latent SVM method (LSVM)~\cite{DPM}. Based on the scoring functions defined in Section~\ref{sec:scoreFunction}, we can re-write the scoring function of applying a given object AOG, $\mathcal{G}$ on a training example (denoted by $I_B$ for simplicity), 
\begin{equation}
\text{Score}(I_B; \mathcal{G})=\max_{pt\in \Omega_{\mathcal{G}}} <\Theta, \Phi(\mathbb{F}, pt)>
\end{equation}
where $pt$ represents a parse tree, $\Omega_{\mathcal{G}}$ the space of parse trees, $\Theta$ the concatenated vector of all parameters, $\Phi(\mathbb{F},pg)$ the concatenated vector of appearance and deformation features in feature pyramid $\mathbb{F}$  w.r.t. parse tree $pt$, and the bias term.

The objective function in estimating parameters is defined by the $l_2$-regularized empirical hinge loss function, 
\begin{align}
\nonumber \mathcal{L}_{D_t}(\Theta) = {1\over 2}||\Theta||_2^2 + &{C\over {|D_t|}}[\sum_{I_B\in D^+_t}\max(0, 1-\text{Score}(I_B; \mathcal{G}))\\
&\sum_{I_B\in D^-_t}\max(0, 1+\text{Score}(I_B; \mathcal{G}))] \label{eqn:loss}
\end{align}
where $C$ is the trade-off parameter in learning. Eqn.(~\ref{eqn:loss}) is a semi-convexity function of the parameters $\Theta$ due to the empirical loss term on positives. 

In optimization, we utilize an iterative procedure in a ``coordinate descent" way. We first convert the objective function to a convex function by assigning latent values for all positives using the spatial DP algorithm. Then, we estimate parameters. While we can use stochastic gradient descent as done in DPMs~\cite{DPM}, we adopt LBFGS method in practice~\footnote{We reimplemented the matlab code available at http://www.cs.ubc.ca/~schmidtm/Software/minConf.html in c++.} \cite{LBFGS} since it is more robust and efficient with parallel implementation as investigated in~\cite{WLLSVM,FFLD}. The detection threshold, $\tau_{\mathcal{G}}$ is estimated as the minimum score of positives.

\begin{figure}
	\centering
	\includegraphics[width = 0.5\textwidth]{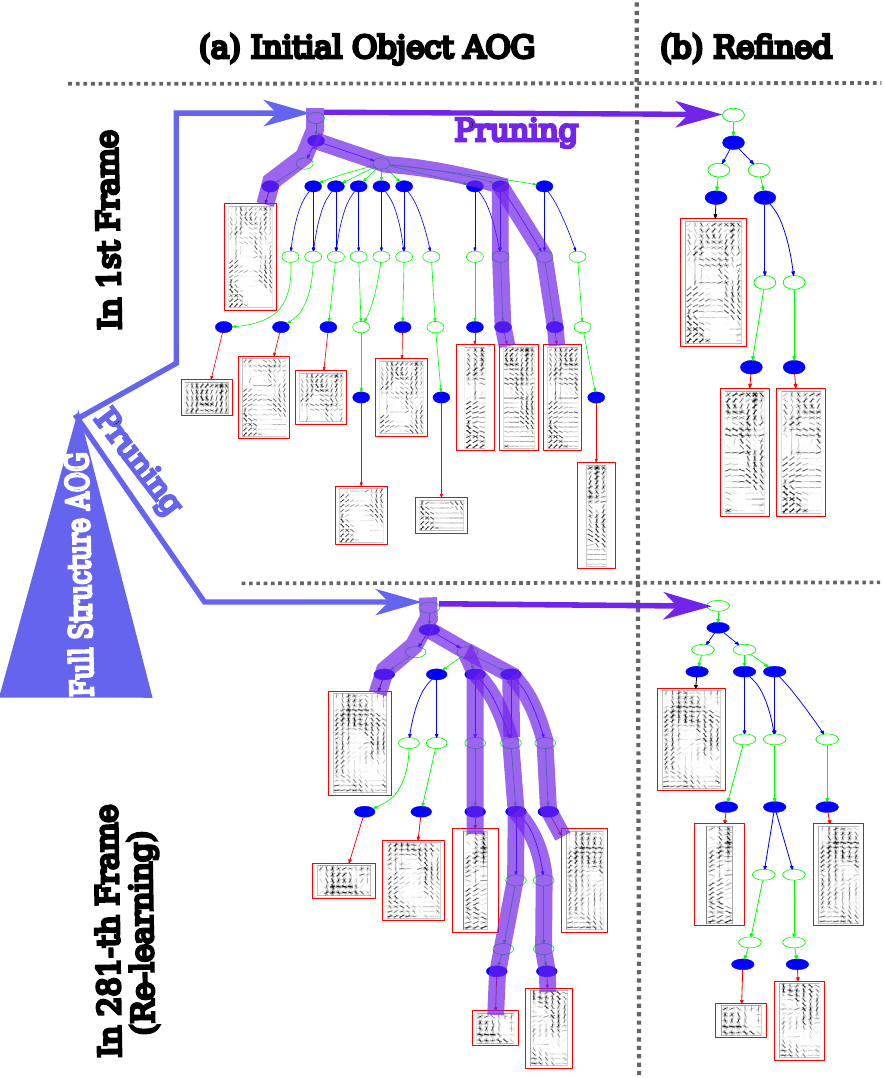}
	\caption{Illustration of learning an object AOG in the first frame (top) and re-learning an object AOG in the 281-th frame when a critical moment has triggered. It consists of two steps: (a) learning initial object AOG by pruning branches of Or-nodes in full structure AOG, and (b) learning refined object AOG by pruning part configurations w.r.t. majority voting in positive re-labeling in LSVM.} \label{fig:initAOG}
	%\vspace{-3mm} 
\end{figure}

%\vspace{-2mm}
\subsection{Learning Object AOGs}
With the training dataset $D_t$ and the full structure AOG constructed based on $B_1$, an object AOG is learned in three steps:

\textit{i) Evaluating the figure of merits of nodes in the full structure AOG.} We first train the root classifier (i.e., object appearance parameters and bias term) by linear SVM using $D^+_t$ and data-mining hard negatives in $D^-_t$. Then, the appearance parameters for each part terminal-node $\mathbf{t}$ is initialized by cropping out the corresponding portion in the object template~\footnote{We also tried to train the linear SVM classifiers for all the terminal-nodes individually using cropped examples, which increases the runtime, but does not improve the tracking performance in experiments. So, we use the simplified method above.}. Following DFS order, we evaluate the figure of merit of each node in the full structure AOG by its training error rate. The error rate is calculated on $D_t$ where the score of a node is computed w.r.t. scoring functions defined in Section \ref{sec:scoreFunction}. 
The smaller the error rate is, the more discriminative a node is. 

\textit{ii) Retrieving an initial object AOG and re-estimating  parameters.}  We retrieve the most discriminative subgraph in the full structure AOG as initial object AOG. Following BFS order, we start from the root Or-node, select for each encountered Or-node the best child node (with the smallest training error rate among all children) and the child nodes whose training error rates are not bigger than that of the best child by some predefined small positive value (i.e., preserving ambiguities), keep the two child nodes for each encountered And-node, and stop at each encountered terminal-node. We show two examples in the left of Fig.~\ref{fig:initAOG}. We train the parameters of initial object AOG using LSVM~\cite{DPM} with two rounds of positive re-labeling and hard negative mining respectively. %within each round of positive re-labeling.  

\begin{figure}
	\centering
	\includegraphics[width = 0.5\textwidth]{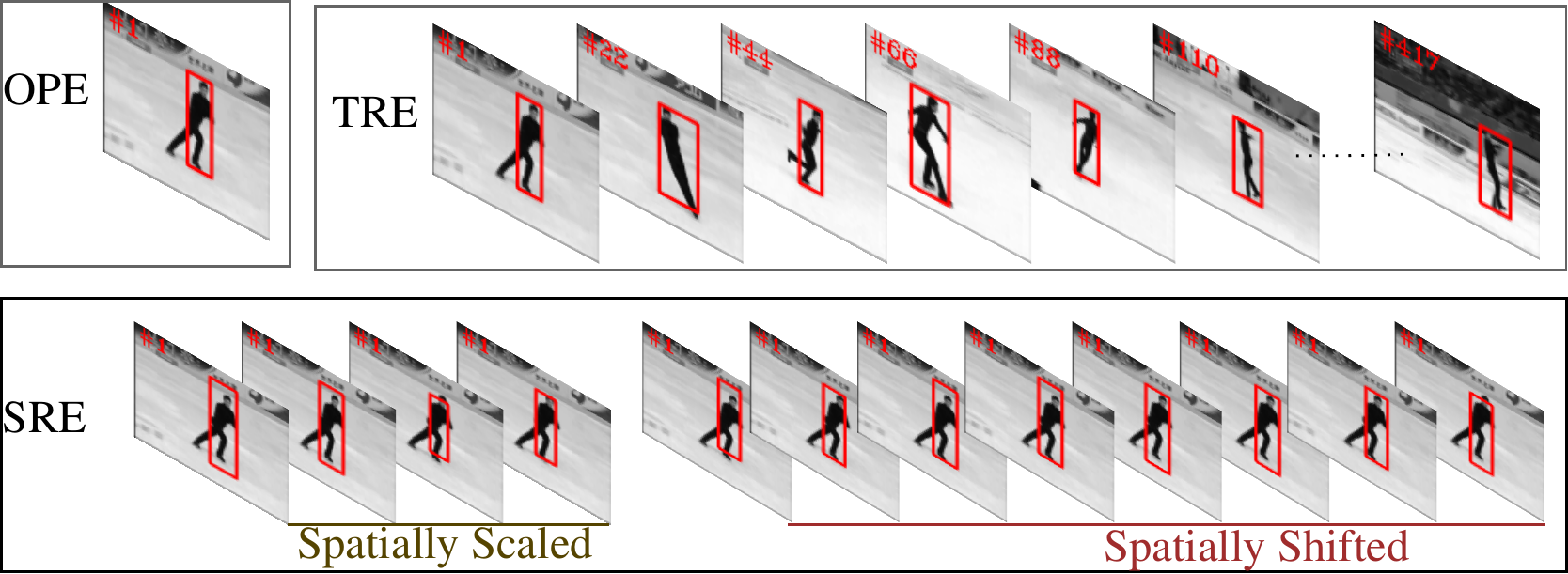}
	\caption{Illustration of the three types of evaluation methods in TB-100/50/CVPR2013. \textit{In one-pass evaluation (OPE)}, a tracker is initialized in the first frame and let it track the target until the end of the sequence. \textit{In temporal robustness evaluation (TRE)}, a tracker starts at different
		starting frames initialized with the corresponding ground-truth bounding boxes and then tracks the object until the end. 20 starting frames (including the first frame) are used in TB-100. \textit{In spatial robustness evaluation (SRE)}, a tracker runs multiple times with spatially scaled (4 types) and shifted (8 types of perturbation) initializations in the first frame.}\label{fig:ope}
	%\vspace{-3mm} 
\end{figure}

\textit{iii) Controlling model complexity.} To do that, a refined object AOG for tracking is obtained by further selecting the most discriminative part configuration(s) in the initial object AOG learned in the step ii). The selection process is based on latent assignment in relabeling positives in LSVM training. A part configuration in the initial object AOG is pruned if it relabeled less than 10\% positives (see the right of Fig.~\ref{fig:initAOG}). We further train the refined object AOG with one round latent positive re-labeling and hard negative mining. By reducing model complexity, we can speed up the tracking-by-parsing procedure. 

\textit{Verification of a refined object AOG.} We run parsing with a refined object AOG in the first frame. The refined object AOG is accepted if the score of the optimal parse  tree is greater than the threshold estimated in training and the IoU overlap between the predicted bounding box and the input bounding box is greater than or equals the IoU NMS threshold, $\tau_{\text{\tiny{NMS}}}$ in detection.  

\textit{Identifying critical moments in tracking.}
A critical moment means a tracker has become ``uncertain" and at the same time accumulated ``enough" new samples, which is triggered in tracking when two conditions were satisfied.  The first is that the number of frames in which a tracked object is ``intrackable" was larger than some value, $N_{\text{\tiny{Intrackable}}}$. The second is that the number of new valid tracking results are greater than some value, $N_{\text{\tiny{NewSample}}}$. Both are accumulated from the last time an  object AOG was re-learned. 

\textit{The spatial resolution of placing parts.} In learning object AOGs, we first place parts at the same spatial resolution as the object. 
If the learned object AOG was not accepted in verification, we then place parts at twice the spatial resolution w.r.t. the object and re-learn the object AOG. In our experiments, the two specifications handled all testing sequences successfully.

\textit{Overall flow of online learning.} In the first frame or when a critical moment is identified in tracking, we learn both structure and  parameters of an object AOG, otherwise we update parameters only if the tracking result is valid in a frame based on tracking-by-parsing.

\begin{table}
\centering
\resizebox{0.5\textwidth}{!}{%
\begin{tabular}{|c|c|c|c|c|c|c|c|c|c|c|c|c|c|c|}
	\hline 
	 & \multicolumn{10}{c|}{Representation} & \multicolumn{4}{c|}{Search}\\	 
	 \cline{2-15}
	 & \rotatebox[origin=c]{90}{Local} & \rotatebox[origin=c]{90}{Template} & \rotatebox[origin=c]{90}{Color}  & \rotatebox[origin=c]{90}{Histogram}  & \rotatebox[origin=c]{90}{Subspace} & \rotatebox[origin=c]{90}{Sparse} & \rotatebox[origin=c]{90}{Binary or Haar} & \rotatebox[origin=c]{90}{Discriminative} & \rotatebox[origin=c]{90}{Generative} & \rotatebox[origin=c]{90}{Model Update} & \rotatebox[origin=c]{90}{Particle Filter} & \rotatebox[origin=c]{90}{MCMC} & \rotatebox[origin=c]{90}{Local Optimum} & \rotatebox[origin=c]{90}{Dense Sampling} \\
	 \hline
	 ASLA~\cite{ASLA} & \checkmark &  \checkmark & & & \checkmark & \checkmark & & & \checkmark & \checkmark & \checkmark & & & \\
	 \hline
	 BSBT~\cite{BSBT} & & & & & & & H & \checkmark & & \checkmark & & & & \checkmark \\
	 \hline
	 CPF~\cite{CPF} & \checkmark & & \checkmark& \checkmark& & &  &  & \checkmark&  & \checkmark& & &   \\
	 \hline
	 CSK~\cite{CSK} & & \checkmark & & & & &  &  \checkmark& & \checkmark & & & &  \checkmark\\ \hline
	 CT~\cite{CT} & & & & & & & H & \checkmark & & \checkmark & & & & \checkmark  \\ \hline
	 CXT~\cite{CXT} & & & & & &  & B & \checkmark & & \checkmark & & & & \checkmark \\ \hline
	 DFT~\cite{DFT} & \checkmark& \checkmark& & & & &  &  & \checkmark& \checkmark & & & \checkmark&  \\ \hline
	 FOT~\cite{FOT} & \checkmark& \checkmark& & & & &  &  & \checkmark& \checkmark & & & \checkmark&  \\ \hline
	 FRAG~\cite{FRAG} & \checkmark& & & \checkmark& & &  &  & \checkmark&  & & & & \checkmark \\ \hline
	 IVT~\cite{IVT} & & \checkmark& & & \checkmark& &  &  & \checkmark& \checkmark & \checkmark& & &  \\ \hline
	 KMS~\cite{KMS} & & & \checkmark& \checkmark& & &  &  & \checkmark&  & & & \checkmark&\checkmark  \\ \hline  
	 L1APG~\cite{L1APG} & & \checkmark& & & \checkmark& \checkmark&  &  & \checkmark&  \checkmark& \checkmark& & &  \\ \hline  
	 LOT~\cite{LOT} & \checkmark& & \checkmark& & & &  &  & \checkmark&  \checkmark& \checkmark& & &  \\ \hline  
	 LSHT~\cite{LSHT} & \checkmark& & \checkmark& \checkmark& & & H &  & \checkmark&  \checkmark& & & &\checkmark  \\ \hline  
	 LSK~\cite{LSK} & \checkmark& \checkmark& & & \checkmark& \checkmark&  &  & \checkmark&  & & & \checkmark&  \\ \hline  
	 LSS~\cite{LSS} & \checkmark& \checkmark& & & \checkmark& \checkmark&  &  & \checkmark&  \checkmark& \checkmark& & &  \\ \hline  
	 MIL~\cite{MIL} & & & & & & &  H& \checkmark & & \checkmark & & & & \checkmark \\ \hline  
	 MTT~\cite{MTT} & \checkmark& & & \checkmark& \checkmark& &  & \checkmark &\checkmark & \checkmark & & & &  \\ \hline  
	 OAB~\cite{OAB} & & & & & & &  H& \checkmark & & \checkmark & & & & \checkmark \\ \hline  
	 ORIA~\cite{ORIA} & & \checkmark& & & \checkmark& &  H& \checkmark & & \checkmark & & & & \checkmark \\ \hline  
	 PCOM~\cite{PCOM} & \checkmark& & & \checkmark& \checkmark& &  & \checkmark &\checkmark & \checkmark & & & &  \\ \hline  
	 SCM~\cite{SCM} & \checkmark& \checkmark& & & \checkmark& \checkmark&  &  \checkmark& \checkmark&  \checkmark& \checkmark& & & \checkmark \\ \hline
	 SMS~\cite{SMS} & & & \checkmark& \checkmark& & &  &  & \checkmark&  & & & \checkmark&  \\ \hline
	 SBT~\cite{SBT} & & & & & & &  H& \checkmark & & \checkmark & & & & \checkmark \\ \hline
	 STRUCK~\cite{STRUCK} & & & & & & &  H& \checkmark & & \checkmark & & & & \checkmark \\ \hline
	 TLD~\cite{TLD} & \checkmark& & & & & &  B& \checkmark & & \checkmark & & & & \checkmark \\ \hline
	 VR~\cite{VR} & & & \checkmark& & & &  & \checkmark & & \checkmark & & & \checkmark &  \\ \hline
	 VTD~\cite{VTD} & & \checkmark& \checkmark& & \checkmark& &  &  & \checkmark& \checkmark & & \checkmark& &  \\ \hline
	 VTS~\cite{VTS} & \checkmark& \checkmark& \checkmark& & \checkmark& &  &  & \checkmark&  \checkmark& & \checkmark& &  \\ \hline
	 AOG & \checkmark& \checkmark& \checkmark& \checkmark& & & HOG [+Color] &  \checkmark& & \checkmark & & & \checkmark& \checkmark \\ \hline
\end{tabular}
}
\caption{Tracking algorithms evaluated in the TB-100 benchmark (reproduced from \cite{trackingBenchmarkPAMI}). }\label{table:trackers}
%\vspace{-3mm}
\end{table}

\begin{table*}
\centering
\resizebox{1.0\textwidth}{!}{%
\begin{tabular}{|c|c|c|c|c|c|c|c|c|c|c|c|c|}
	\hline Metric & \multicolumn{9}{c|}{Success Rate / Precision Rate}  \\ 
	\hline  Evaluation & \multicolumn{3}{c|}{OPE} &   \multicolumn{3}{c|}{SRE} & \multicolumn{3}{c|}{TRE} \\ 
	\hline  Subset & 100 & 50 & CVPR2013 & 100 & 50 & CVPR2013 & 100 & 50 & CVPR2013 \\ 
	\hline \rule[0ex]{0pt}{3ex} AOG Gain & 13.93 / 18.06  & 16.84 / 22.23 & 2.74 / 19.37  & 11.47 / 16.79   & 12.52 / 17.82  & 11.89 / 17.55   & 9.25 / 11.06 & 11.37 / 14.61  & 11.59 / 14.38  \\ 
	\hline 
	Runner-up & \multicolumn{2}{c|}{STRUCK\cite{STRUCK}} & SO-DLT\cite{rcnnTracker} / STRUCK\cite{STRUCK} & \multicolumn{6}{c|}{STRUCK\cite{STRUCK}}  \\
	\hline  \hline 
\end{tabular}}

\resizebox{1.0\textwidth}{!}{%
	\begin{tabular}{|p{3cm}|c|c|c|c|c|c|c|c|c|c|c|c|}
		\hline \rule[0ex]{0pt}{3ex} Subsets in TB-50 &  DEF(23)  & FM(25) & MB(19) & IPR(29) & BC(20) &OPR(32) & OCC(29) & IV(22) & LR(8) & SV(38) & OV(11)    \\ 
		\hline \rule[0ex]{0pt}{3ex} AOG Gain (success rate) &  15.89   & 15.56 & 17.29 & 12.29 & 17.81 & 14.04 & 14.7 & 15.73 & 6.65 &  18.38 & 15.99  \\
		\hline 
		Runner-up &  \multicolumn{2}{c|}{STRUCK\cite{STRUCK}} & \multicolumn{2}{c|}{TLD\cite{TLD} } &  \multicolumn{6}{c|}{SCM\cite{SCM}} & MIL\cite{MIL} \\
		\hline
	\end{tabular}}		
	\caption{Performance gain (in \%) of our AOGTracker in term of  success rate and precision rate in the benchmark~\cite{trackingBenchmarkPAMI}. Success plots of TB-100/50/CVPR2013 are shown in Fig.~\ref{fig:TB-AUC}. The success plots of the 11 subsets in TB-50 are shown in Fig.~\ref{fig:OPE-SbusetTB50}. Precision plots are provided in the supplementary material due to space limit here. } \label{table:gain}
	%\vspace{-2mm} 
\end{table*}

\begin{figure*}[t!]
	\centering
		\begin{subfigure}[t]{0.32\textwidth}
			\centering
			\includegraphics[width=1.0\textwidth, height=0.8\textwidth] {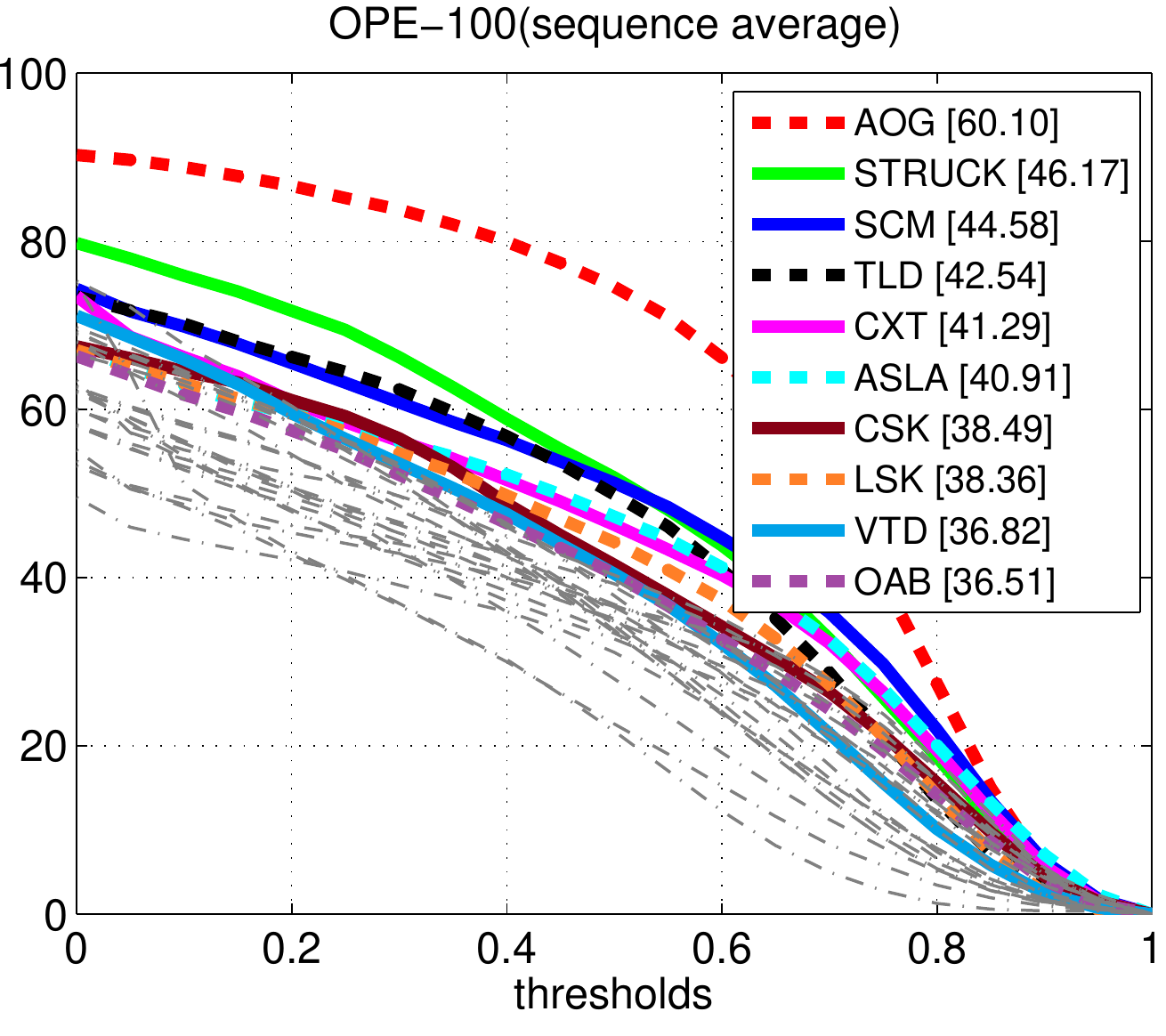}
		\end{subfigure}%
		~ 
		\begin{subfigure}[t]{0.32\textwidth}
			\centering
			\includegraphics[width=1.0\textwidth, height=0.8\textwidth]{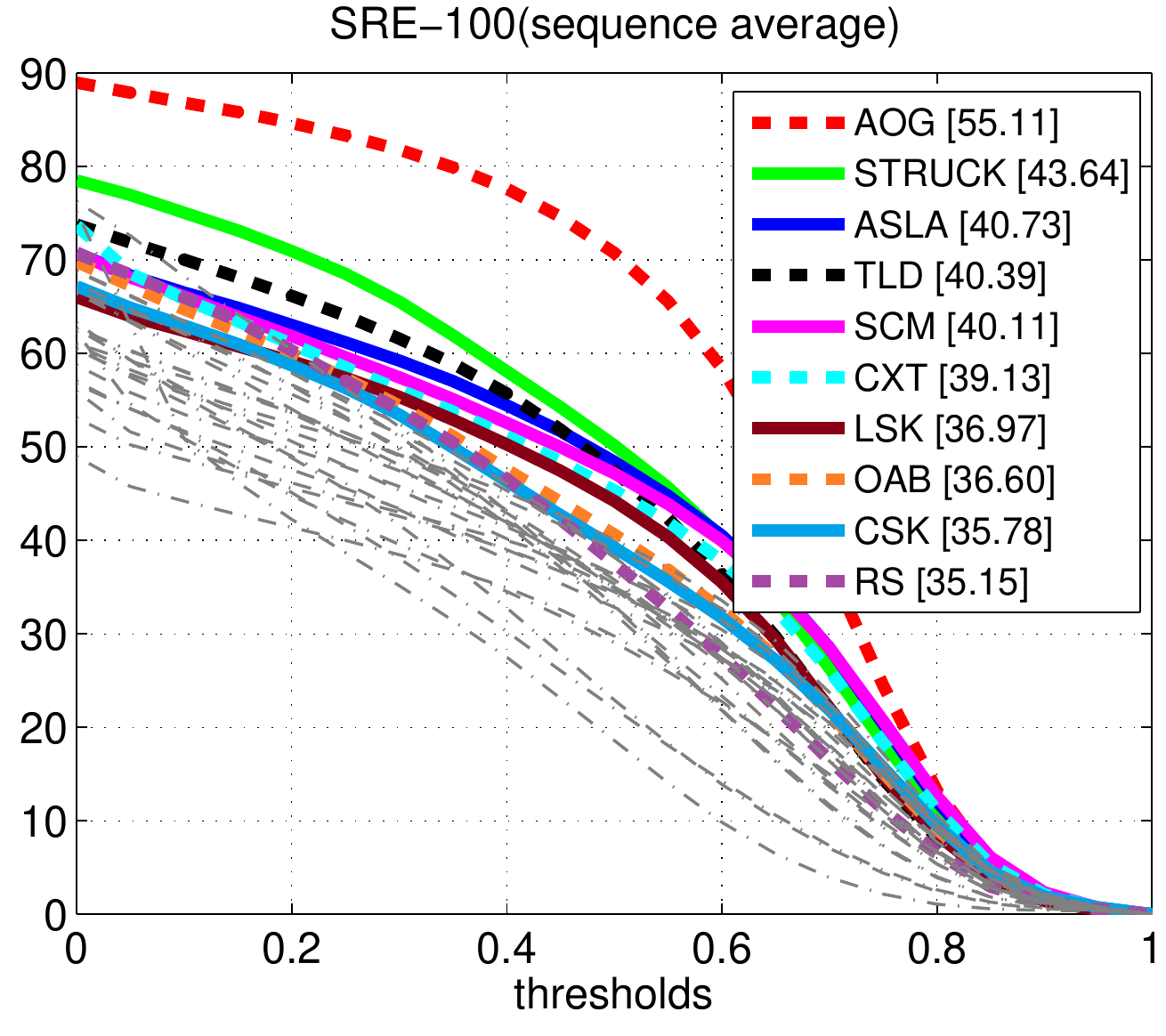}	
		\end{subfigure}%
		~
		\begin{subfigure}[t]{0.32\textwidth}
			\centering
			\includegraphics[width=1.0\textwidth, height=0.8\textwidth]{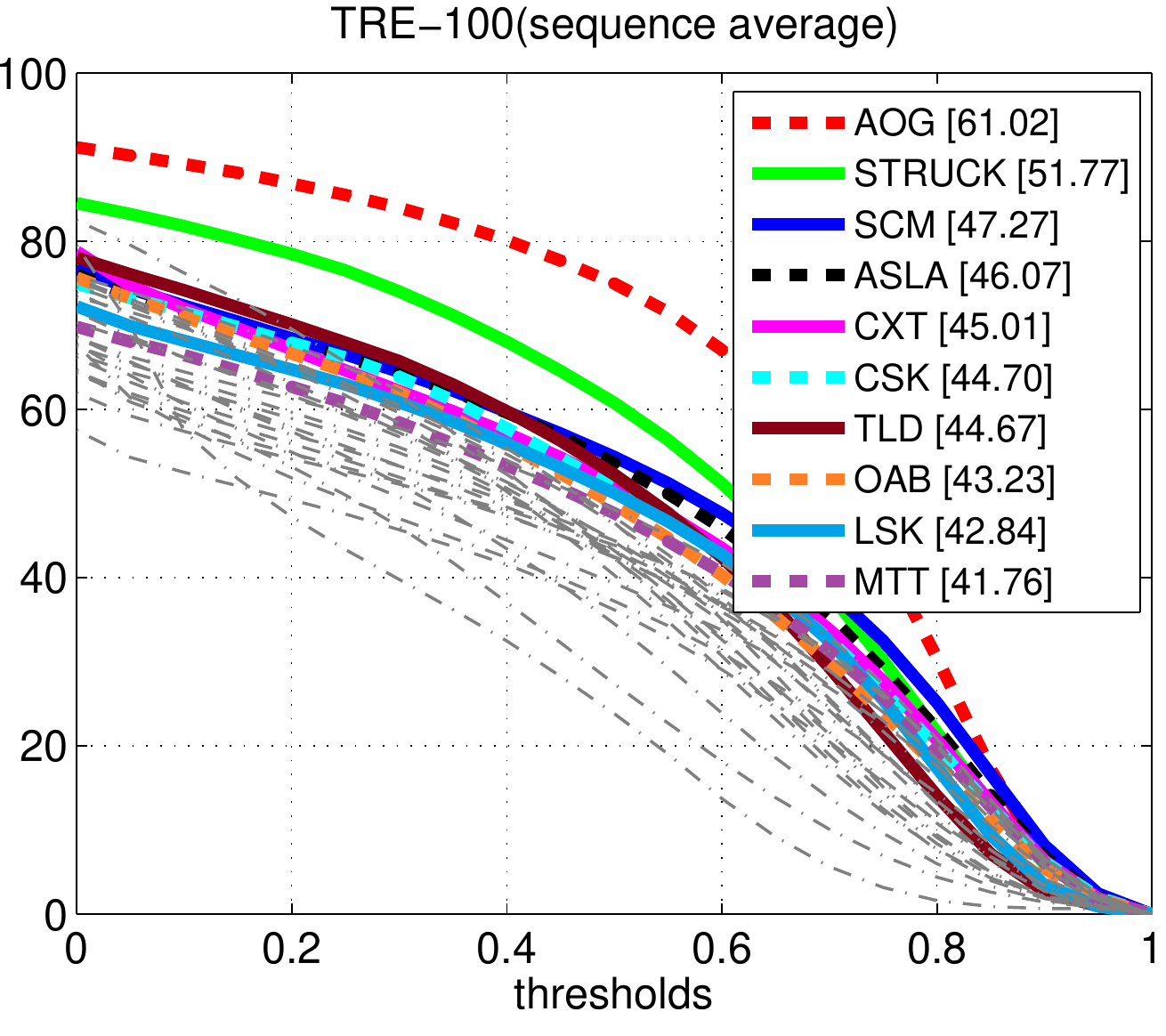}
		\end{subfigure}%
		\vspace{1mm}
		\begin{subfigure}[t]{0.32\textwidth}
			\centering
			\includegraphics[width=1.0\textwidth, height=0.8\textwidth]{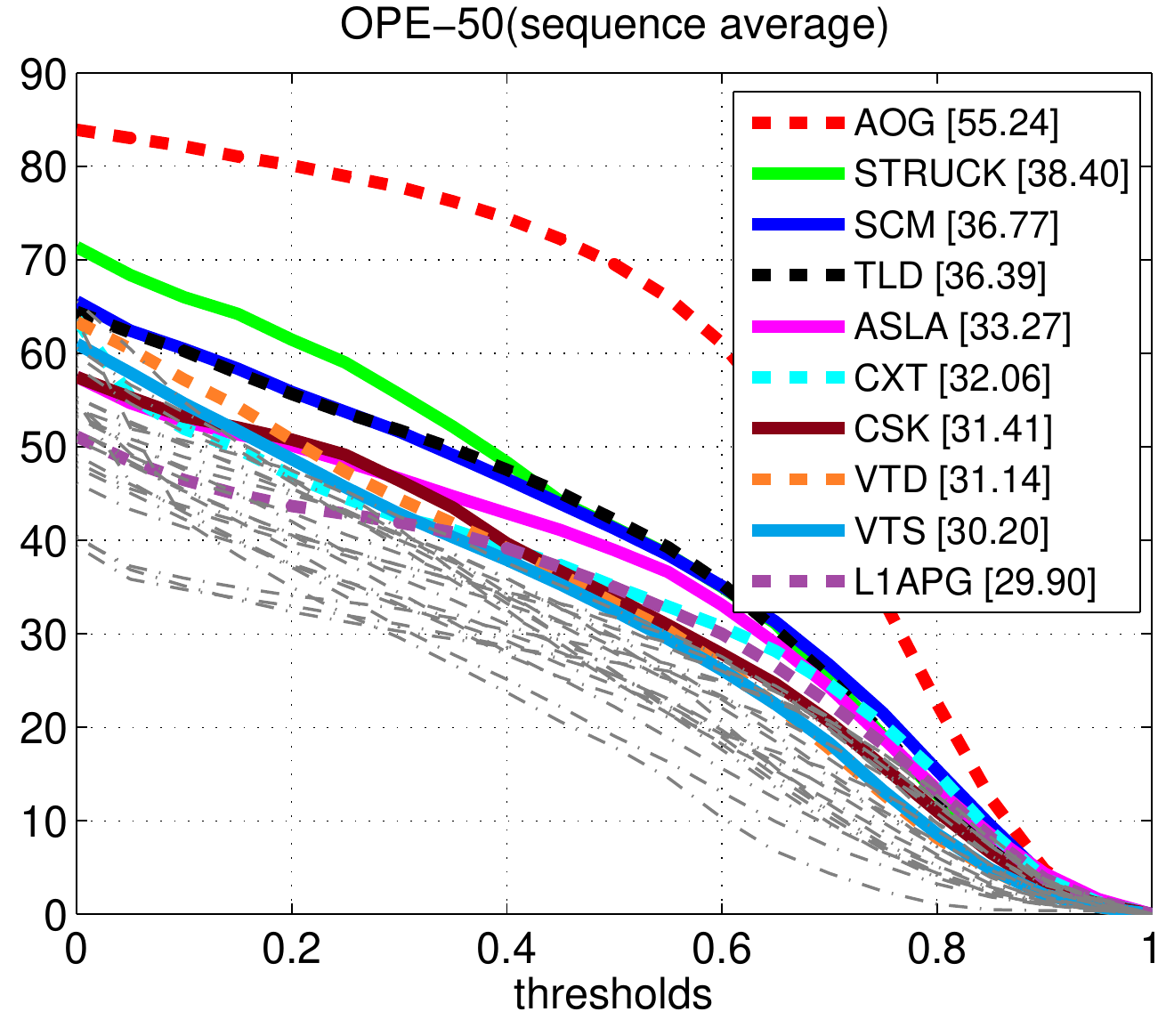}
		\end{subfigure}%
		~ 
		\begin{subfigure}[t]{0.32\textwidth}
			\centering
			\includegraphics[width=1.0\textwidth, height=0.8\textwidth]{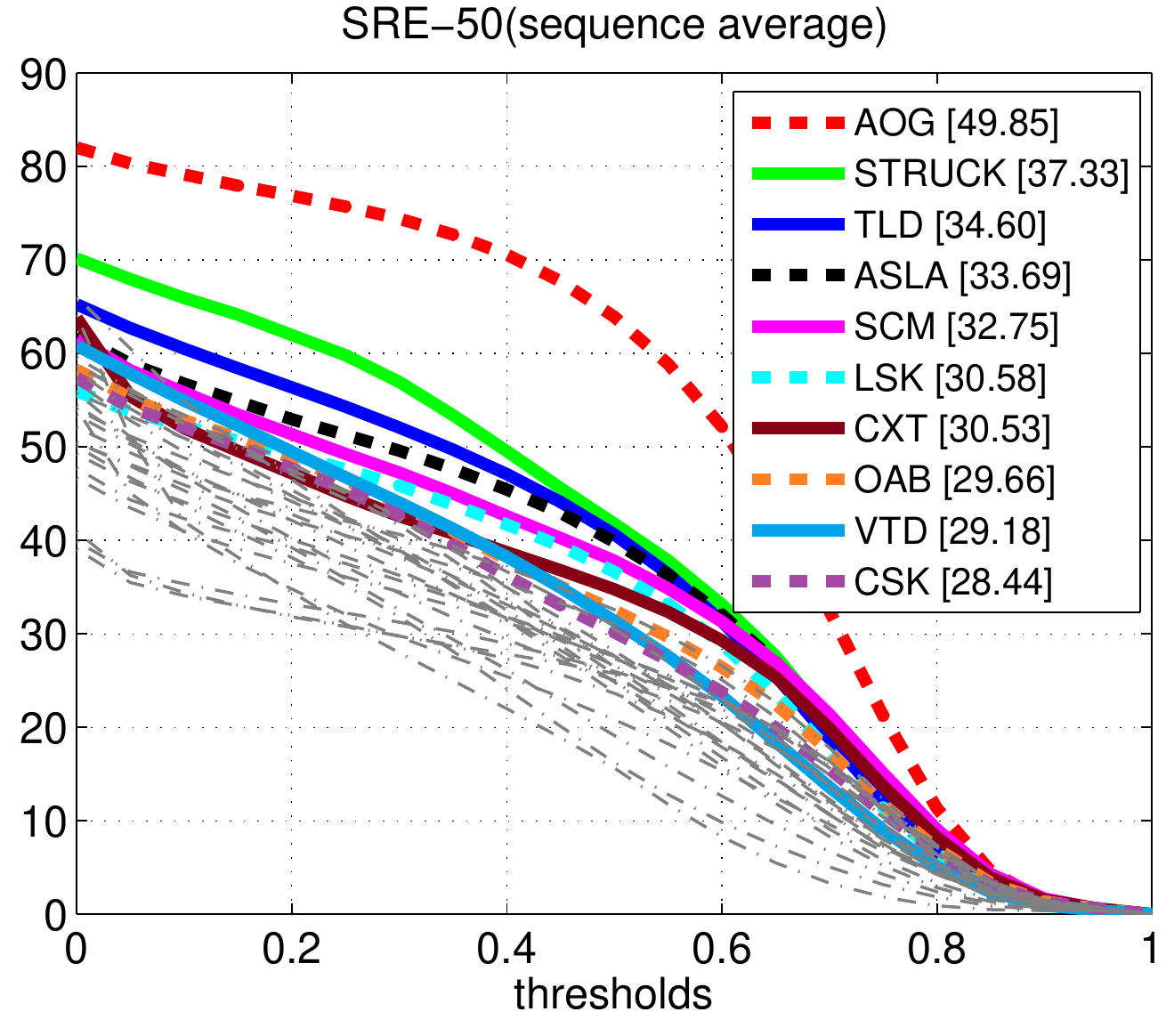}
		\end{subfigure}%
		~
		\begin{subfigure}[t]{0.32\textwidth}
			\centering
			\includegraphics[width=1.0\textwidth, height=0.8\textwidth]{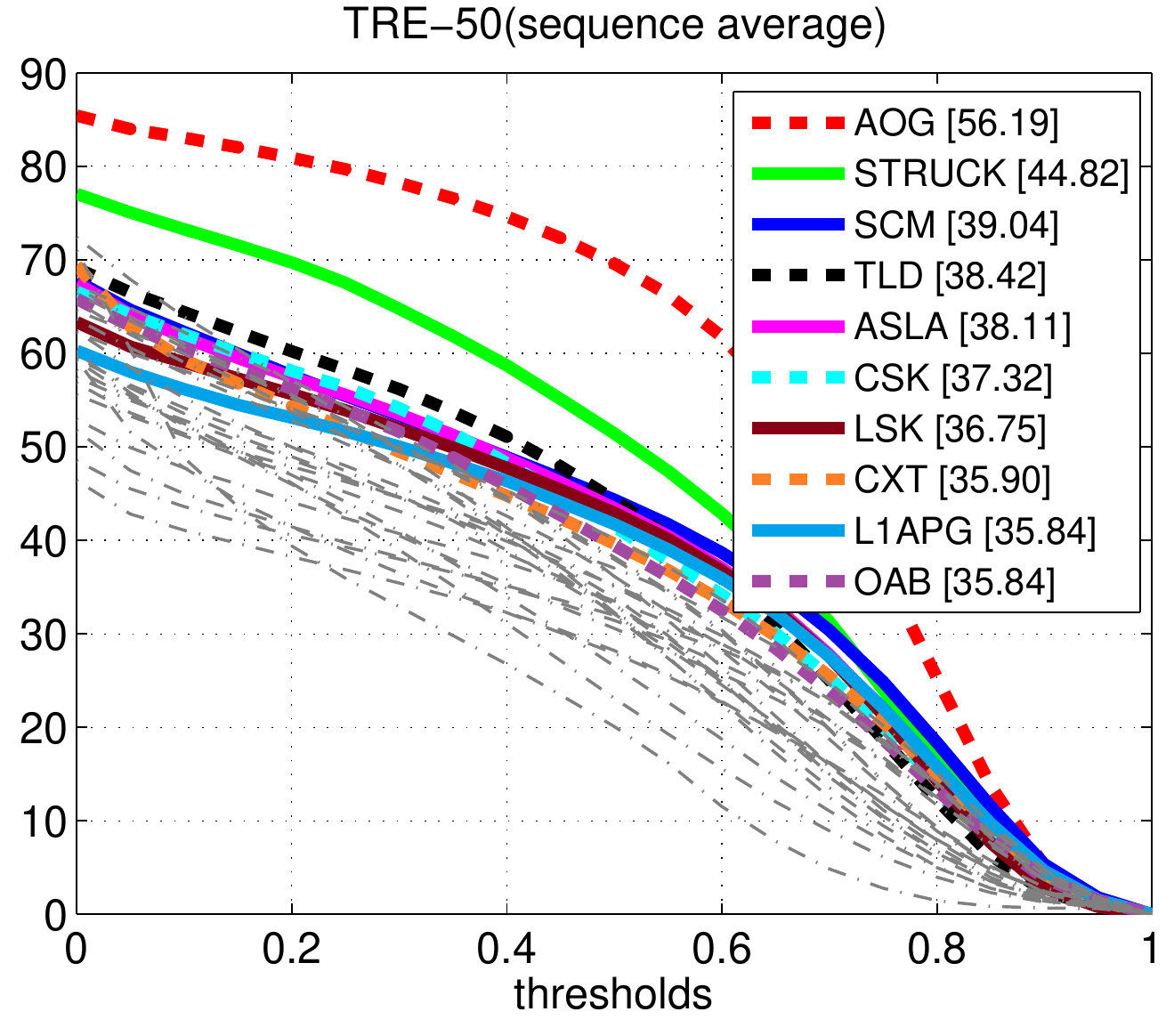}
		\end{subfigure}%
		\vspace{1mm}
		\begin{subfigure}[t]{0.32\textwidth}
			\centering
			\includegraphics[width=1.0\textwidth, height=0.8\textwidth]{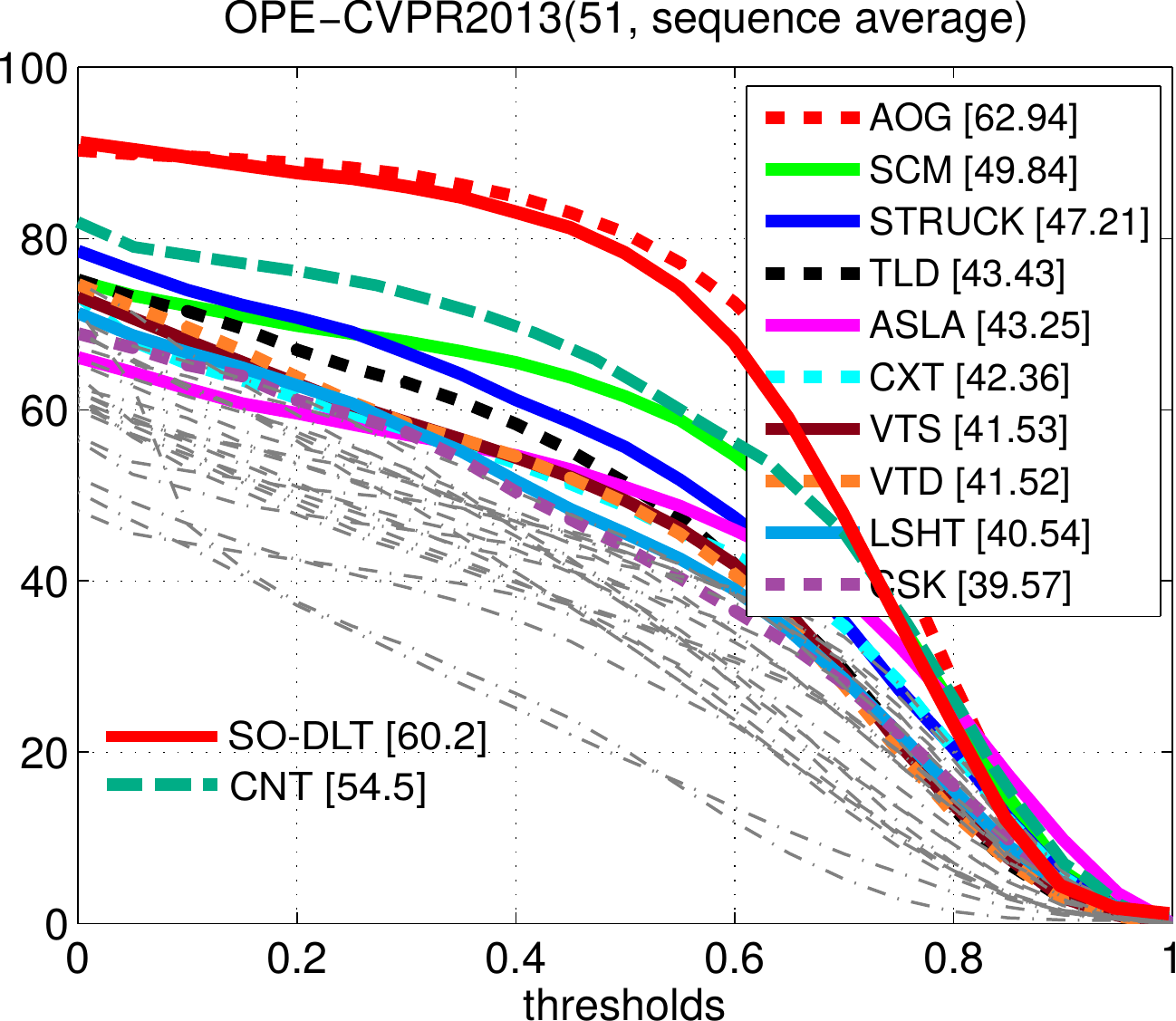}
		\end{subfigure}%
		~ 
		\begin{subfigure}[t]{0.32\textwidth}
			\centering
			\includegraphics[width=1.0\textwidth, height=0.8\textwidth]{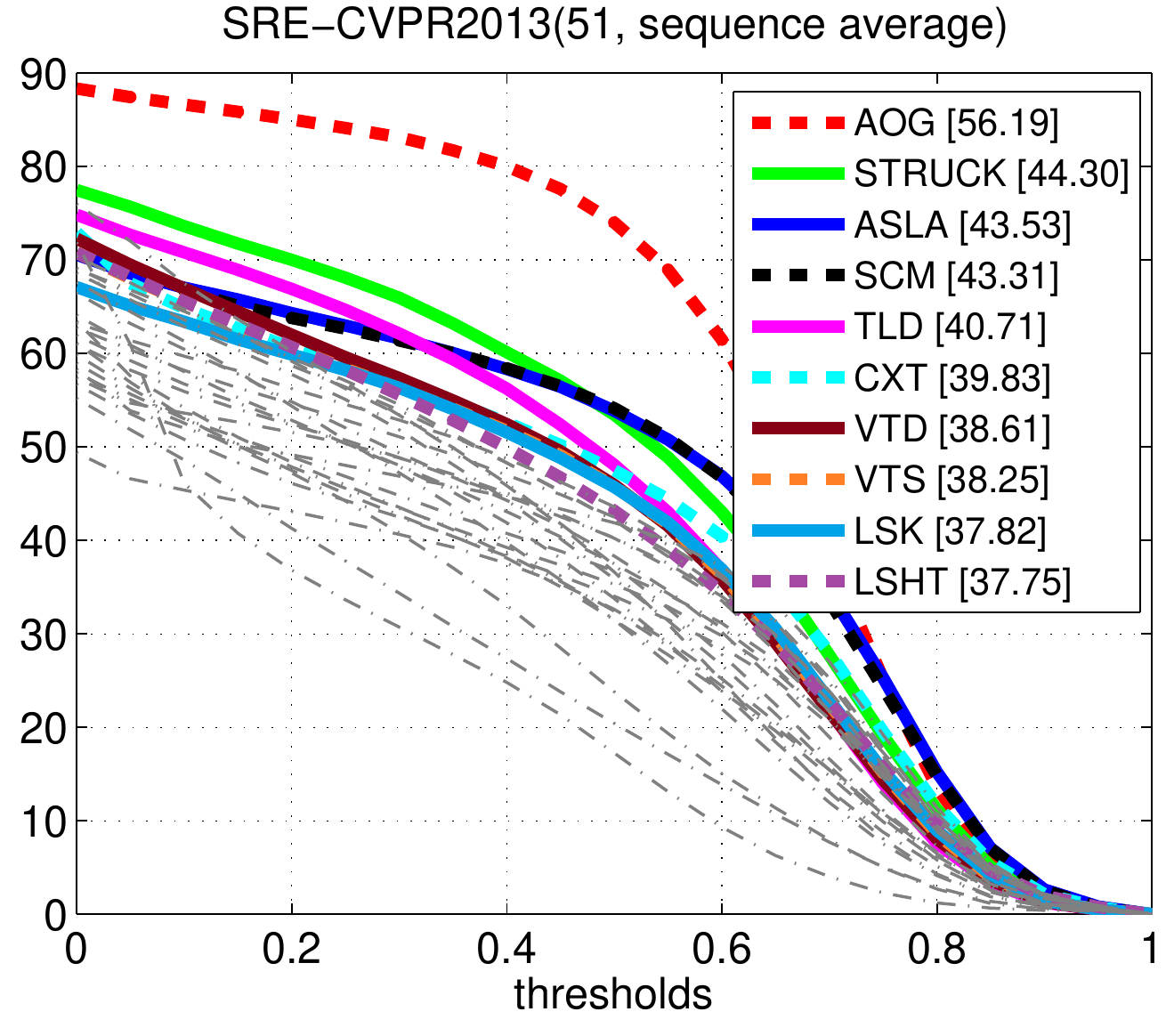}
		\end{subfigure}%
		~
		\begin{subfigure}[t]{0.32\textwidth}
			\centering
			\includegraphics[width=1.0\textwidth, height=0.8\textwidth]{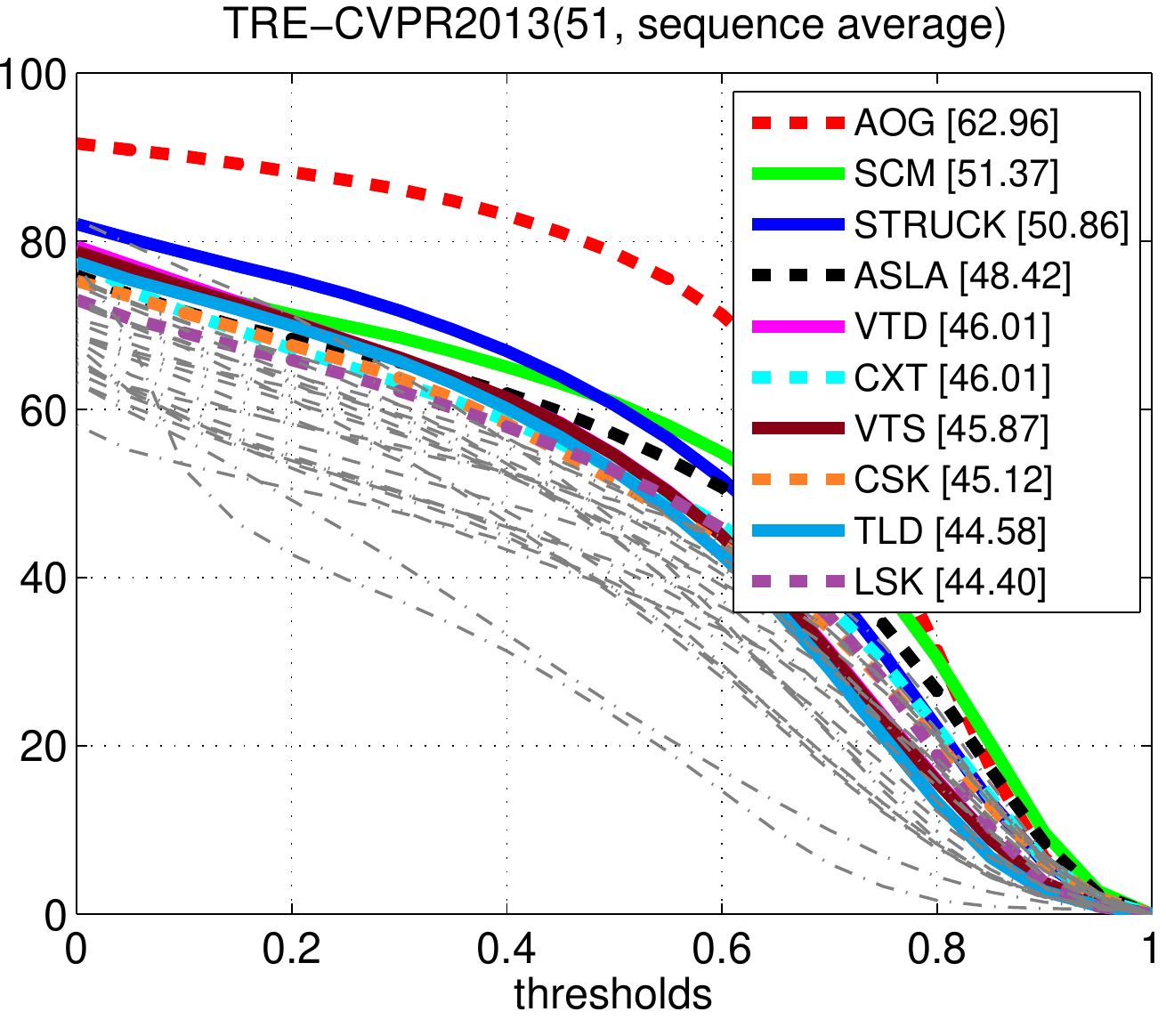}
		\end{subfigure}%
	\caption{Performance comparison in TB-100 (1st row), TB-50 (2nd row) and TB-CVPR2013 (3rd row) in term of success plots of OPE (1st column), SRE (2nd column) and TRE (3rd colum). For clarity, only top 10 trackers are shown in color curves and listed in the legend. Two deep learning based trackers, CNT\cite{cnnTracker} and SO-DLT\cite{rcnnTracker}, are evaluated in TB-CVPR2013 using OPE (with their performance plots manually added in the left-bottom figure). \textit{We note that the plots are reproduced with the raw results provided at http://cvlab.hanyang.ac.kr/tracker\_benchmark/.} (Best viewed in color and with magnification)} \label{fig:TB-AUC}
	%\vspace{-3mm} 
	% (x-axis represents intersection-over-union overlap thresholds and y-axis success rates) 
\end{figure*}

\begin{figure*}[!t]
	\centering
	\begin{subfigure}[t]{0.32\textwidth}
		\centering
		\includegraphics[width=1.0\textwidth, height=0.8\textwidth]{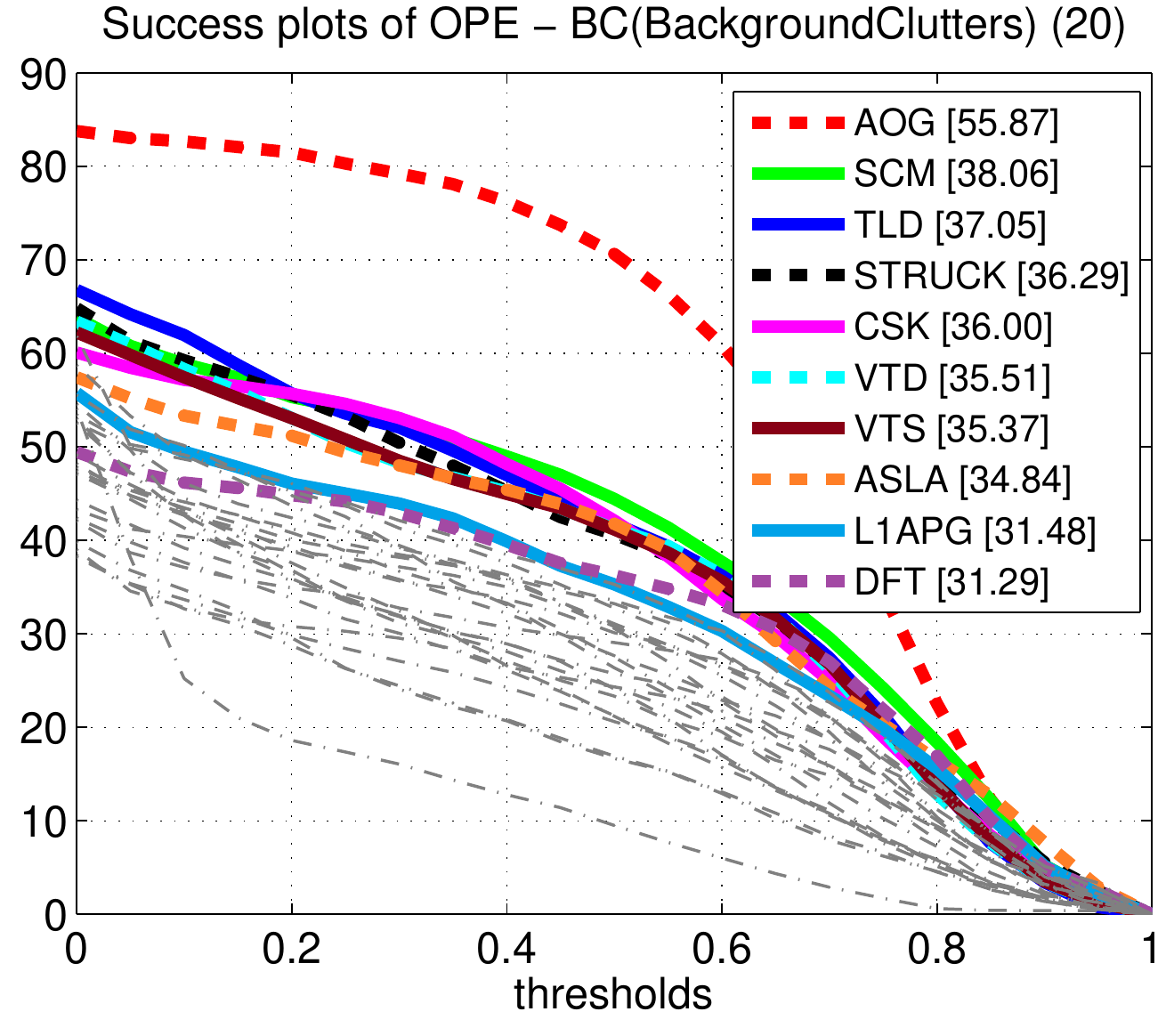}		
	\end{subfigure}%
	~ 
	\begin{subfigure}[t]{0.32\textwidth}
		\centering
		\includegraphics[width=1.0\textwidth, height=0.8\textwidth]{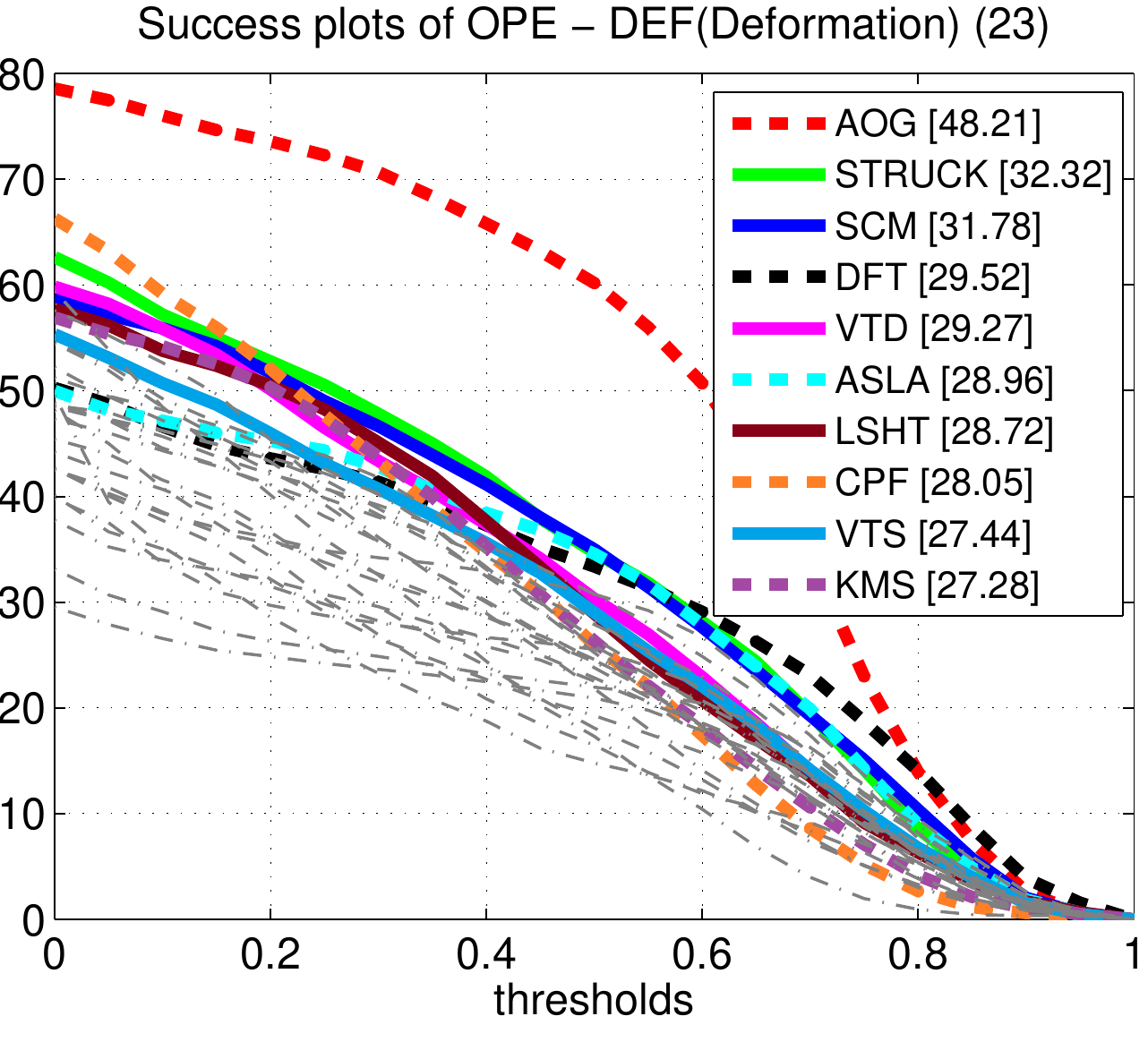}		
	\end{subfigure}%
	~
	\begin{subfigure}[t]{0.32\textwidth}
		\centering
		\includegraphics[width=1.0\textwidth, height=0.8\textwidth]{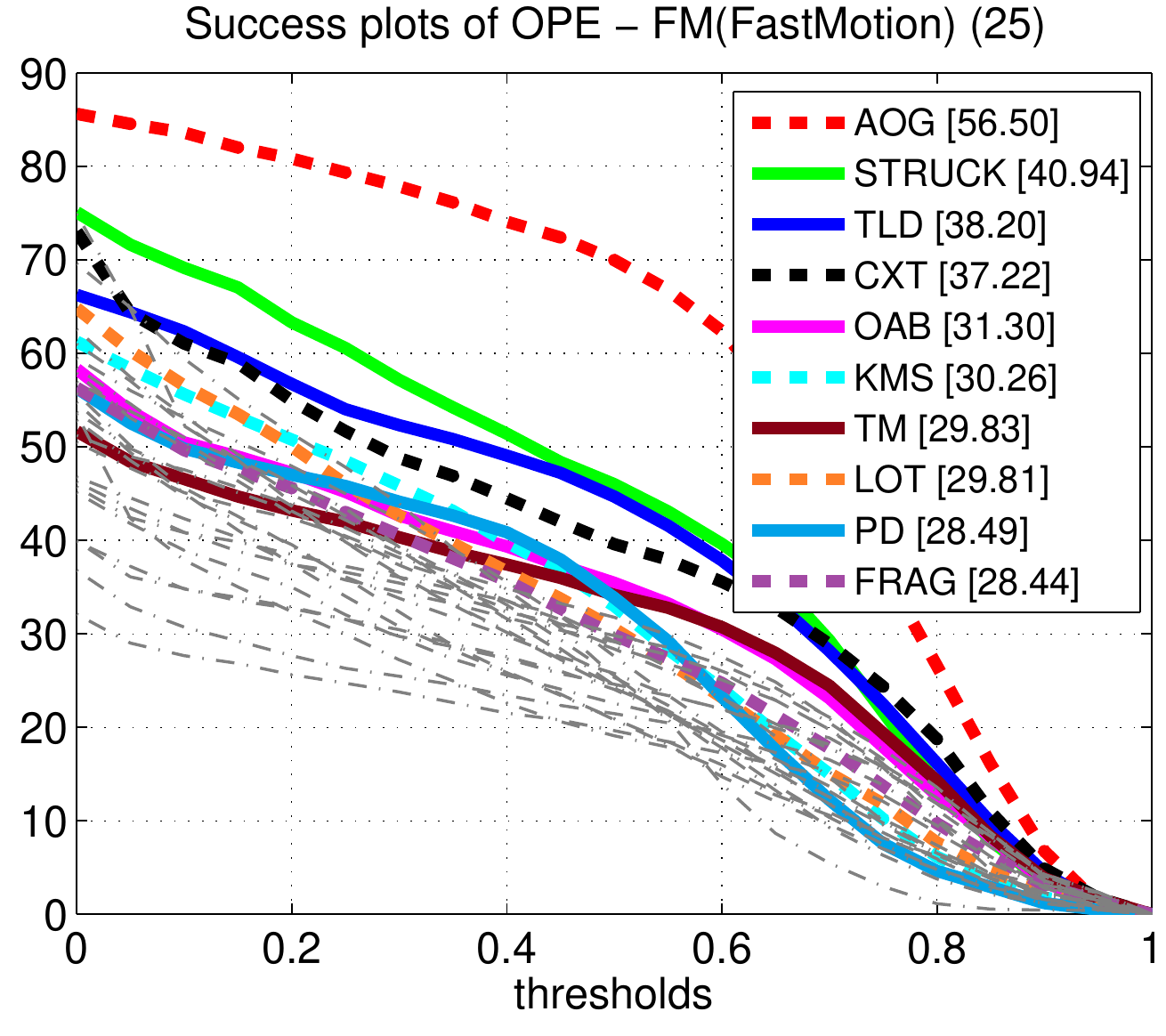}		
	\end{subfigure}%
	\vspace{1mm}
	\begin{subfigure}[t]{0.32\textwidth}
		\centering
		\includegraphics[width=1.0\textwidth, height=0.8\textwidth]{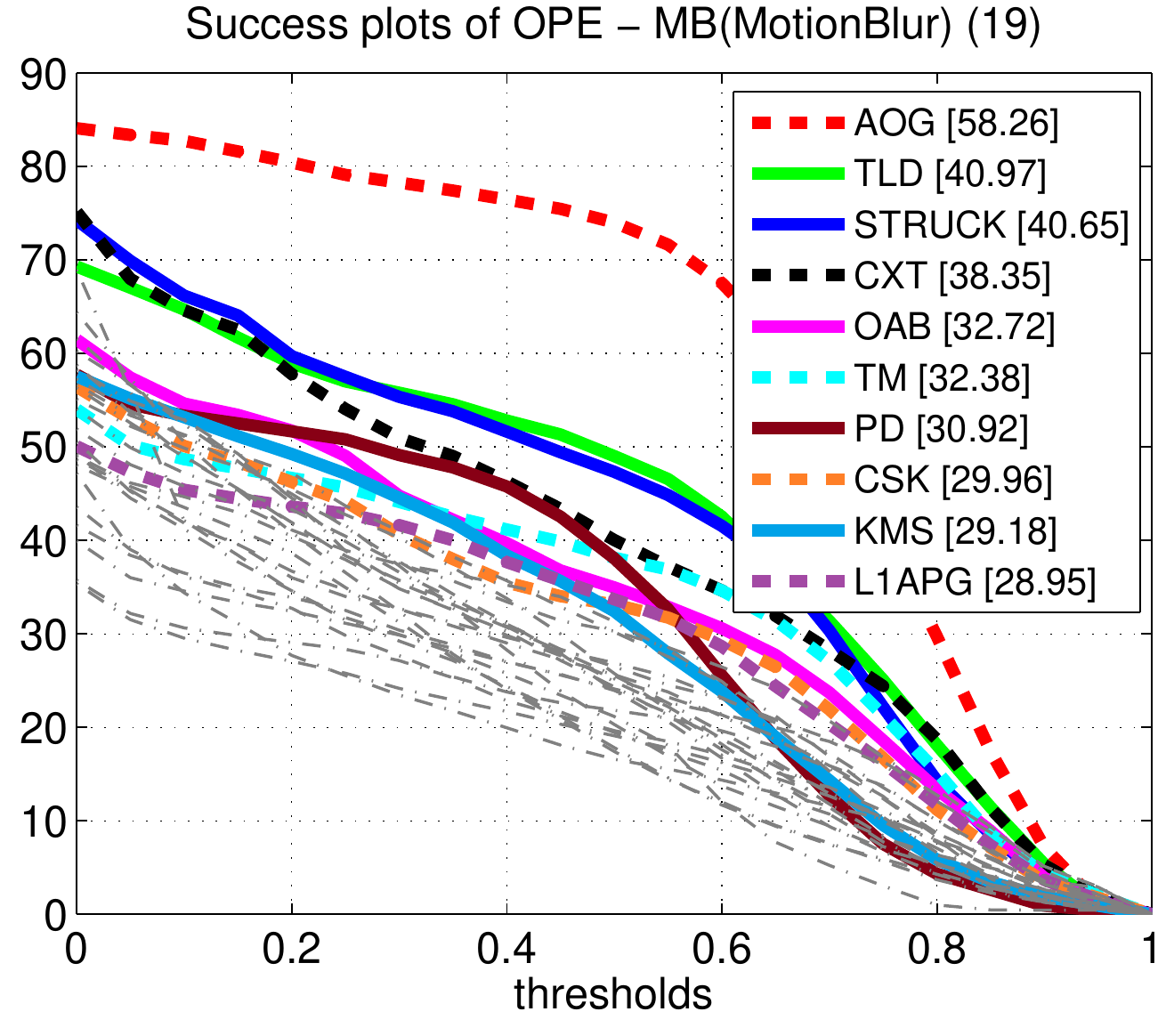}		
	\end{subfigure}%
	~ 
	\begin{subfigure}[t]{0.32\textwidth}
		\centering
		\includegraphics[width=1.0\textwidth, height=0.8\textwidth]{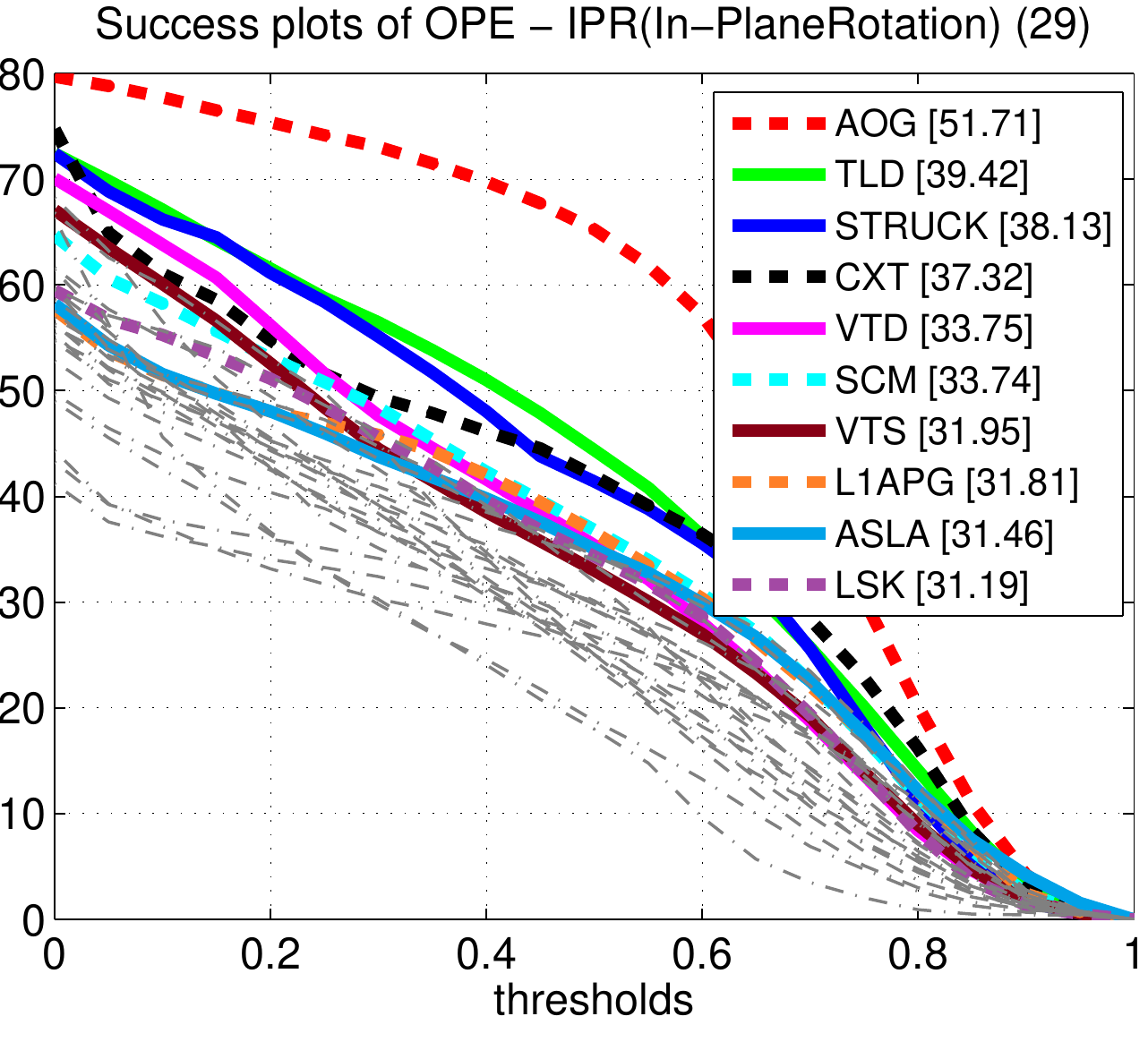}		
	\end{subfigure}%
	~
	\begin{subfigure}[t]{0.32\textwidth}
		\centering
		\includegraphics[width=1.0\textwidth, height=0.8\textwidth]{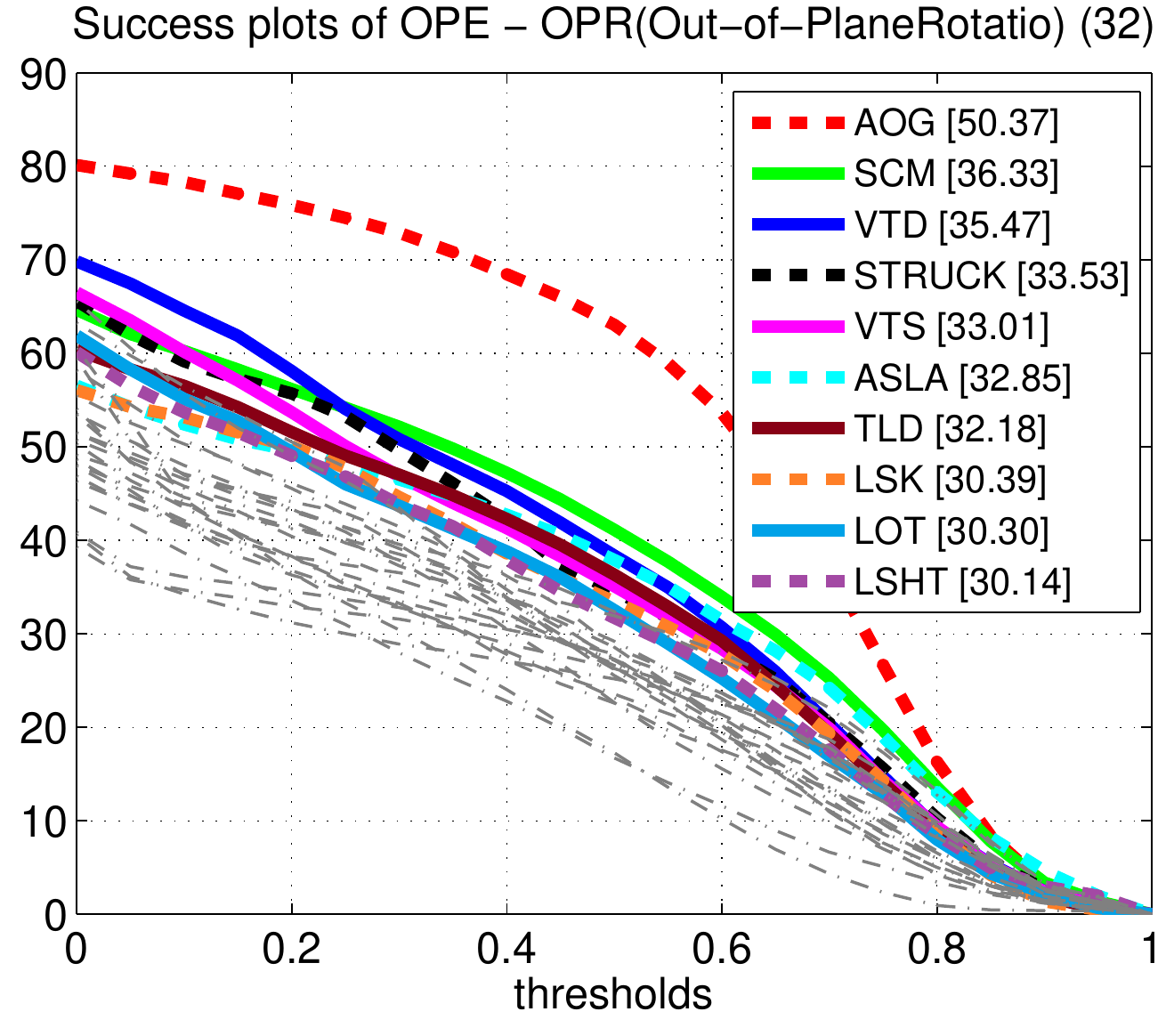}		
	\end{subfigure}%
	\vspace{1mm}
	\begin{subfigure}[t]{0.32\textwidth}
		\centering
		\includegraphics[width=1.0\textwidth, height=0.8\textwidth]{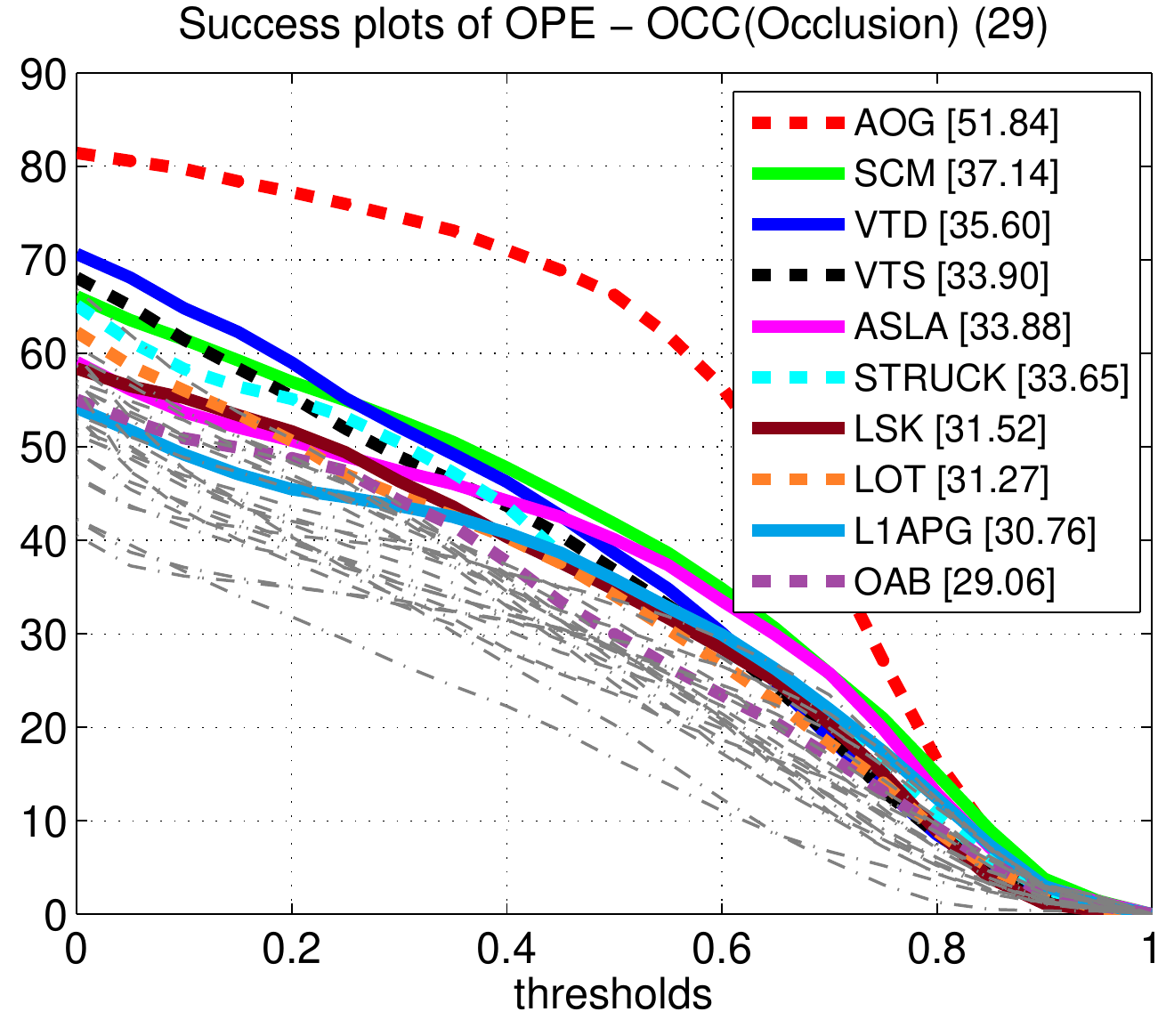}		
	\end{subfigure}%
	~ 
	\begin{subfigure}[t]{0.32\textwidth}
		\centering
		\includegraphics[width=1.0\textwidth, height=0.8\textwidth]{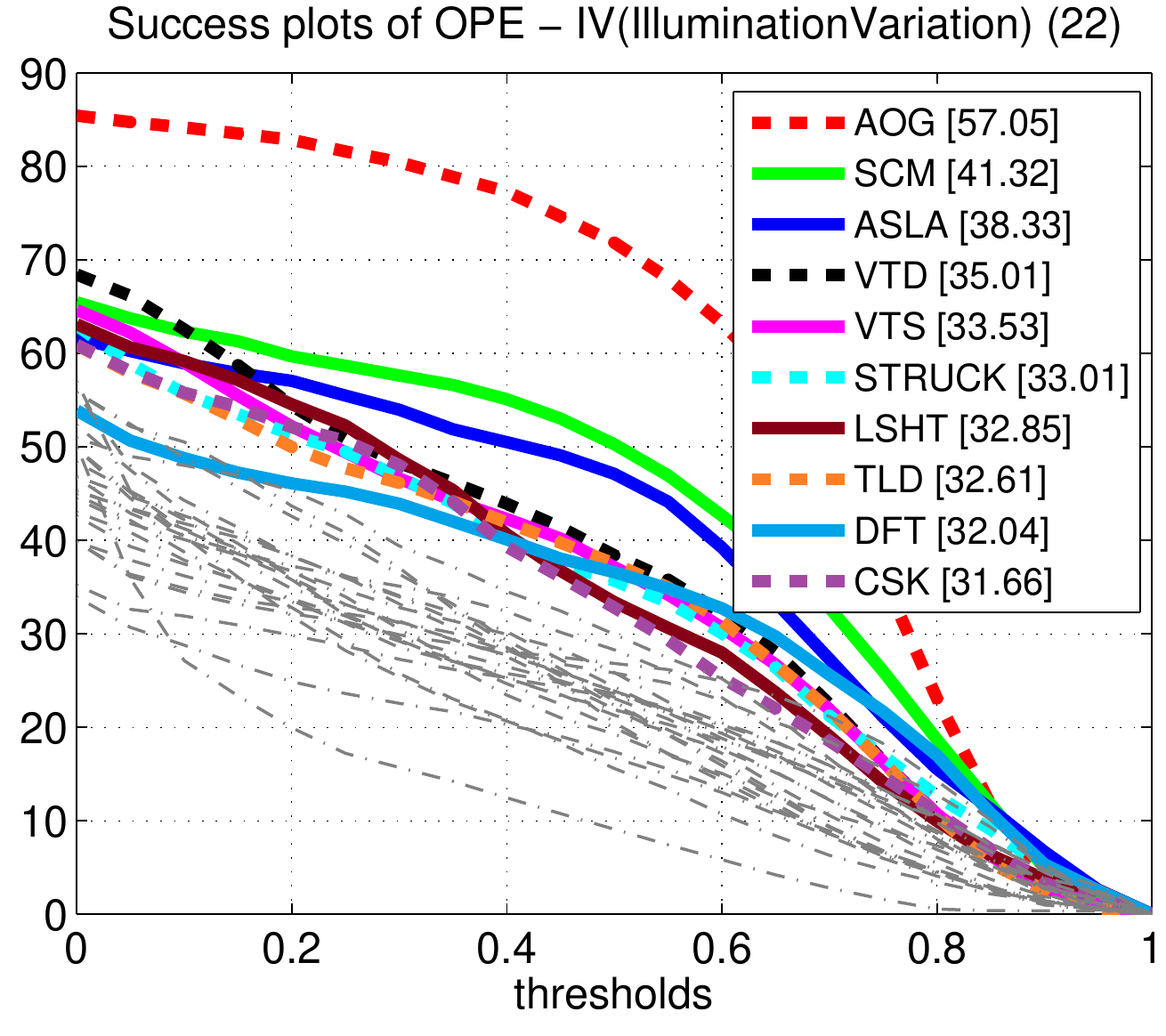}		
	\end{subfigure}%
	~
	\begin{subfigure}[t]{0.32\textwidth}
		\centering
		\includegraphics[width=1.0\textwidth, height=0.8\textwidth]{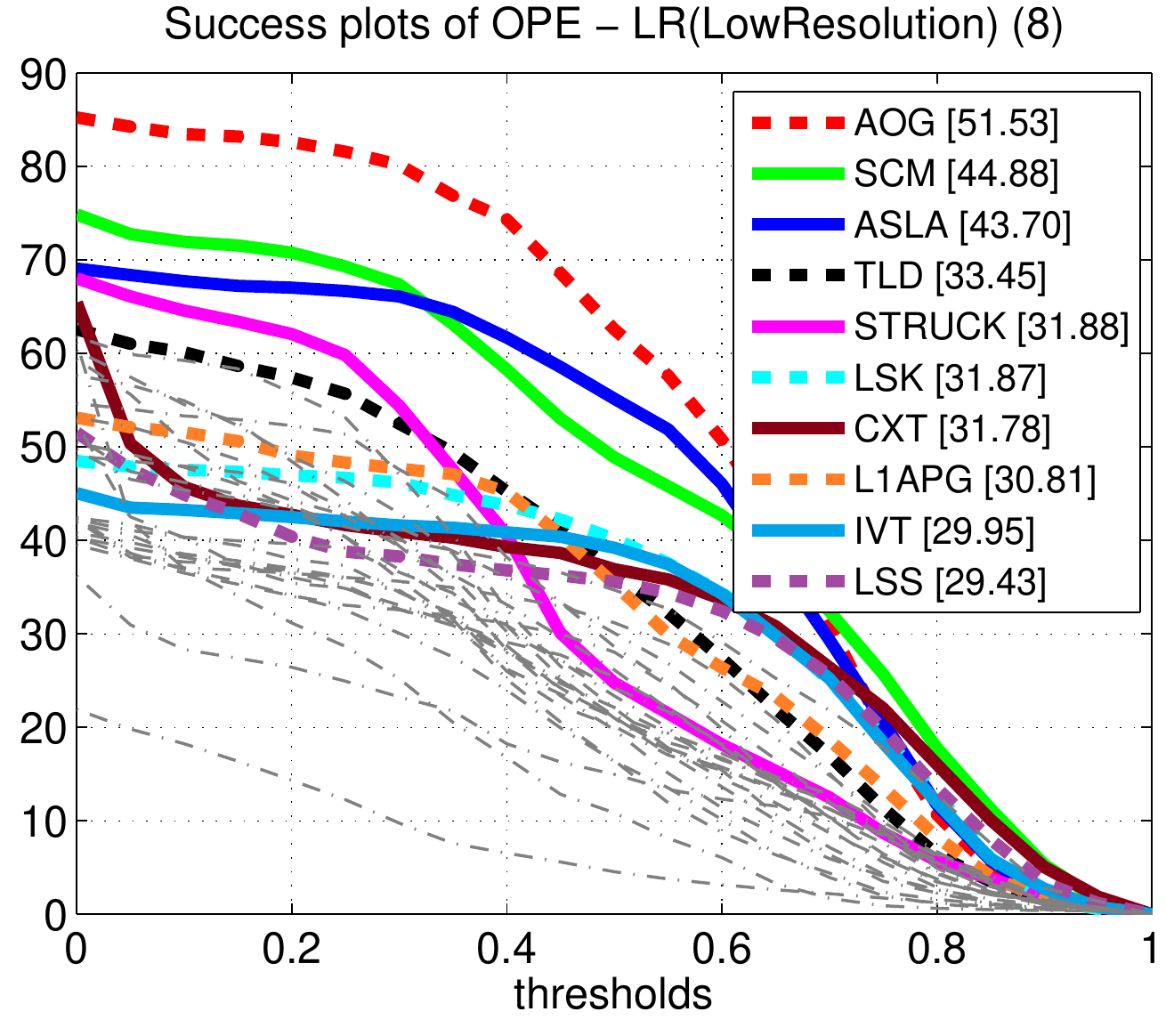}		
	\end{subfigure}%
	\vspace{1mm}
	\begin{subfigure}[t]{0.32\textwidth}
		\centering
		\includegraphics[width=1.0\textwidth, height=0.8\textwidth]{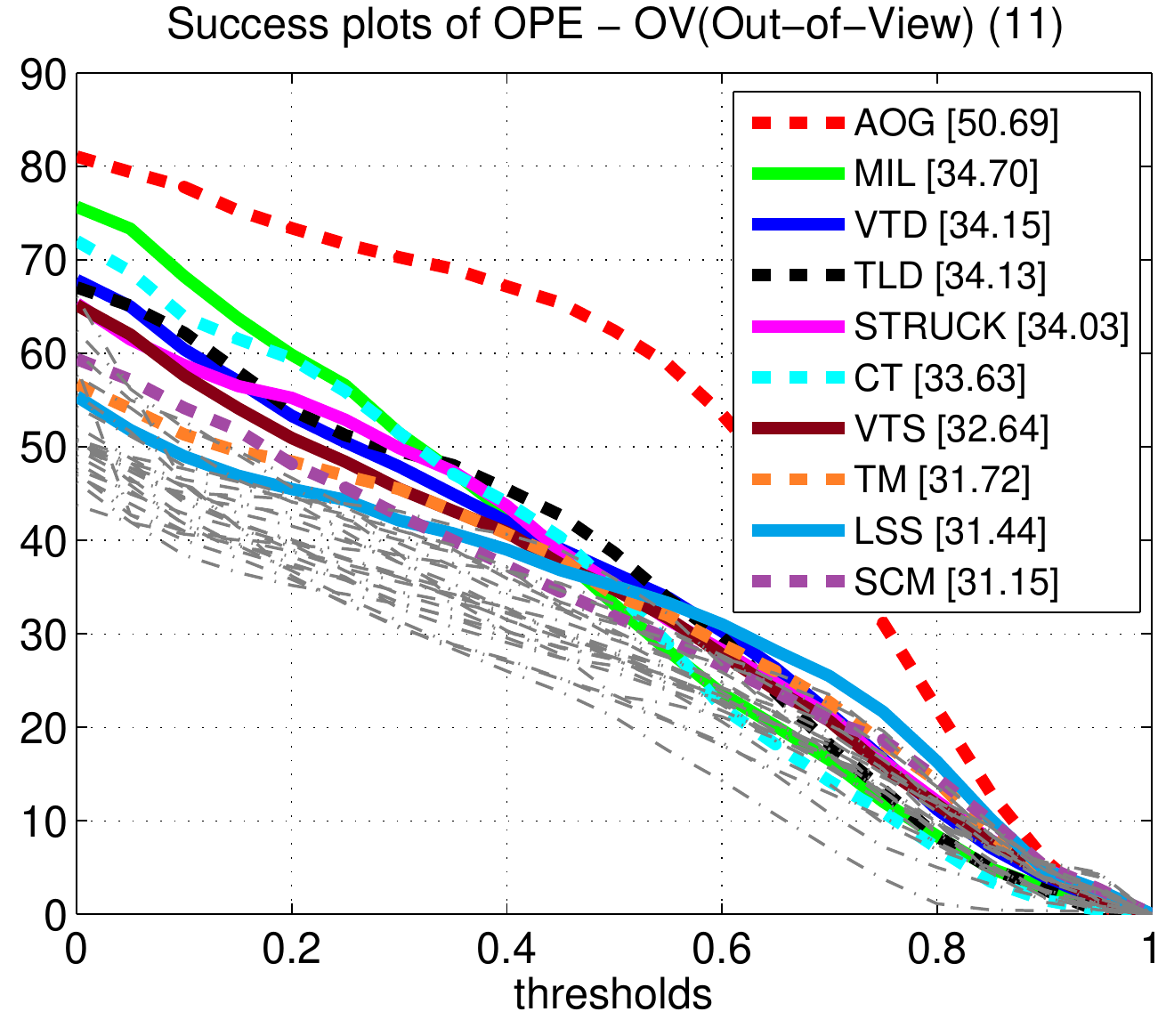}		
	\end{subfigure}%
	~ 
	\begin{subfigure}[t]{0.32\textwidth}
		\centering
		\includegraphics[width=1.0\textwidth, height=0.8\textwidth]{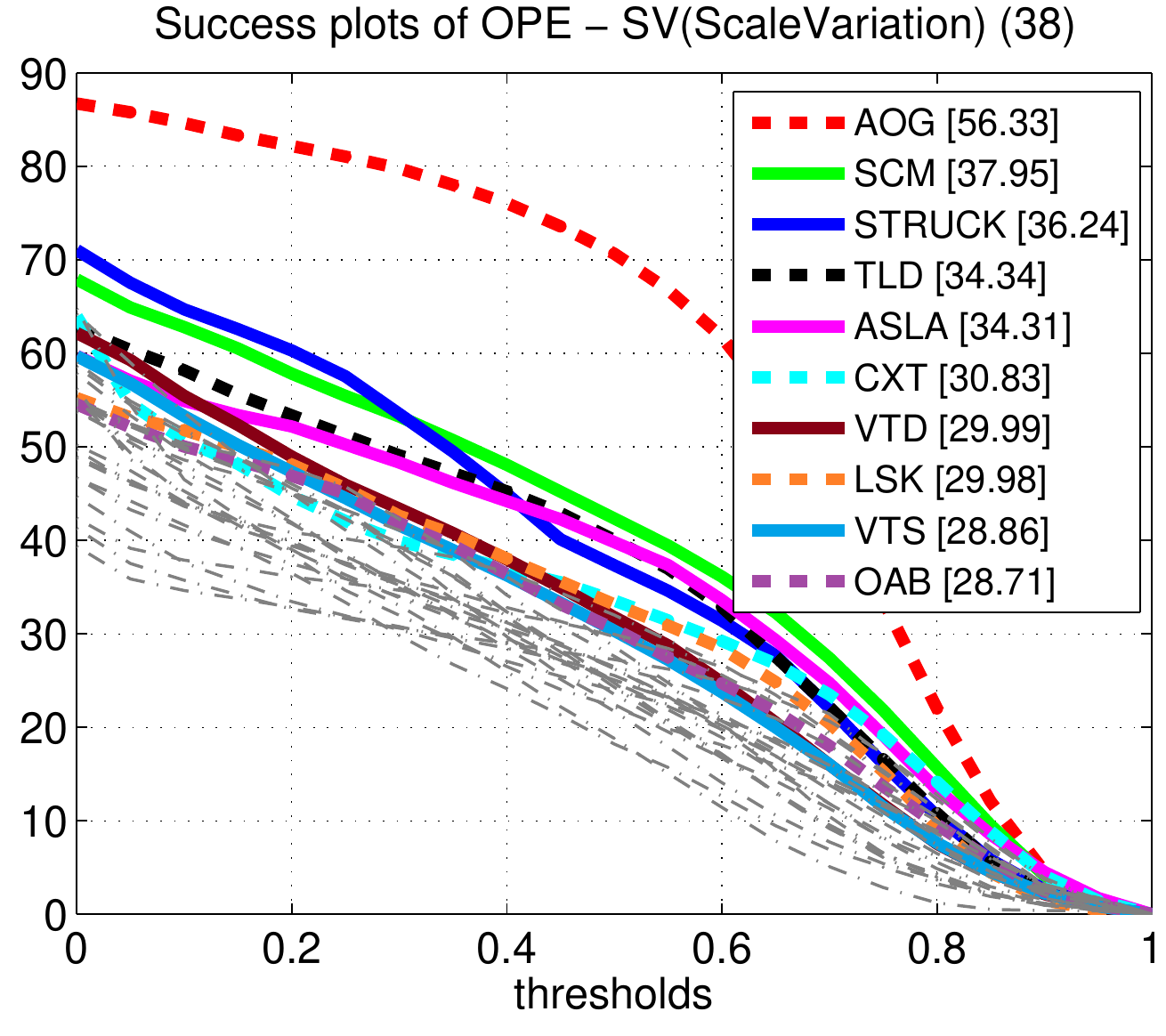}		
	\end{subfigure}%
%	~
%	\begin{subfigure}[t]{0.3\textwidth}
%		\centering
%		\includegraphics[width=0.6\textwidth]{Fig/resultFigs/attributes.pdf}		
%	\end{subfigure}%
	\caption{Performance comparison in the 11 subsets (with different attributes and different number of sequences as shown by the titles in the sub-figures) of TB-50 based on the success plots of OPE.} \label{fig:OPE-SbusetTB50}
\end{figure*}

 \begin{figure*}
 	\centering
 	\includegraphics[width=0.95\textwidth]{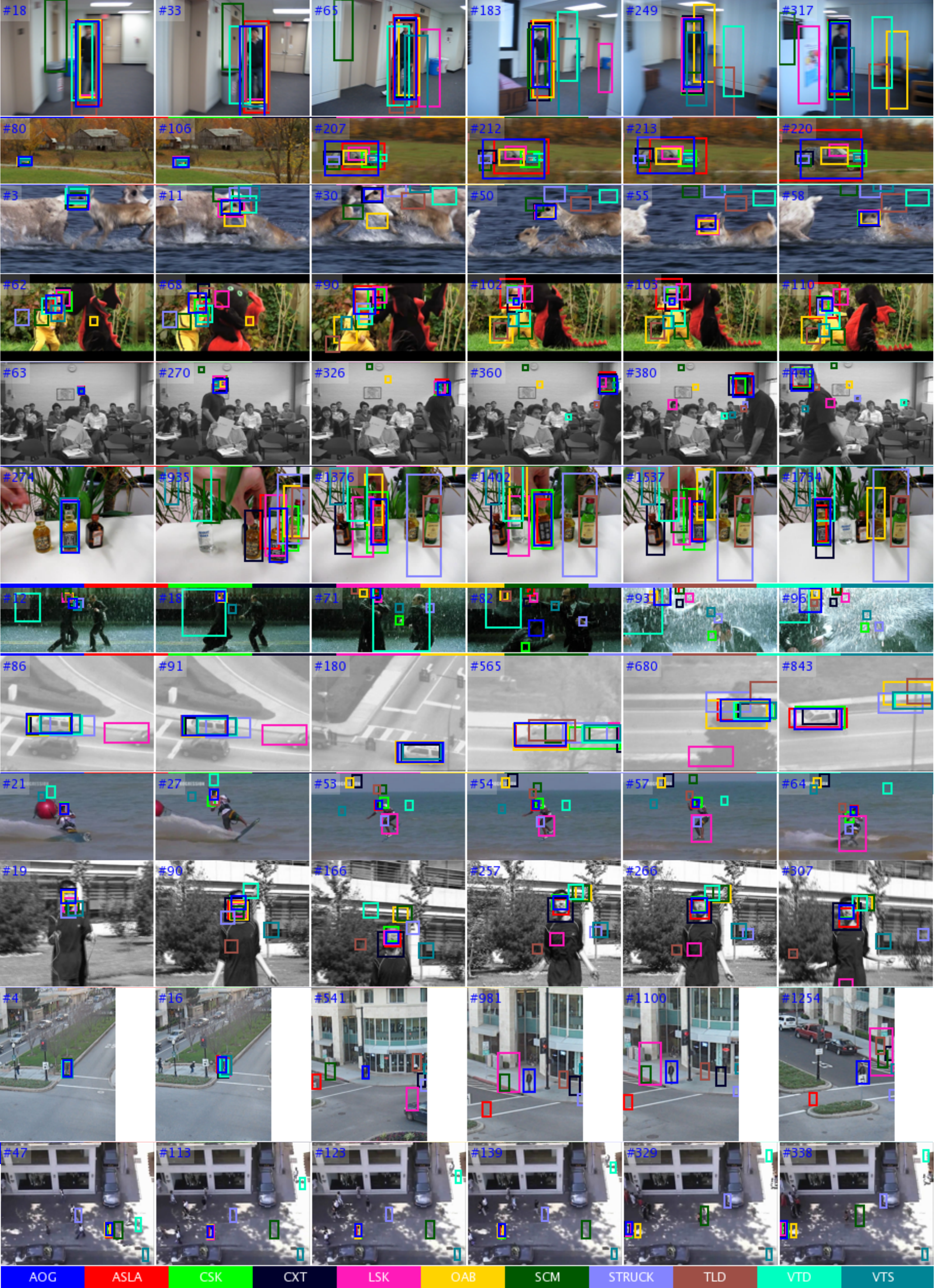}
 	\caption{Qualitative results. For clarity, we show tracking results (bounding boxes) in 6 randomly sampled frames for the top 10 trackers according to their OPE performance in TB-100. (Best viewed in color and with magnification.)}	
 	\label{fig:BBs}	
 	%\vspace{-5mm}
 \end{figure*}

%\vspace{-2mm}
\section{Experiments}\label{sec:exp}
In this section, we present comparison results on the TB-50/100/CVPR2013 benchmarks \cite{trackingBenchmarkPAMI,trackingBenchmark} and the VOT benchmarks \cite{VOT}. We also analyze different aspects of our method. 
The source code \footnote{Available at  https://github.com/tfwu/RGM-AOGTracker}  is released with this paper for reproducing all results. We denote the proposed method by \textit{AOG} in tables and plots.  

\textit{Parameter Setting.} We use the same parameters for all experiments since we emphasize online learning in this paper. In learning object AOGs, the side length of the grid used for constructing the full structure AOG is either $3$ or $4$ depending the slide length of input bounding box (to reduce the time complexity of online learning).  The number of intervals in computing feature pyramid is set to $6$ with cell size being $4$. The factor $s$ in computing search ROI is set to $s_{\text{\tiny{ROI}}}=3$. The NMS IoU threshold is set to $\tau_{\text{\tiny{NMS}}}=0.7$. The number of top parse trees kept after spatial DP parsing is set $N_{\text{\tiny{Best}}}=10$. The time range in temporal DP algorithm is set to $\Delta t=5$. In identifying critical moments, we set $N_{\text{\tiny{Intrackable}}}=5$ and $N_{\text{\tiny{NewSample}}}=10$. The LSVM trade-off parameter in Eqn.(\ref{eqn:loss}) is set to $C=0.001$. 
When re-learning structure and parameters, we could use all the frames with valid tracking results. To reduce the time complexity, the number of frames used in re-learning is at most $100$ in our experiments. At time $t$, we first take the first $10$ frames with valid tracking results in $[1, t]$ with the underlying intuition that they have high probabilities of being tracked correctly (note that we alway use the first frame since the ground-truth bounding box is given), and then take the remaining frames in reversed time order.

\textit{Speed.} In our current c++ implementation, we adopt FFT in computing score pyramids as done in \cite{FFLD} which also utilizes multi-threads with OpenMP. We also provide a distributed version based on MPI~\footnote{https://www.mpich.org/}  in evaluation. The FPS is about 2 to 3. We are experimenting GPU implementations to speed up our TLP.

    \begin{figure*} 
           	\centering
           	\begin{subfigure}[t]{0.32\textwidth}
           		\centering
           		\includegraphics[width=1.0\textwidth, height=0.8\textwidth] {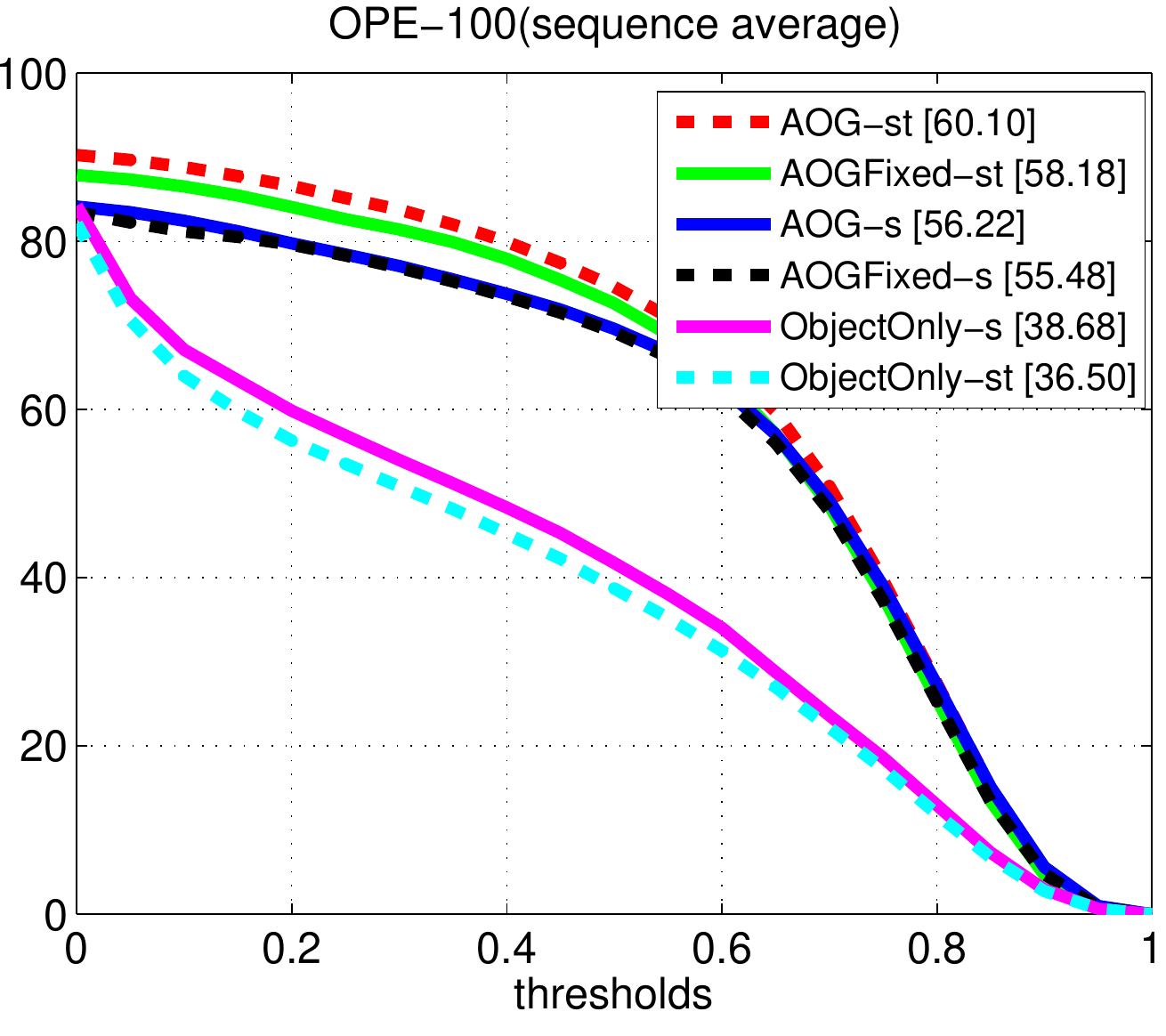}
           	\end{subfigure}%
           	~ 
           	\begin{subfigure}[t]{0.32\textwidth}
           		\centering
           		\includegraphics[width=1.0\textwidth, height=0.8\textwidth]{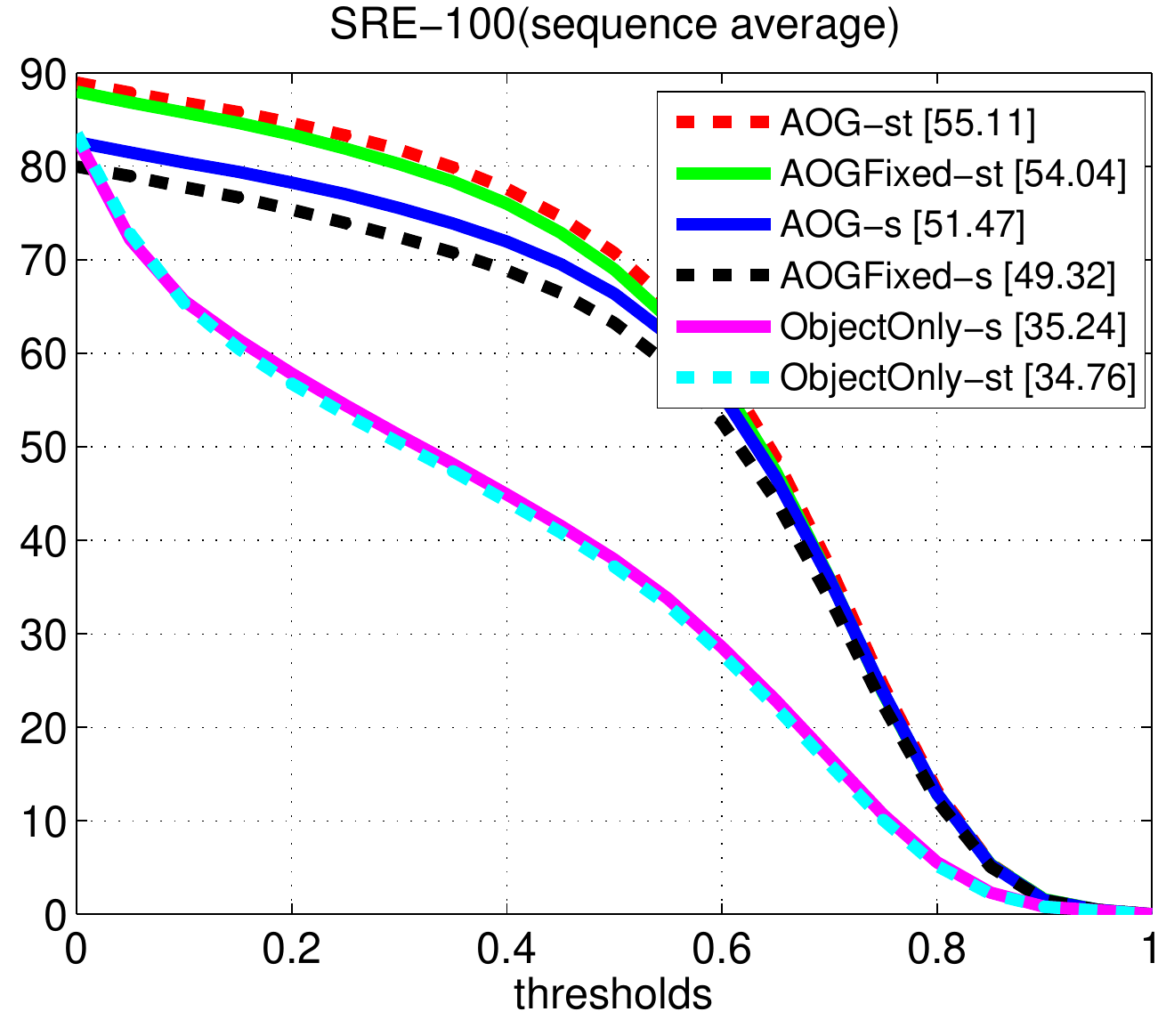}
           	\end{subfigure}%
           	~
           	\begin{subfigure}[t]{0.32\textwidth}
           		\centering
           		\includegraphics[width=1.0\textwidth, height=0.8\textwidth]{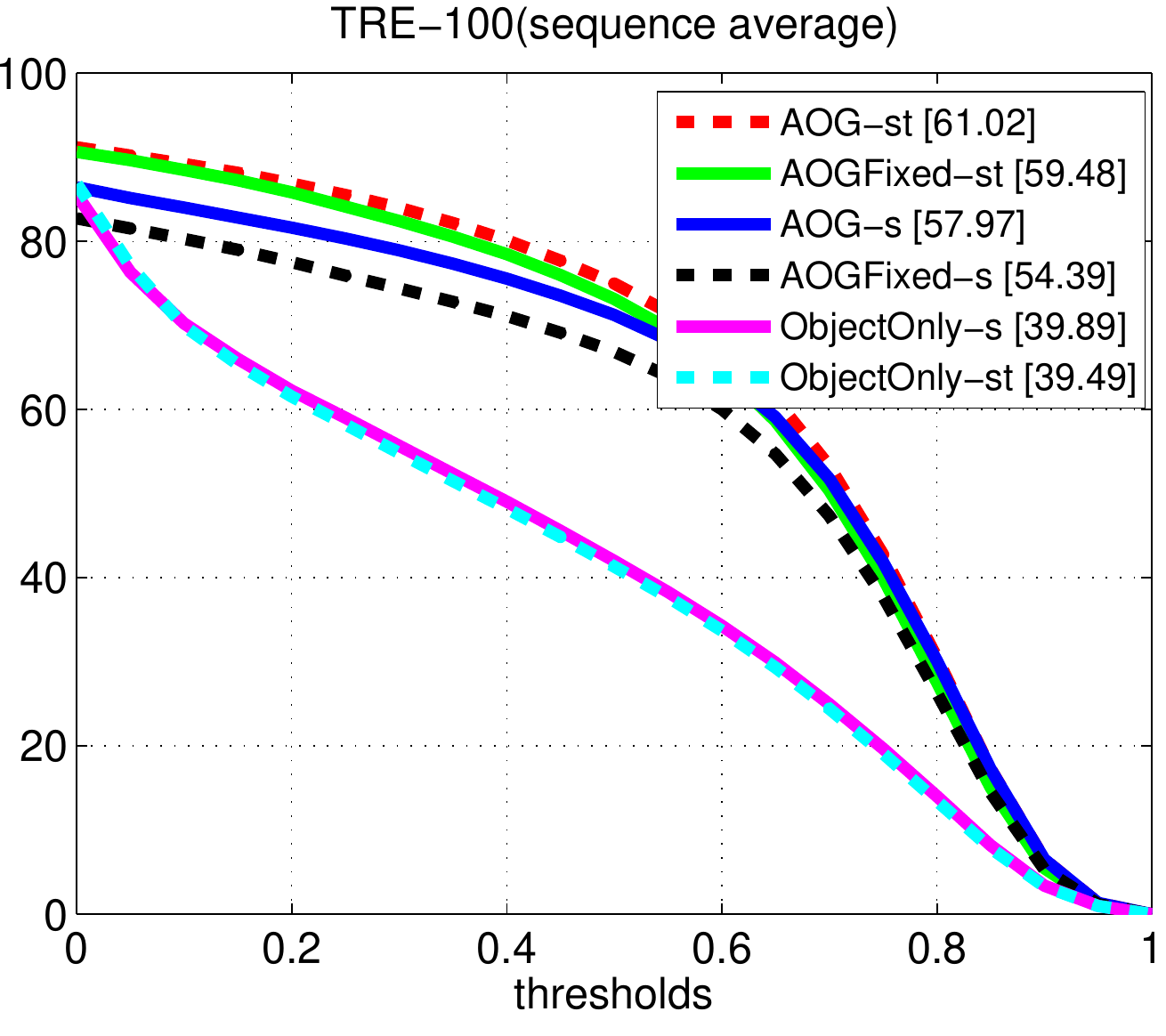}
           	\end{subfigure}%
           	\vspace{1mm}
           	\begin{subfigure}[t]{0.32\textwidth}
           		\centering
           		\includegraphics[width=1.0\textwidth, height=0.8\textwidth]{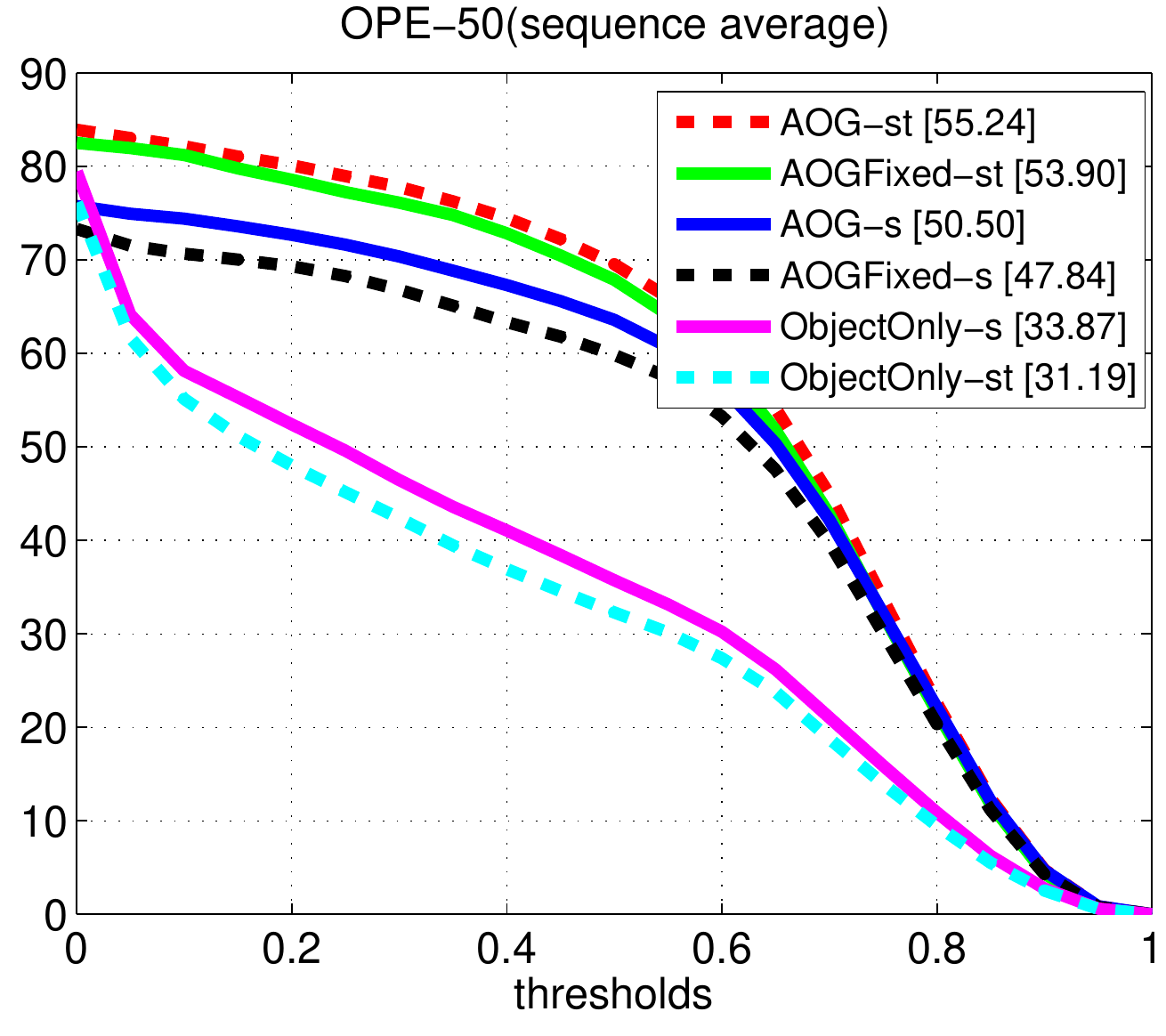}
           	\end{subfigure}%
           	~ 
           	\begin{subfigure}[t]{0.32\textwidth}
           		\centering
           		\includegraphics[width=1.0\textwidth, height=0.8\textwidth]{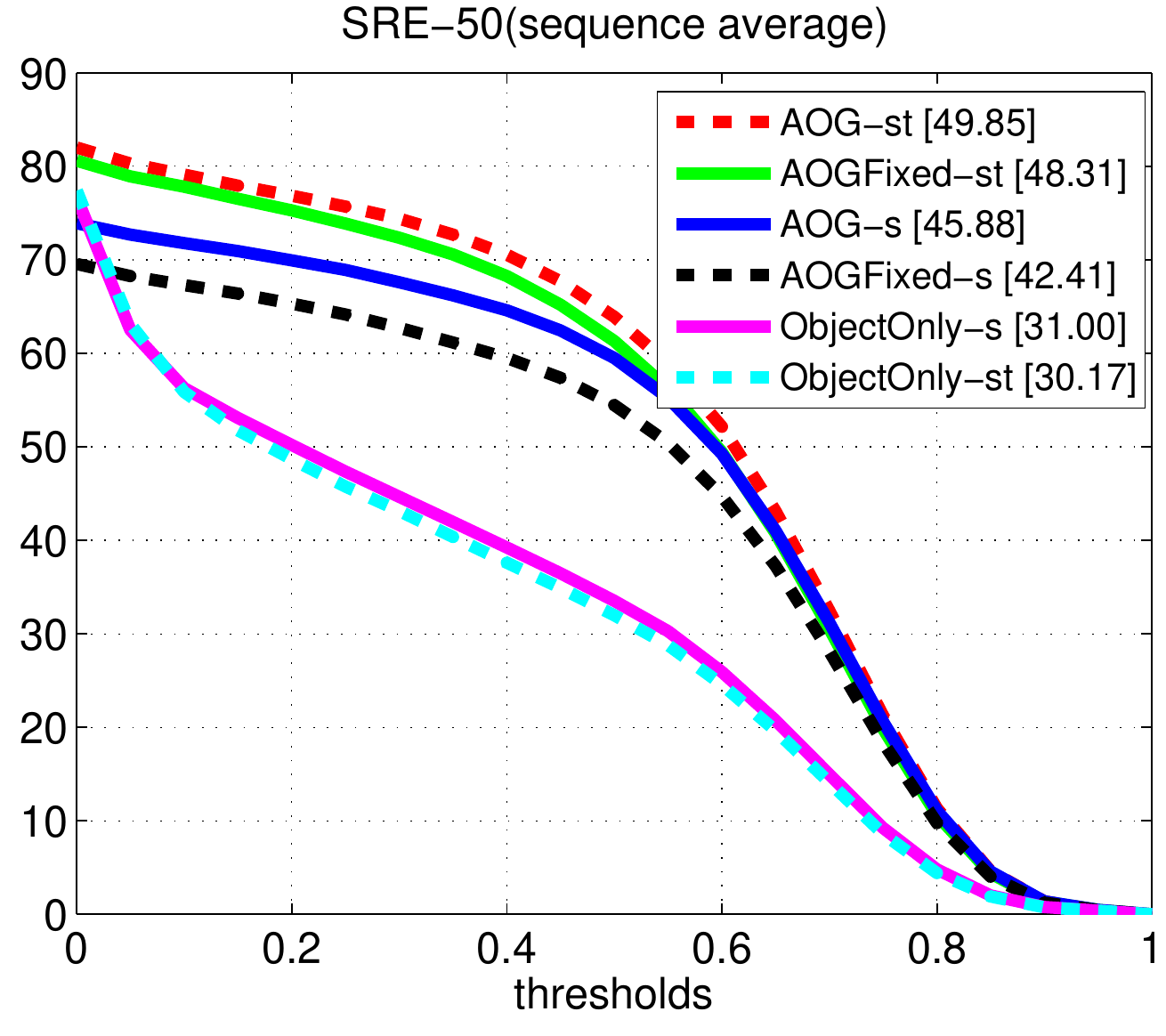}
           	\end{subfigure}%
           	~
           	\begin{subfigure}[t]{0.32\textwidth}
           		\centering
           		\includegraphics[width=1.0\textwidth, height=0.8\textwidth]{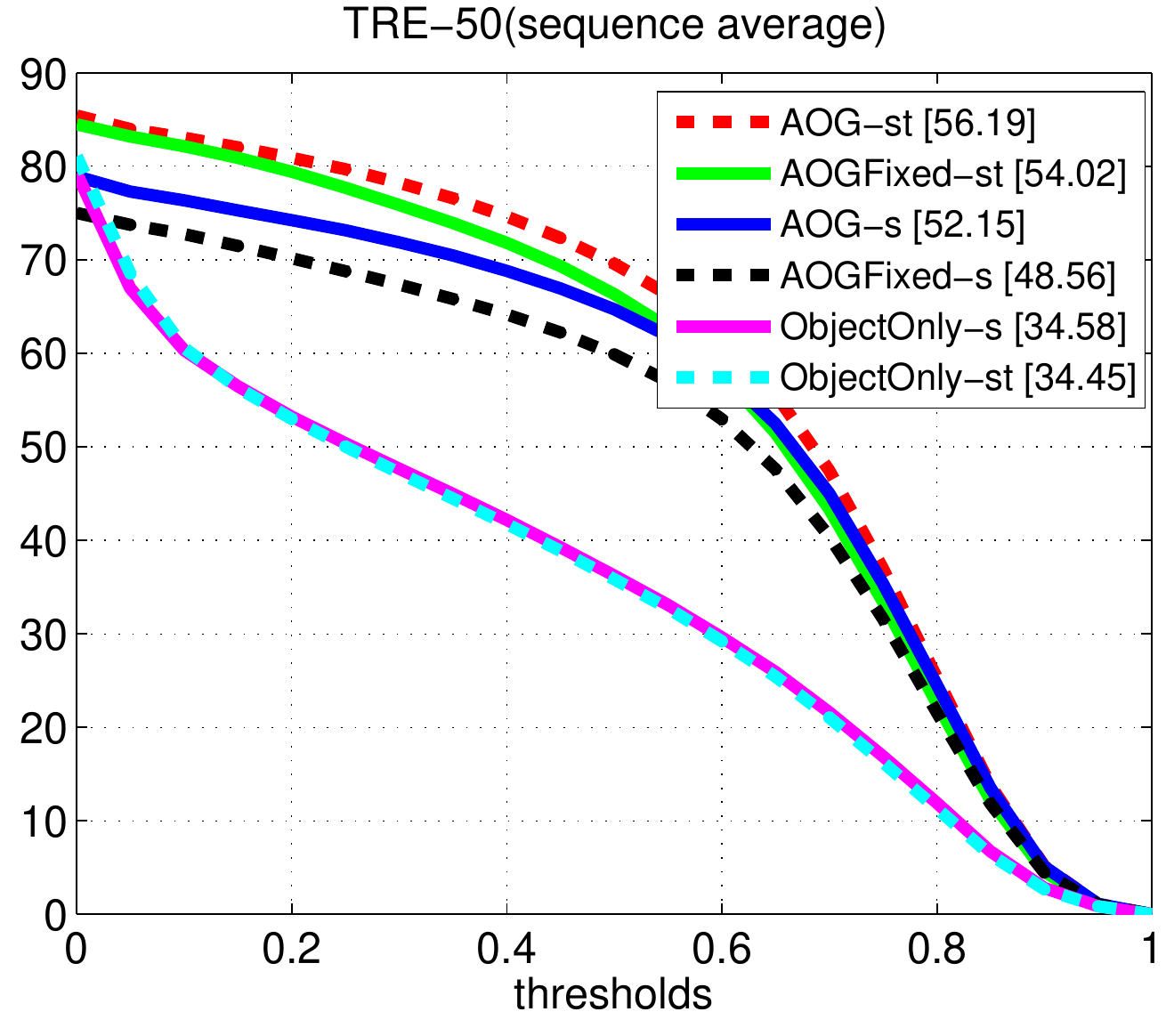}
           	\end{subfigure}%
           	\vspace{1mm}
           	\begin{subfigure}[t]{0.32\textwidth}
           		\centering
           		\includegraphics[width=1.0\textwidth, height=0.8\textwidth]{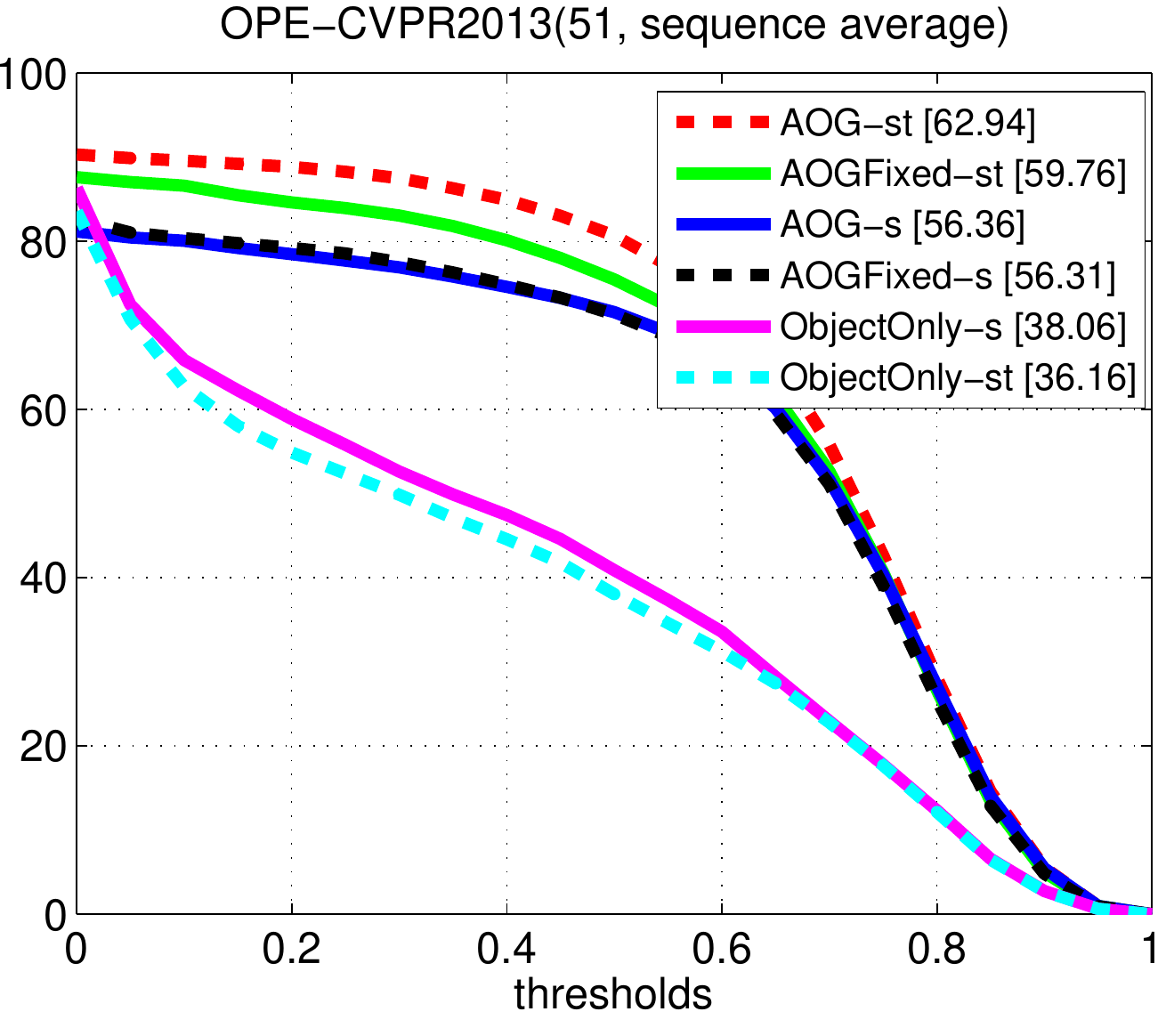}
           	\end{subfigure}%
           	~ 
           	\begin{subfigure}[t]{0.32\textwidth}
           		\centering
           		\includegraphics[width=1.0\textwidth, height=0.8\textwidth]{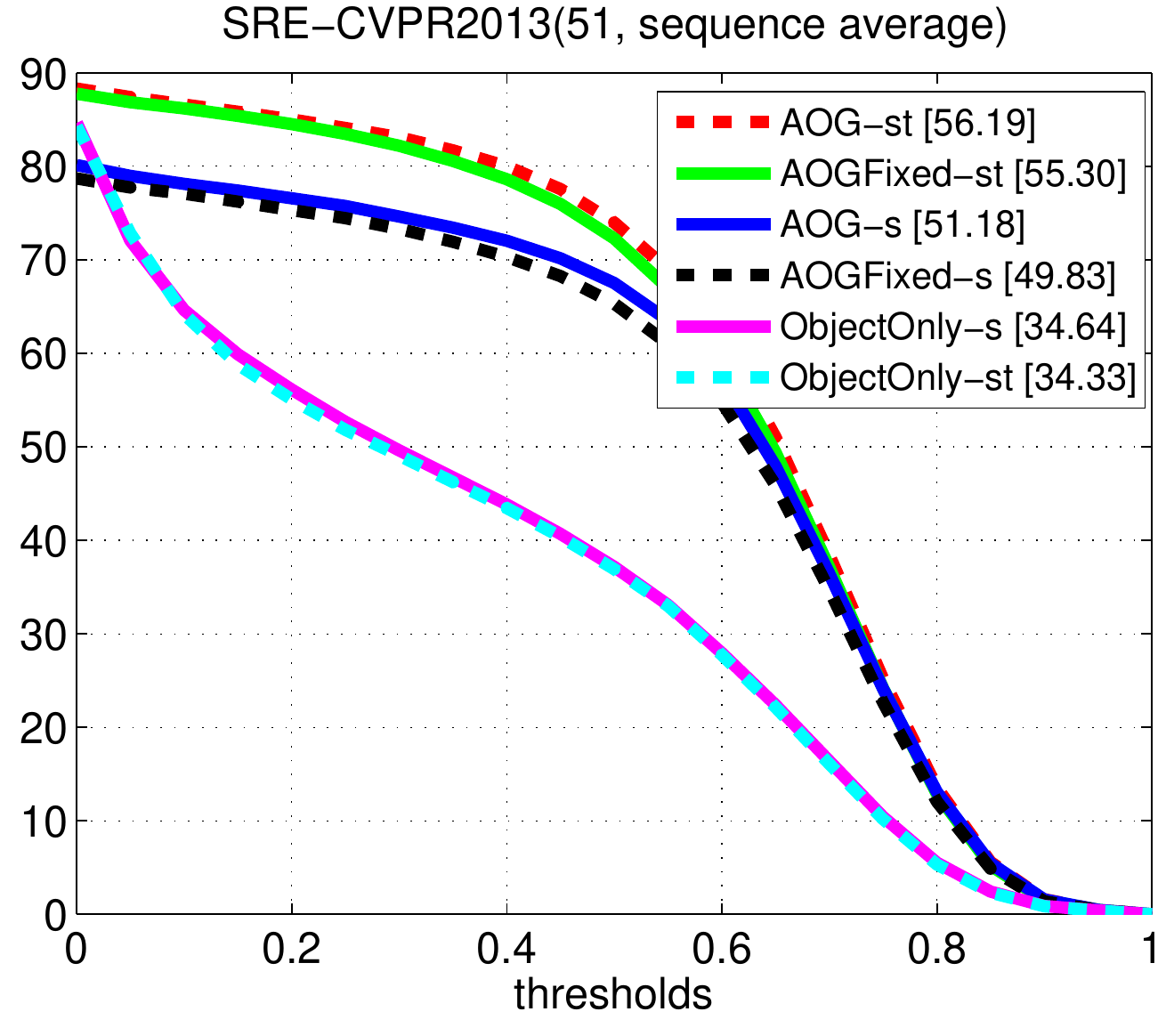}
           	\end{subfigure}%
           	~
           	\begin{subfigure}[t]{0.32\textwidth}
           		\centering
           		\includegraphics[width=1.0\textwidth, height=0.8\textwidth]{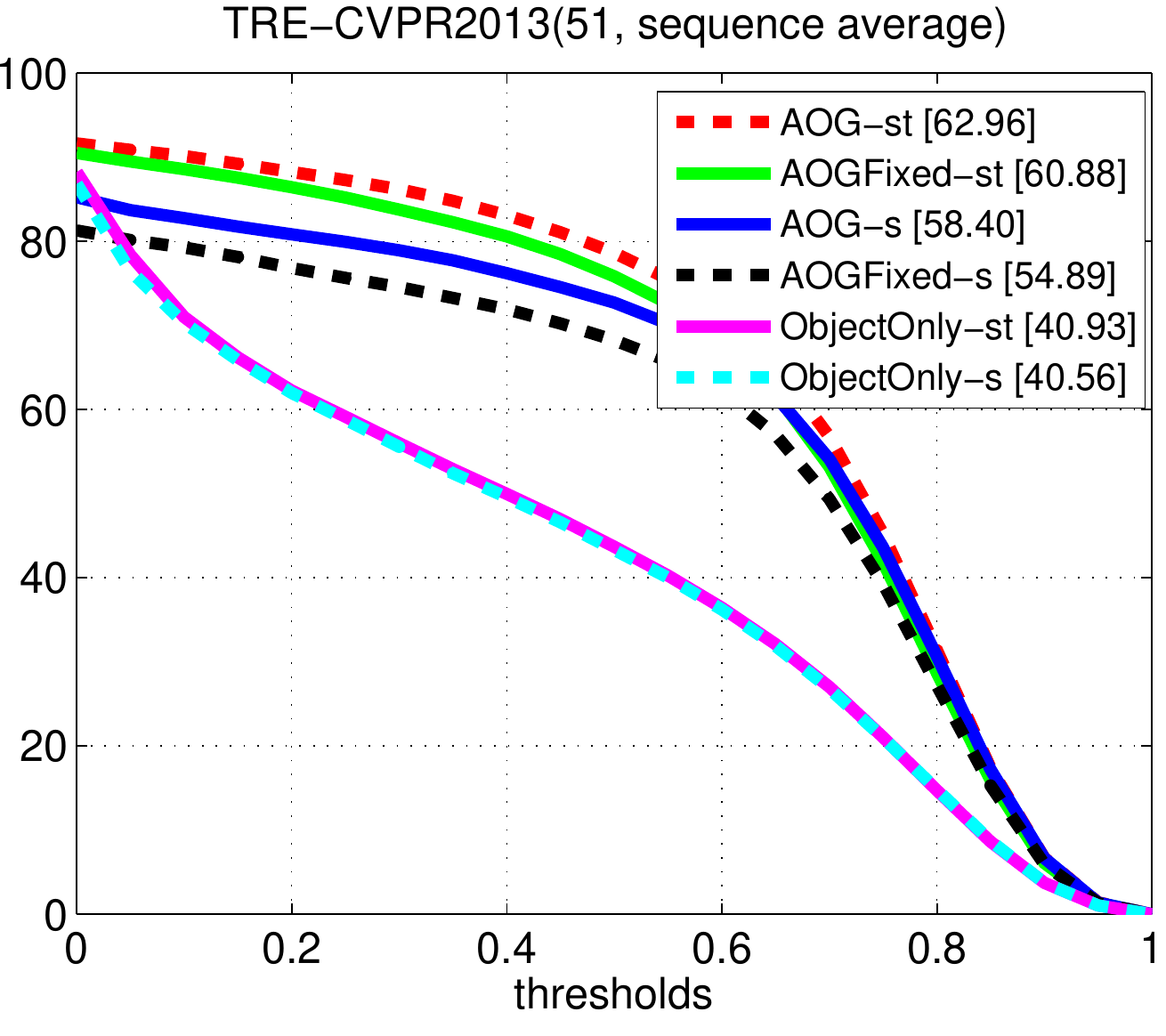}
           	\end{subfigure}%
           	\caption{Performance comparison of the six variants of our AOGTracker in TB-100/50/CVPR2013 in term of the success plots of OPE (1st column), SRE (2nd column) and TRE (3rd colum). } \label{fig:TB-AUC-AOGversions}
           	
           	% (x-axis represents intersection-over-union overlap thresholds and y-axis success rates) 
           \end{figure*} 

%\vspace{-3mm}
\subsection{Results on TB-50/100/CVPR2013}
 The TB-100 benchmark has 100 target objects ($58,897$ frames in total) with 29 publicly available trackers evaluated. It is extended from a previous benchmark with 51 target objects released at CVPR2013 (denoted by TB-CVPR2013). Further, since some target objects are similar or less challenging, a subset of 50 difficult and representative ones (denoted by TB-50) is selected for an in-depth analysis. Two types of performance metric are used, the \textbf{precision plot} (i.e., the percentage of frames in which estimated locations are within a given threshold distance of ground-truth positions) and the \textbf{success plot} (i.e., based on IoU overlap scores which are commonly used in object detection benchmarks, e.g., PASCAL VOC \cite{VOC}). The higher a success rate or a precision rate is, the better a tracker is. Usually, success plots are preferred to rank trackers \cite{trackingBenchmarkPAMI,VOT} (thus we focus on success plots in comparison). Three types of evaluation methods are used as illustrated in Fig.\ref{fig:ope}. 

To account for different factors of a test sequence affecting performance, the testing sequences are further categorized w.r.t. 11 attributes for more ind-depth comparisons:
(1) Illumination Variation (IV, 38/22/21 sequences in TB-100/50/CVPR2013),
(2) Scale Variation (SV, 64/38/28 sequences),
(3) Occlusion (OCC, 49/29/29 sequences),
(4) Deformation (DEF, 44/23/19 sequences),  
(5) Motion Blur (MB, 29/19/12 sequences), 
(6) Fast Motion (FM, 39/25/17 sequences), 
(7) In-Plane Rotation (IPR, 51/29/31 sequences),
(8) Out-of-Plane Rotation (OPR, 63/32/39 sequences),
(9) Out-of-View (OV, 14/11/6 sequences),
(10) Background Clutters (BC, 31/20/21 sequences), and
(11) Low Resolution (LR, 9/8/4 sequences). 
More details on the attributes and their distributions in the benchmark are referred to \cite{trackingBenchmarkPAMI,trackingBenchmark}.

Table.~\ref{table:trackers} lists the 29 evaluated tracking algorithms which are categorized based on representation and search scheme. 
See more details about categorizing these trackers in \cite{trackingBenchmarkPAMI}. 
In TB-CVPR2013, two recent trackers trained by deep convolutional network (CNT\cite{cnnTracker}, SO-DLT\cite{rcnnTracker}) were evaluated using OPE. 

We summarize the performance gain of our AOGTracker in Table.\ref{table:gain}. 
Our AOGTracker obtains significant improvement (more than 12\%) in the 10 subsets in TB-50. Our AOGTracker handles out-of-view situations much better than other trackers since it is capable of re-detecting target objects in the whole image, and it performs very well in the scale variation subset (see examples in the second and fourth rows in Fig.~\ref{fig:BBs}) since it searches over feature pyramid explicitly (with the expense of more computation). 
Our AOGTracker obtains the least improvement in the low-resolution subset since it uses HOG features and the discrepancy between HOG cell-based coordinate and pixel-based one can cause some loss in overlap measurement, especially in the low resolution subset. We will add automatic selection of feature types (e.g., HOG v.s. pixel-based features such as intensity and gradient) according to the resolution, as well as other factors in future work. 

Fig.\ref{fig:TB-AUC} shows success plots of OPE, SRE and TRE in TB-100/50/CVPR2013. Our AOGTracker consistently outperforms all other trackers. We note that for OPE in TB-CVPR2013, although the improvement of our AOGTracker over the SO-DLT\cite{rcnnTracker} is not very big, the SO-DLT utilized two deep convolutional networks with different model update strategies in tracking, both of which are pretrained on the ImageNet \cite{imagenet}. Fig.~\ref{fig:BBs} shows some qualitative results.

   %\vspace{-2mm} 
   \subsection{Analyses of AOG models and the TLP Algorithm}
   To analyze contributions of different components in our AOGTracker, we compare performance of six different variants-- three different object representation schema: AOG with and without structure re-learning (denoted by \textit{AOG} and \textit{AOGFixed} respectively), and whole object template only (i.e., without part configurations, denoted by \textit{ObjectOnly}), and two different inference strategies for each representation scheme: inference with and without temporal DP (denoted by \text{-st} and \textit{-s} respectively). As stated above, we use a very simple setting for temporal DP which takes into account $\Delta t=5$ frames, $[t-5, t]$ in our experiments.
      
      Fig.~\ref{fig:TB-AUC-AOGversions} shows performance comparison of the six variants. \textit{AOG-st} obtains the best overall performance consistently. Trackers with AOG perform better than those with whole object template only. AOG structure re-learning has consistent overall performance improvement. But, we observed that \textit{AOGFixed-st} works slightly better than \textit{AOG-st} on two subsets out of 11, Motion-Blur and Out-of-View, on which the simple intrackability measurement is not good enough. For trackers with AOG, temporal DP helps improve performance, while for trackers with whole object templates only, the one without temporal DP (\textit{ObjectOnly-s}) slightly outperform the one with temporal DP (\textit{ObjectOnly-st}), which shows that we might need strong enough object models in integrating spatial and temporal information for better performance.

\subsection{Comparison with State-of-the-Art Methods}
We explain why our AOGTracker outperforms other trackers on the TB-100 benchmark in terms of representation, online learning and inference. 

\textit{Representation Scheme}. Our AOGTracker utilizes three types of complementary features (HOG+LBP+Color) jointly to capture appearance variations, while most of other trackers use simpler ones (e.g., TLD~\cite{TLD} uses intensity based Haar like features). More importantly, we address the issue of learning the optimal deformable part-based configurations in the quantized space of latent object structures, while most of other trackers focus on either whole objects~\cite{CSK} or implicit configurations (e.g., the random fern forest used in TLD). These two components are integrated in a latent structured-output discriminative learning framework, which improves the overall tracking performance (e.g., see comparisons in Fig.~\ref{fig:TB-AUC-AOGversions}).  

\textit{Online Learning}. Our AOGTracker includes two components which are not addressed in all other trackers evaluated on TB-100: online structure re-learning based on intrackability, and a simple temporal DP for computing optimal joint solution. Both of them improve the performance based on our ablation experiments. The former enables our AOGTracker to capture both large structural and sudden appearance variations automatically, which is especially important for long-term tracking. In addition to improve the prediction performance, the latter improves the capability of maintaining the purity of online collected training dataset. 

\textit{Inference}. Unlike many other trackers which do not handle scale changes explicitly (e.g., CSK~\cite{CSK} and STRUCK~\cite{STRUCK}), our AOGTracker runs tracking-by-parsing in feature pyramid to detect scale changes (e.g., the car example in the second row in Fig.~\ref{fig:BBs}). Our AOGTracker also utilizes a dynamic search strategy which re-detects an object in whole frame if local ROI search failed. For example, our AOGTracker handles out-of-view situations much better than other trackers due to the re-detection component (see examples in the fourth row in Fig.~\ref{fig:BBs}). 

\textit{Limitations}. All the performance improvement stated above are obtained at the expense of more computation in learning and tracking.
Our AOGTracker obtains the least improvement in the low-resolution subset since it uses HOG features and the discrepancy between HOG cell-based coordinate and pixel-based one can cause some loss in overlap measurement, especially in the low resolution subset. We will add automatic selection of feature types (e.g., HOG v.s. pixel-based features such as intensity and gradient) according to the resolution, as well as other factors in future work.   
        
         \begin{figure}
                        	\centering
                        	\includegraphics[width=0.5\textwidth]{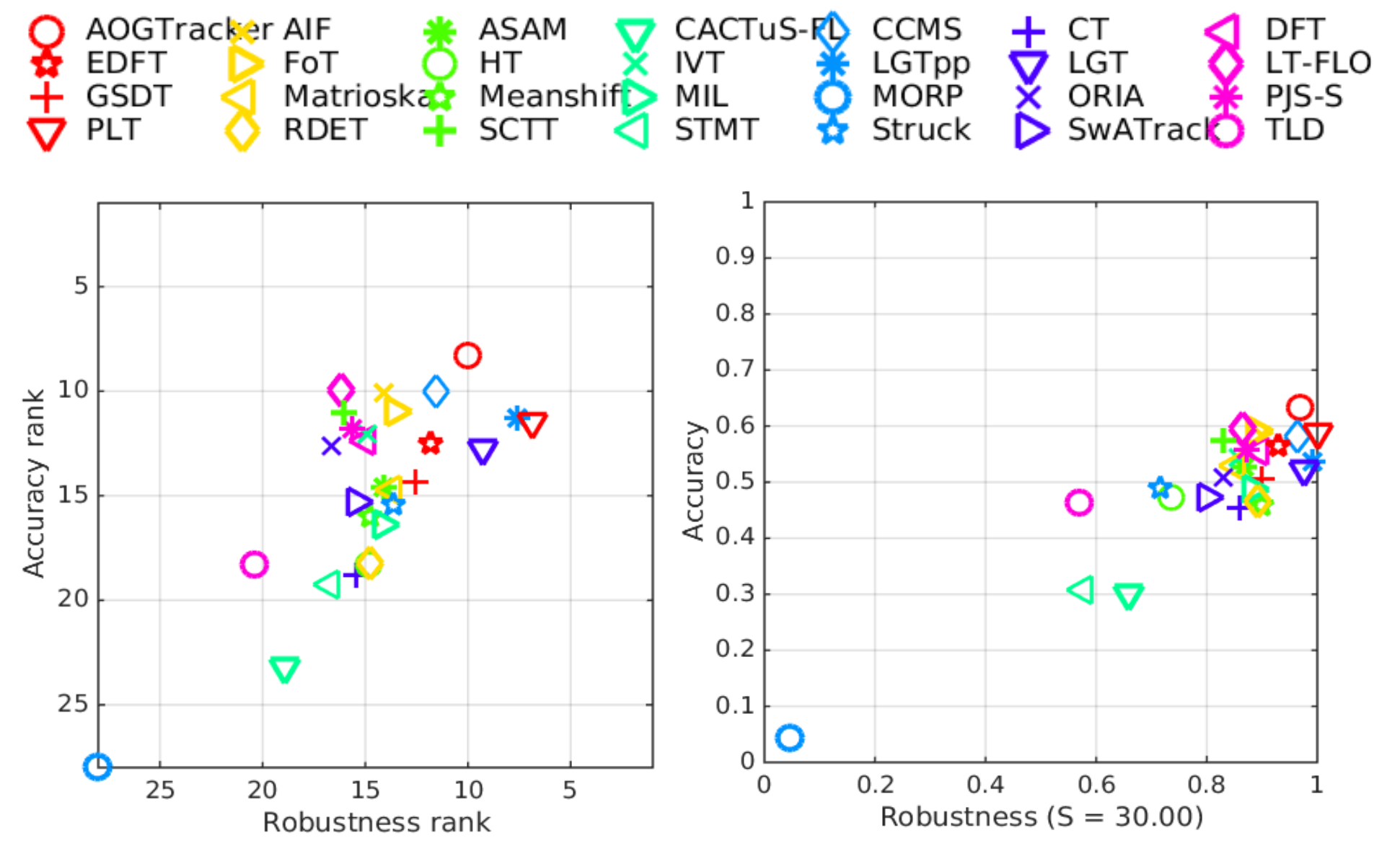}
                        	\caption{Performance comparison in VOT2013. \textit{Left}: Ranking plot for the baseline experiment. The smaller the rank number is, the better a tracker is w.r.t. accuracy and/or robust (i.e., the right-top region indicates better performance) \textit{Right}: Accuracy-Robustness plot. The larger the rate is, the better a tracker is.}	
                        	\label{fig:vot2013}	
                        	%\vspace{-4mm}
                        \end{figure}
                        
              \begin{figure}
                             	\centering
                             	\includegraphics[width=0.5\textwidth]{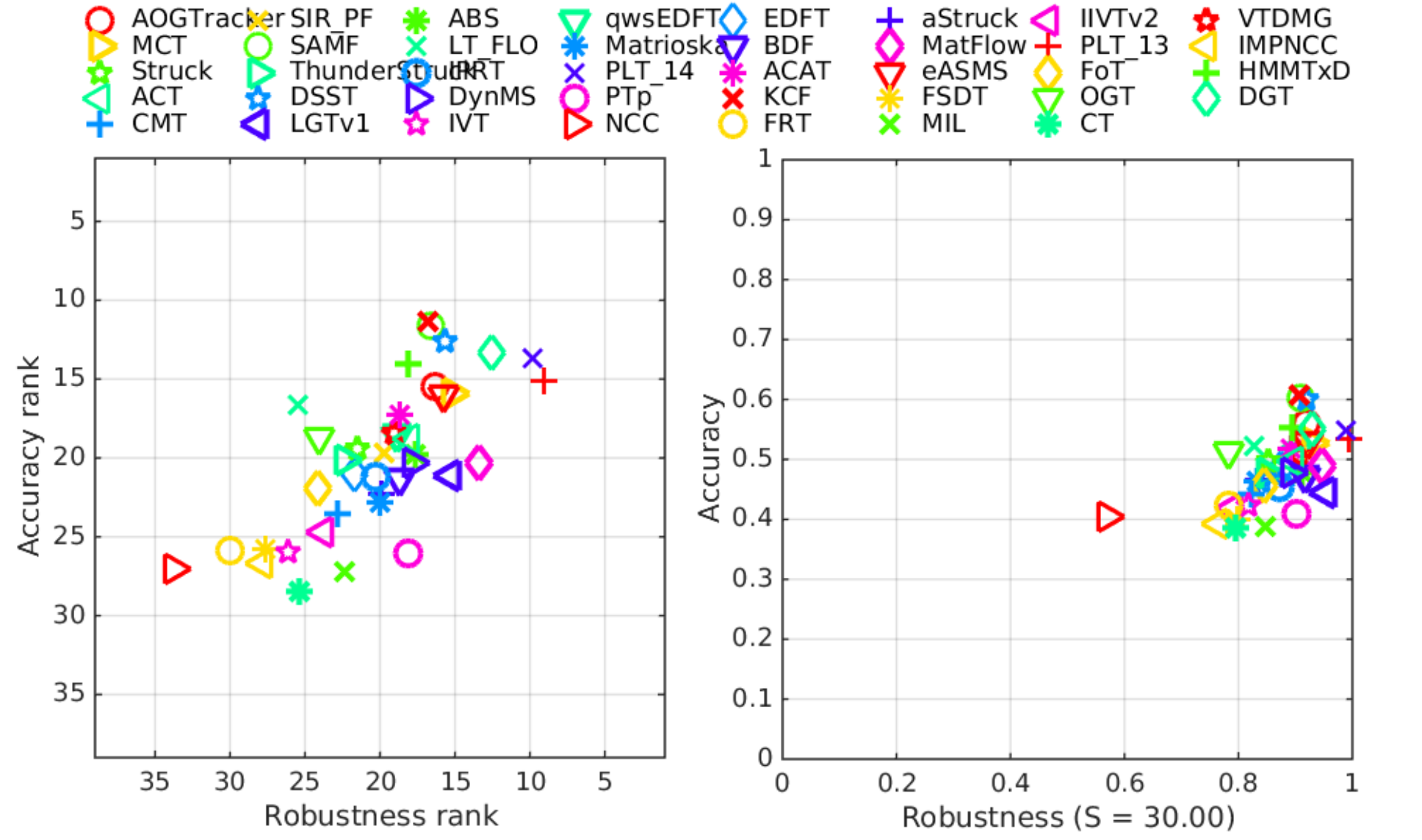}
                             	\caption{Performance comparison in VOT2014. %\textit{Left}: Ranking plot for the baseline experiment. \textit{Right}: Accuracy-Robustness (AR) plot.
                             		}	
                             	\label{fig:vot2014}	
                             	 	%\vspace{-4mm}
                             \end{figure}

%\vspace{-2mm}
\subsection{Results on VOT}
 In VOT, the evaluation focuses on short-term tracking (i.e., a tracker is not expected to perform re-detection after losing a target object), so the evaluation toolkit will re-initialize a tracker after it loses the target (w.r.t. the condition the overlap between the predicted bounding box and the ground-truth one drops to zero) with the number of failures counted. In VOT protocol, a tracker is tested on each sequence multiple times. The performance is measured in terms of accuracy and robustness. \textbf{Accuracy} is computed as the average of per-frame accuracies which themselves are computed by taking the average over the repetitions. \textbf{Robustness} is computed as the average number of failure times over repetitions.

We integrate our AOGTracker in the latest VOT toolkit\footnote{Available at https://github.com/votchallenge/vot-toolkit, version 3.2} to run  experiments with the baseline protocol and to generate plots \footnote{The plots for VOT2013 and 2014 might be different compared to those in the original VOT reports \cite{vot2013,vot2014} due to the new version of  vot-toolkit.}. 

The VOT2013 dataset \cite{vot2013} has 16 sequences which was selected from
a large pool such that various visual phenomena like occlusion and illumination changes, were still represented well
within the selection. 7 sequences are also used in TB-100. There are 27 trackers evaluated. The readers are referred to the VOT technical report~\cite{vot2013} for details.
%: two background subtraction based methods MORP\cite{vot2013} and STMT\cite{vot2013}, two key-point feature based methods Matrioska\cite{Matrioska} and SCTT\cite{vot2013}, IVT\cite{IVT}, MS\cite{KMS} and its improved version CCMS\cite{vot2013}, DFT\cite{DFT} and EDFT\cite{EDFT}, AIF\cite{AIF} and CactusFI\cite{vot2013}, MIL\cite{MIL}, STRUCK\cite{STRUCK} and its derivative PLT\cite{vot2013}, CT\cite{CT} and its derivative RDET\cite{RDET}, ORIA\cite{ORIA} and ASAM\cite{vot2013}, GSDT\cite{GSDT}, some part-based methods including HT\cite{HT}, LGT\cite{LGT} and its extension LGTpp\cite{LGTpp}, LT-FLO\cite{LTFLO}, FOT\cite{FOT}, TLD\cite{TLD}, and a few baseline trackers implemented by authors of the benchmark. 

Fig.\ref{fig:vot2013} shows the ranking plot and AR plot in VOT2013. Our AOGTracker obtains the best accuracy while its robustness is slightly worse than three other trackers (i.e., PLT\cite{vot2013}, LGT\cite{LGT} and LGTpp\cite{LGTpp}, and PLT was the winner in VOT2013 challenge). Our AOGTracker obtains the best overall rank.

The VOT2014 dataset \cite{vot2014} has 25 sequences extended from VOT2013. The annotation is based on rotated bounding box instead of up-right rectangle. There are 33 trackers evaluated. Details on the trackers are referred to \cite{vot2014}. 
Fig.\ref{fig:vot2014} shows the ranking plot and AR plot. Our AOGTracker is comparable to other trackers. One main limitation of AOGTracker is that it does not handle rotated bounding boxes well.  

The VOT2015 dataset~\cite{vot2015} consists of 60 short sequences (with rotated bounding box annotations) and VOT-TIR2015 comprises 20 sequences (with bounding box annotations). There are 62 and 28 trackers evaluated in VOT2015 and VOT-TIR2015 respectively. Our AOGTracker obtains $51$\% and 65\% (tied for third place) in accuracy in VOT2015 and VOT-TIR2015 respectively. The details are referred to the reports~\cite{vot2015} due to  space limit here.

%\vspace{-3mm}
\section{Discussion and Future Work}\label{sec:conclusion}
We have presented a tracking, learning and parsing (TLP) framework and derived a spatial dynamic programming (DP) and a temporal DP algorithm for online object tracking with AOGs. We also have presented a method of online learning object AOGs including its structure and parameters. In experiments, we test our method on two main public benchmark datasets and experimental results show better or comparable performance. 
 
 In our on-going work, we are studying \textit{more flexible computing schemes} in tracking with AOGs. The compositional property embedded in an AOG naturally leads to different bottom-up/top-down computing schemes such as the three computing processes studied by Wu and Zhu~\cite{3Channels}. We can track an object by matching the object template directly (i.e. $\alpha$-process), or computing some discriminative parts first and then combine them into object ($\beta$-process), or doing both ($\alpha+\beta$-process, as done in this paper). In tracking, as time evolves, the object AOG might grow through online learning, especially for objects with large variations in long-term tracking. Thus, faster inference is entailed for the sake of real time applications. We are trying to learn near optimal decision policies for tracking using the framework proposed by Wu and Zhu~\cite{DecisionPolicy}. 
  
 In our future work, we will extend the TLP framework by incorporating generic category-level AOGs \cite{DisAOT} to scale up the TLP framework. The generic AOGs are pre-trained offline (e.g., using the PASCAL VOC \cite{VOC} or the imagenet \cite{imagenet}), and will help the online learning of specific AOGs for a target object (e.g., help to maintain the purity of the positive and negative datasets collected online). The generic AOGs will also be updated online together with the specific AOGs. By integrating generic and specific AOGs, we aim at the life-long learning of objects in videos without annotations. Furthermore, we are also interested in integrating scene grammar \cite{sceneGrammar} and event grammar \cite{eventGrammar} to leverage more top-down information.

\ifCLASSOPTIONcompsoc
  % The Computer Society usually uses the plural form
  \section*{Acknowledgments}
\else
  % regular IEEE prefers the singular form
  \section*{Acknowledgment}
\fi

This work is supported by the DARPA SIMPLEX Award N66001-15-C-4035, the ONR MURI grant N00014-16-1-2007, and NSF IIS-1423305. T. Wu was also supported by the ECE startup fund 201473-02119 at NCSU.  We thank Steven Holtzen for proofreading this paper. We also gratefully acknowledge the support of NVIDIA Corporation with the donation of one GPU.

% Can use something like this to put references on a page
% by themselves when using endfloat and the captionsoff option.
\ifCLASSOPTIONcaptionsoff
  \newpage
\fi

% trigger a \newpage just before the given reference
% number - used to balance the columns on the last page
% adjust value as needed - may need to be readjusted if
% the document is modified later
%\IEEEtriggeratref{8}
% The "triggered" command can be changed if desired:
%\IEEEtriggercmd{\enlargethispage{-5in}}

% references section

% can use a bibliography generated by BibTeX as a .bbl file
% BibTeX documentation can be easily obtained at:
% http://mirror.ctan.org/biblio/bibtex/contrib/doc/
% The IEEEtran BibTeX style support page is at:
% http://www.michaelshell.org/tex/ieeetran/bibtex/
\bibliographystyle{IEEEtran} %{IEEEtran}
% argument is your BibTeX string definitions and bibliography database(s)
%\bibliography{IEEEabrv,../bib/paper}
\bibliography{IEEEabrv,egbib}
%
% <OR> manually copy in the resultant .bbl file
% set second argument of \begin to the number of references
% (used to reserve space for the reference number labels box)
%\begin{thebibliography}{1}
%
%\bibitem{IEEEhowto:kopka}
%H.~Kopka and P.~W. Daly, \emph{A Guide to \LaTeX}, 3rd~ed.\hskip 1em plus
%  0.5em minus 0.4em\relax Harlow, England: Addison-Wesley, 1999.
%
%\end{thebibliography}

% biography section
% 
% If you have an EPS/PDF photo (graphicx package needed) extra braces are
% needed around the contents of the optional argument to biography to prevent
% the LaTeX parser from getting confused when it sees the complicated
% \includegraphics command within an optional argument. (You could create
% your own custom macro containing the \includegraphics command to make things
% simpler here.)
%\begin{IEEEbiography}[{\includegraphics[width=1in,height=1.25in,clip,keepaspectratio]{mshell}}]{Michael Shell}
% or if you just want to reserve a space for a photo:

\begin{IEEEbiography}[ {
\includegraphics*[width=.9in,clip]{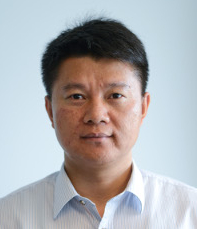} } ]{Tianfu Wu}
received Ph.D. degree in Statistics from University of California, Los Angeles (UCLA) in 2011.
He joined NC State University in August 2016 as a Chancellor’s Faculty Excellence Program cluster hire in Visual Narrative. He is currently assistant professor in the Department of Electrical and Computer Engineering. His research focuses on explainable and improvable visual Turing test and robot autonomy through life-long communicative learning by pursuing a unified framework for machines to ALTER (Ask, Learn, Test, Explain, and Refine) recursively in a principled way: (i) Statistical learning of large scale and highly expressive hierarchical and compositional models from visual big data (images and videos). (ii) Statistical inference by learning near-optimal cost-sensitive decision policies. (iii) Statistical theory of performance guaranteed learning algorithm and optimally scheduled inference procedure.
\end{IEEEbiography}

% if you will not have a photo at all:
\begin{IEEEbiography}[ {
\includegraphics*[width=.9in,clip]{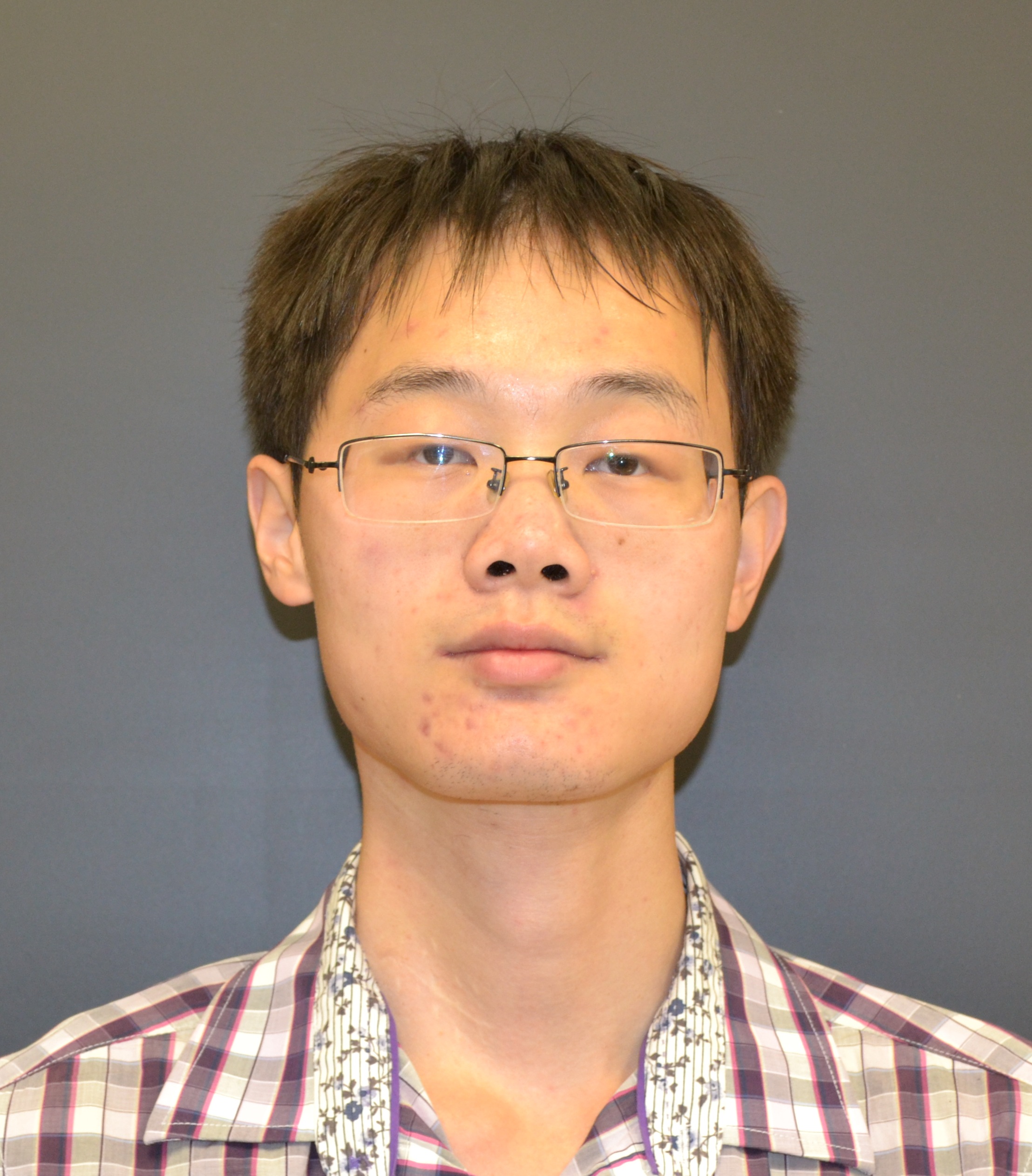} } ]{Yang Lu}
is currently Ph. D. student in the Center for Vision, Cognition, Learning and Autonomy at the University of California, Los
Angeles. He received B.S. degree and M.S. degree in Computer Science from
Beijing Institute of Technology, China, in 2009 and in 2012 respectively. He received the University Fellowship from UCLA and National
Fellowships from Department of Education at China. His current research interests
include Computer Vision and Statistical Machine Learning. Specifically, his research interests focus on statistical modeling of natural images and videos, and
structure learning of hierarchical models.
\end{IEEEbiography}

% insert where needed to balance the two columns on the last page with
% biographies
%\newpage

\begin{IEEEbiography}[ {
\includegraphics*[width=.9in,clip]{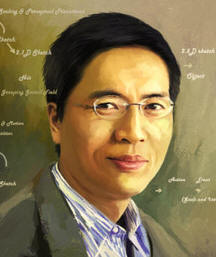} } ]{Song-Chun Zhu}
received Ph.D. degree from Harvard University in 1996.
He is currently professor of Statistics and Computer Science at
UCLA, and director of Center for Vision, Cognition, Learning and Autonomy.
He received a number of honors, including the J.K. Aggarwal prize
from the Int'l Association of Pattern Recognition in 2008 for
"contributions to a unified foundation for visual pattern
conceptualization,  modeling,  learning, and inference",
the David Marr Prize in 2003 with Z. Tu et al. for image parsing,
twice Marr Prize honorary nominations in 1999 for texture modeling and
in 2007 for object modeling with Z. Si and Y.N. Wu. He received the
Sloan Fellowship in 2001, a US NSF Career Award in 2001, and an US ONR
Young Investigator Award in 2001.  He received the Helmholtz Test-of-time award in ICCV 2013, and he is a Fellow of IEEE since 2011. 
\end{IEEEbiography}

% You can push biographies down or up by placing
% a \vfill before or after them. The appropriate
% use of \vfill depends on what kind of text is
% on the last page and whether or not the columns
% are being equalized.

%\vfill

% Can be used to pull up biographies so that the bottom of the last one
% is flush with the other column.
%\enlargethispage{-5in}

% that's all folks
\end{document}